\documentclass[10pt,twocolumn,letterpaper]{article}
\pdfoutput=1 
\usepackage{iccv}
\usepackage{times}
\usepackage{epsfig}
\usepackage{graphicx}
\usepackage{amsmath}
\usepackage{amssymb}
\usepackage{mathdef}
\usepackage{macros}
\usepackage{xcolor}
\usepackage{authblk}

\usepackage[breaklinks=true,bookmarks=false]{hyperref}

\iccvfinalcopy 

\def\iccvPaperID{} 

\ificcvfinal\fi

\newcommand{\Figs}{Figs/}

\newcommand\Alpha{\mathrm{A}}
\newcommand\Beta{\mathrm{B}}

\begin{document}

\title{Continuous Cost Aggregation for Dual-Pixel Disparity Extraction }

\author[1,2,$\ast$]{Sagi Monin}
\author[2,$\star$]{Sagi Katz}
\author[2]{Georgios Evangelidis}
\affil[1]{Department of ECE, Technion, Israel}
\affil[2]{Snap Inc.}

\maketitle
\ificcvfinal\thispagestyle{empty}\fi

\begin{abstract}
    Recent works have shown that depth information can be obtained from Dual-Pixel (DP) sensors. 
    A DP arrangement provides two views in a single shot, thus resembling a stereo image pair with a tiny baseline. However, the different point spread function (PSF) per view, as well as the small disparity range, makes the use of typical stereo matching algorithms problematic. To address the above shortcomings, we propose a Continuous Cost Aggregation (CCA) scheme within a semi-global matching framework that is able to provide accurate continuous disparities from DP images. The proposed algorithm fits parabolas to matching costs and aggregates parabola coefficients along image paths. The aggregation step is performed subject to a quadratic constraint that not only enforces the disparity smoothness but also maintains the quadratic form of the total costs. This gives rise to an inherently efficient disparity propagation scheme with a pixel-wise minimization in closed-form. Furthermore, the continuous form allows for a robust multi-scale aggregation that better compensates for the varying PSF. Experiments on DP data from both DSLR and phone cameras show that the proposed scheme attains state-of-the-art performance in DP disparity estimation.  \let\thefootnote\relax\footnotetext{$\ast$ This work was done as part of Sagi Monin's internship at Snap Inc.}
    \let\thefootnote\relax\footnotetext{$\star$ This work was done as part of Sagi Katz's work at Snap Inc.}    
\end{abstract}

\interfootnotelinepenalty=10000

\section{Introduction}
\label{sec:intro}



Depth is an important cue for disparate applications such as recognition, navigation, and graphic manipulation. The respective sensors typically use stereo, structured light, or time-of-flight techniques~\cite{Sarbolandi2015KinectRS, Horaud2016overview} for depth estimation. These techniques require special hardware while the form factors may not easily get miniaturized. With simpler hardware, deep neural networks (DNN) obtain depth from monocular images~\cite{MING202114}, but still show sub-optimal results in comparison to depth sensors.

\begin{figure}[t]
    \setlength\abovecaptionskip{-0.7\baselineskip} 
    \setlength\belowcaptionskip{-20pt} 
  \begin{center}
     \begin{tabular}{c@{~}c@{~}c@{~}c}

            \raisebox{0.8cm}{\rotatebox[origin=c]{90}{~\scriptsize{DSLR camera~\cite{Punnappurath-ICCP2020-modelingDefocus}}}}&
           \includegraphics[width=0.3\linewidth]{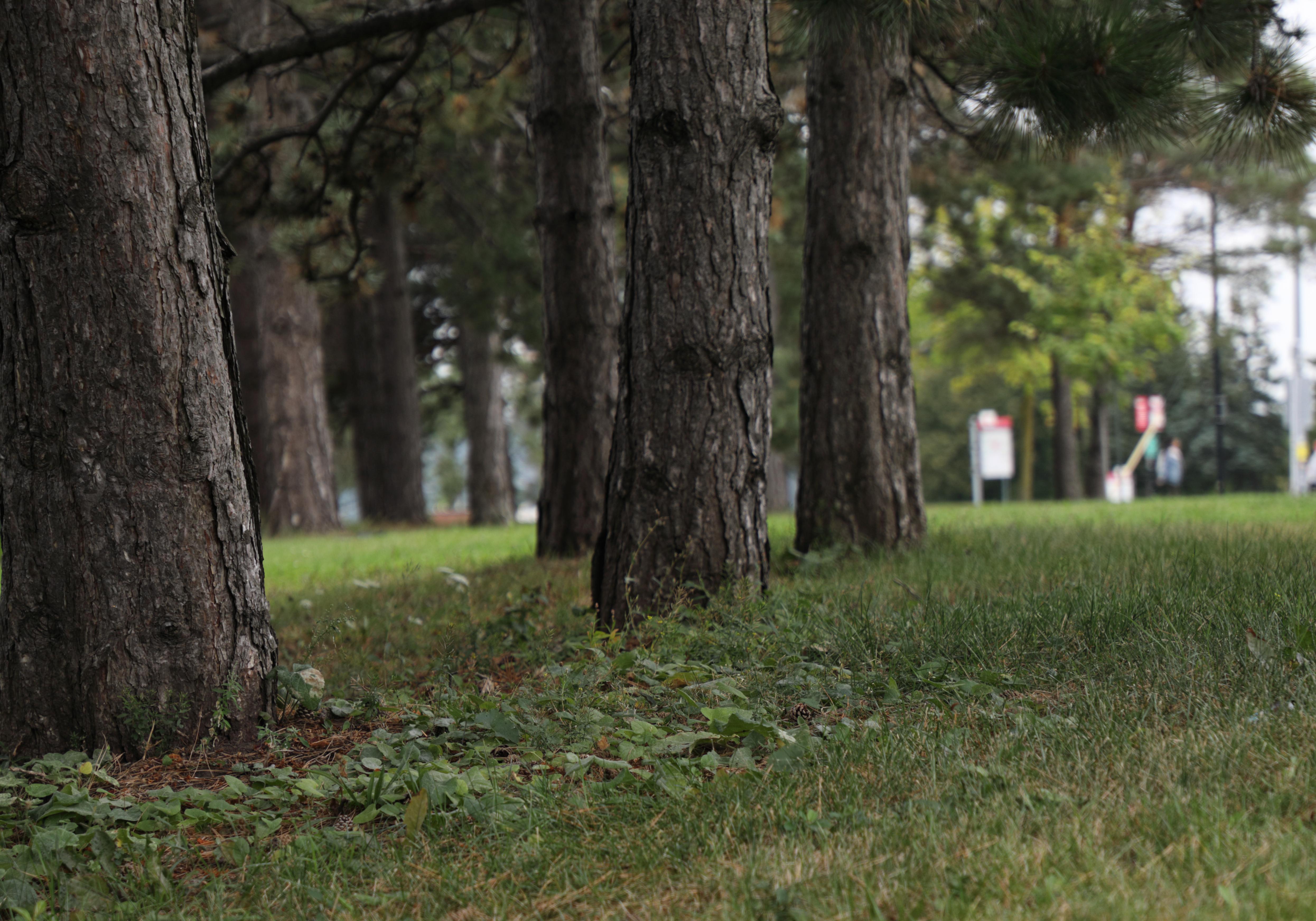} &
            \includegraphics[width=0.3\linewidth]{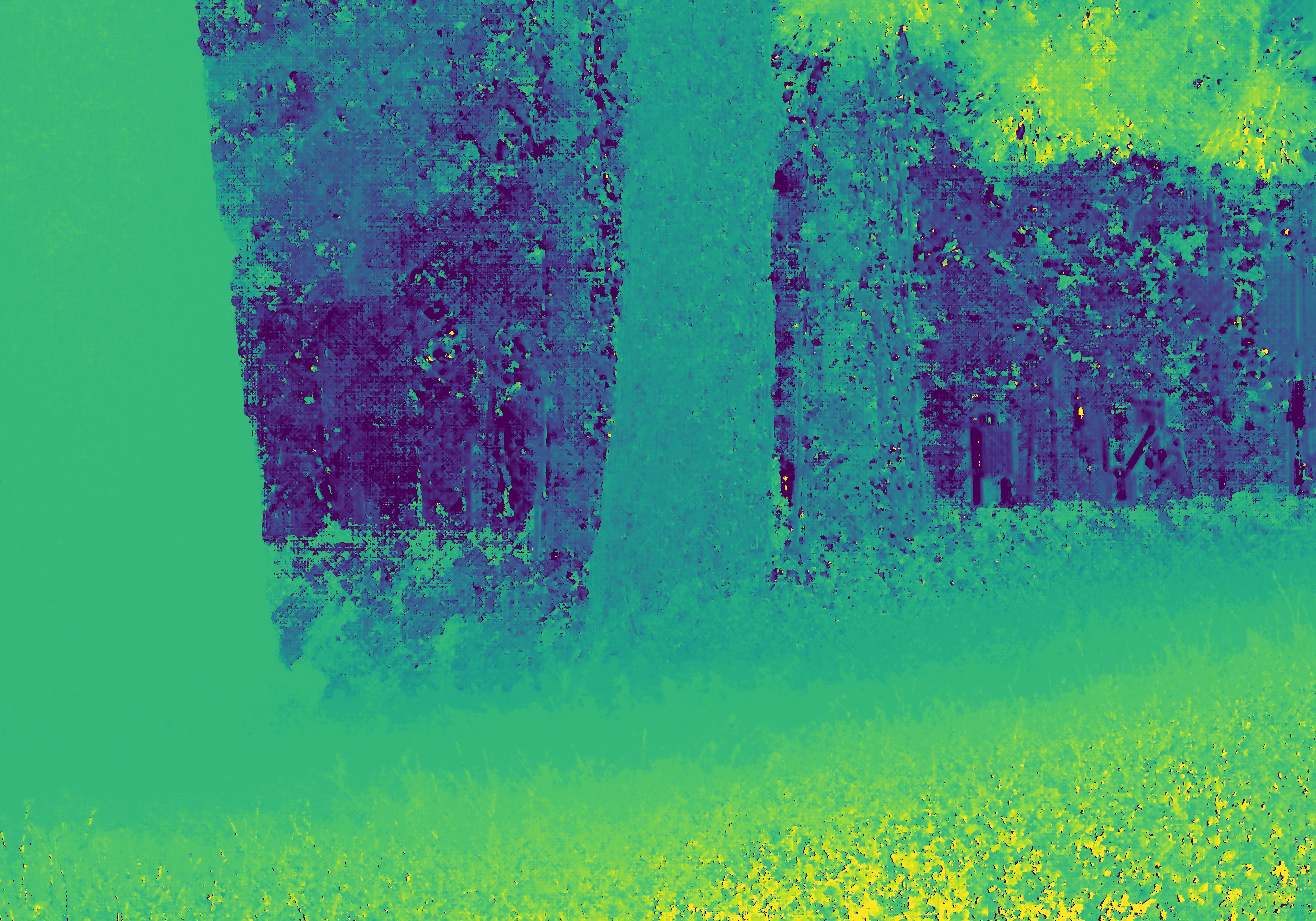} & \includegraphics[width=0.3\linewidth]{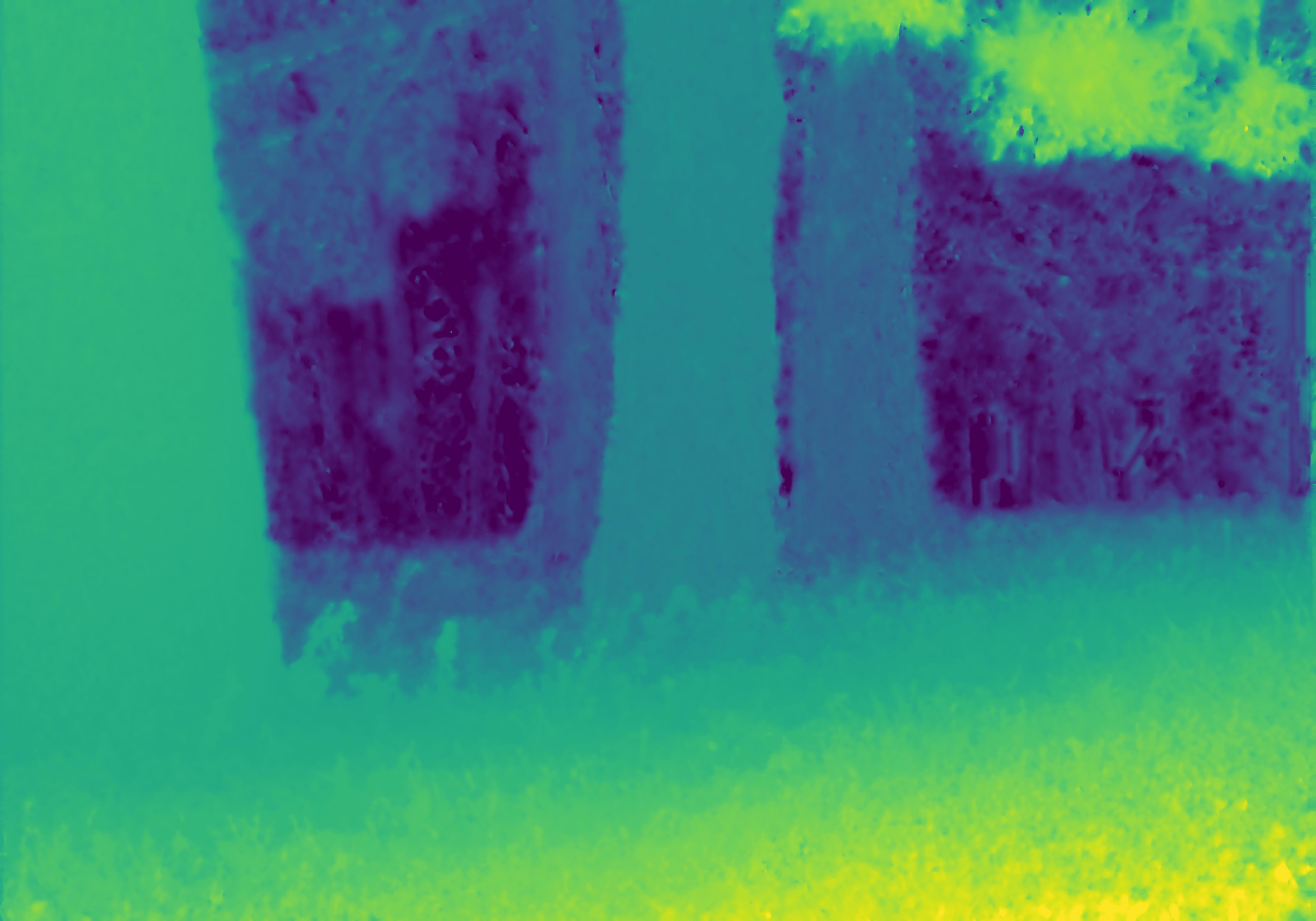} \\

        \raisebox{0.8cm}{\rotatebox[origin=c]{90}{~\scriptsize{Phone camera~\cite{Wadhwa_SIGGRAPH2018_syntheticDoF}}}}&         \includegraphics[width=0.3\linewidth]{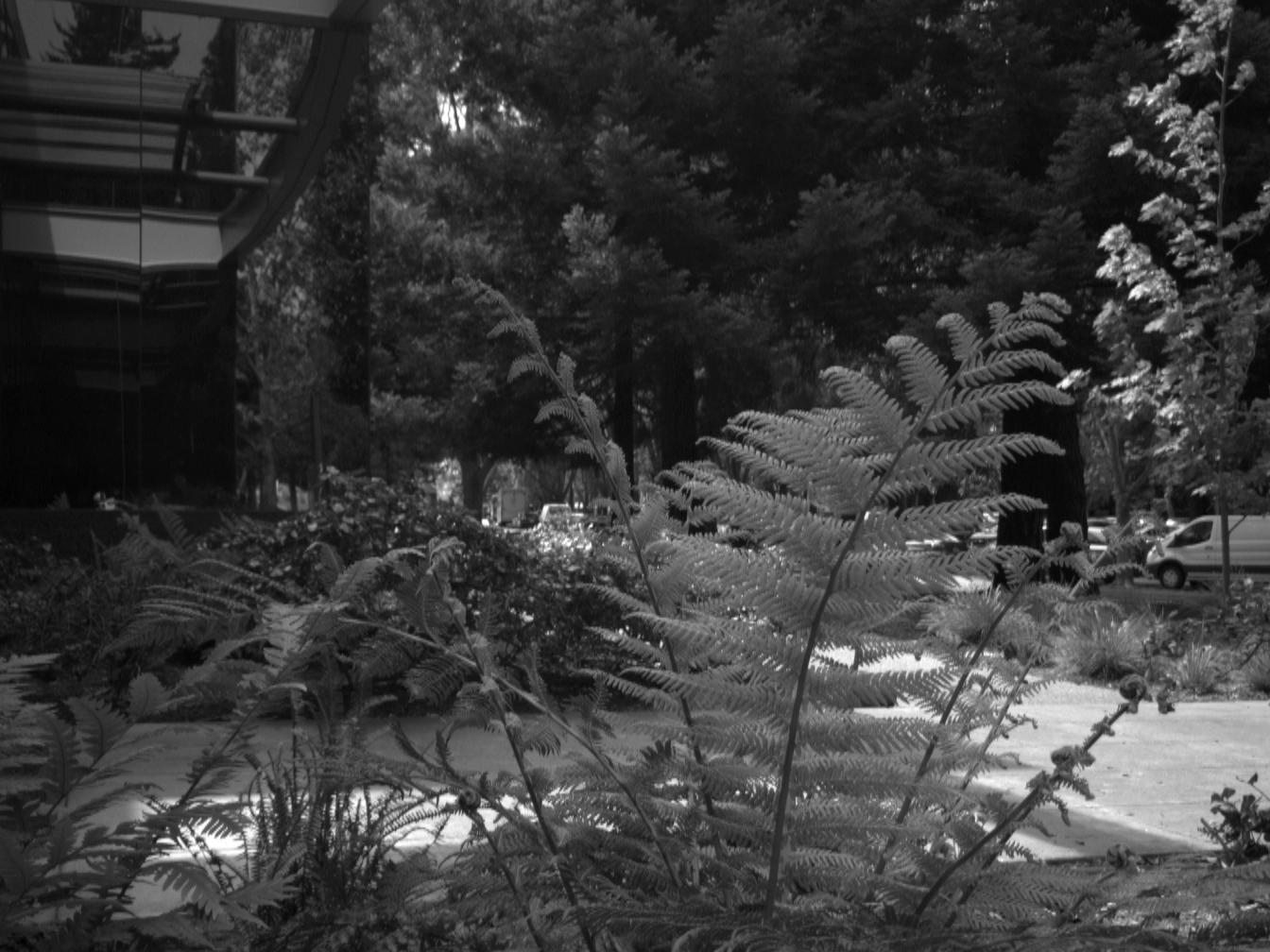} &
        \includegraphics[width=0.3\linewidth]{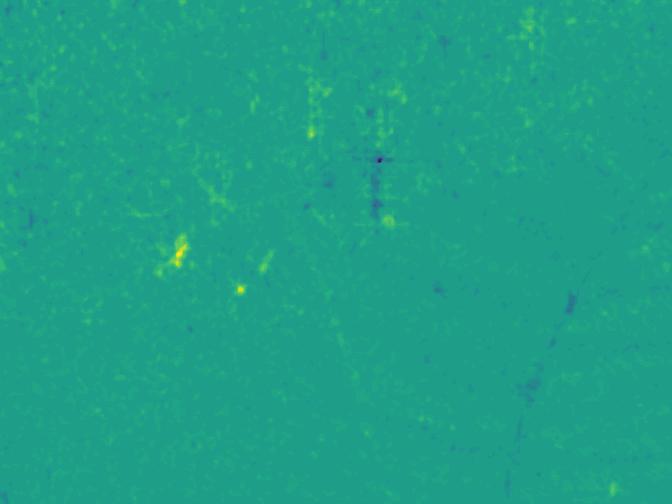} & \includegraphics[width=0.3\linewidth]{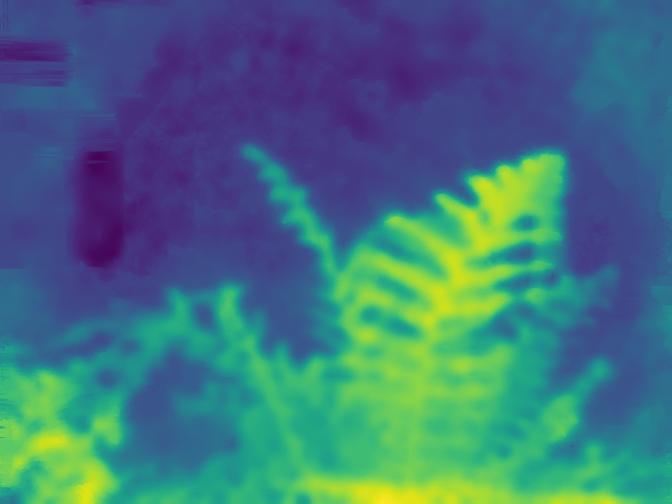} \\
       &\footnotesize{(a) Left DP image} &
       \footnotesize{(b) SGM disparities} &
       \footnotesize{(c) CCA disparities}

    \end{tabular}
      
       

       

    \end{center}
\caption{
SGM is a widely-used stereo disparity estimation method but it underperforms on DP images, in particular when the disparities are quite small. Instead, the proposed Continouous Cost Aggregation (CCA) algorithm  attains state-of-the-art performance in DP disparity estimation. Results from DSLR (\emph{top}) and Phone cameras (\emph{bottom}) are shown; CCA estimates disparities in the range $[-12,6]$ (DSLR) and $[-1.3,0.5]$ (Phone), respectively. 
}


 \label{fig:Teaser}
\end{figure}

\begin{figure*}[t]
    \setlength\abovecaptionskip{-0.6\baselineskip} 
    \setlength\belowcaptionskip{-15pt} 
  \begin{center}
  \includegraphics[width=1\linewidth]{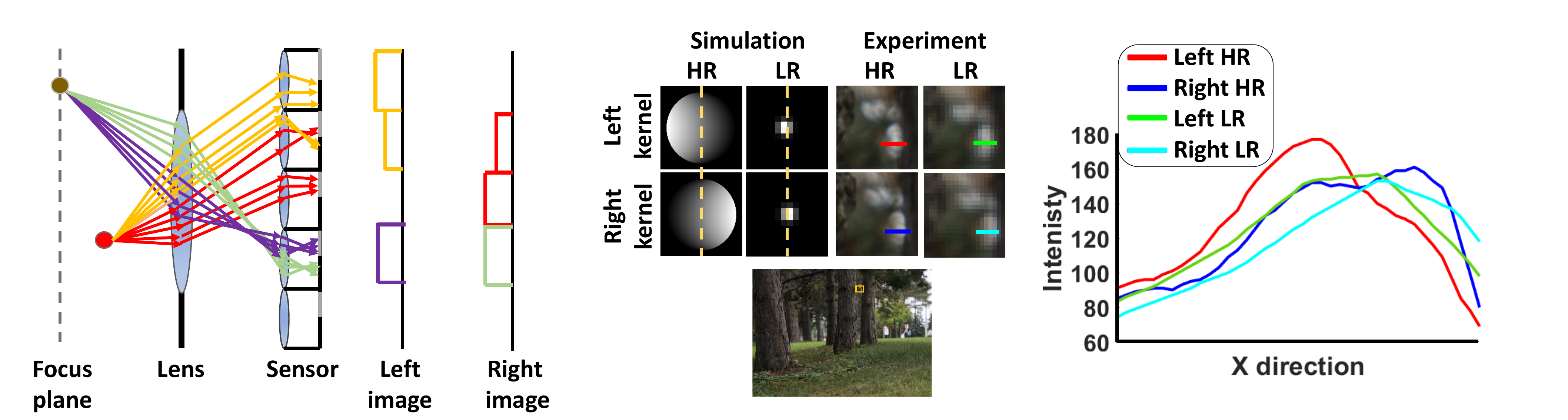}
  \end{center}
 \caption{
A simple illustration of the DP image formation (\emph{left}). Unlike in-focus scene points, out-of-focus points will be projected to different pixels at the two views, hence the disparity. Based on the DP image model, the two views have different PSF, here represented as the 2D kernels of ~\cite{Punnappurath-ICCP2020-modelingDefocus} (\emph{middle}). The higher the resolution is, the more different the PSFs are, and the less similar to stereo-image the model is. As seen, the relation between the intensity profiles of the two LR views of a cross-section is better explained by a shift, compared to the HR profiles(\emph{right}).
 For a more detailed study on DP image formation, we refer the reader to~\cite{Punnappurath-ICCP2020-modelingDefocus, Wadhwa_SIGGRAPH2018_syntheticDoF, Garg-ICCV2019-learningDual}.
 }
 \label{fig:DP_fig}
\end{figure*}

Interestingly, dual-pixel (DP) sensors, which are becoming increasingly common on modern cameras, have been recently employed for depth estimation \cite{Wadhwa_SIGGRAPH2018_syntheticDoF, Garg-ICCV2019-learningDual}. Although these sensors were originally designed to assist with auto-focus, the resemblance to tiny-baseline stereo view as well as being inherently rectified made them attractive for depth computation. However, a DP image is not strictly equivalent to a stereo image pair\cite{Punnappurath-ICCP2020-modelingDefocus} and the use of off-the-shelf stereo algorithms on DP images seems to be inadequate (\figref{fig:Teaser}). This is because DP sensors have a split-pixel arrangement, thereby two sub-aperture views from opposing directions are delivered. As a result, the point spread function (PSF) is different per view and depth does not correspond to pure shifts (\figref{fig:DP_fig}). In addition, even if high-resolution DP images are provided, processing on edge devices~\cite{Wadhwa_SIGGRAPH2018_syntheticDoF} requires working on lower resolution DP-data. This, along with the tiny baseline, implies very small disparity values (\figref{fig:Teaser}). Therefore, unlike traditional stereo algorithms, sub-pixel disparity estimation becomes a prerequisite. 

To address the above-mentioned shortcomings, we propose a continuous cost optimization in a semi-global manner. Inspired by Semi-Global Matching (SGM)~\cite{Hirschmuller-PAMI2008-sgm}, we adopt the pathwise optimization from multiple directions through the image. While SGM aggregates cost at integer disparities, our algorithm aggregates the coefficients of parabolas that represent costs of continuous disparities, thus easily modelling small and subpixel disparities. This aggregation step is performed subject to a constraint that enforces smoothness between successive pixels along the paths. By utilizing a quadratic constraint, the total cost remains a parabola, which gives rise to a very efficient disparity propagation strategy and makes the minimization very simple. The inherent efficiency of the proposed algorithm rises from cost aggregation over a 2D image space, unlike the aggregation of SGM over the 3D image-disparity space.
In addition, the continuous scheme favors a multi-scale approach that fuses disparity estimation from multiple resolutions, thus better dealing with varying PSFs and improving disparity estimation in blurred regions (\figref{fig:DP_fig}). 

Our contributions are summarized as follows:
\begin{enumerate}
    \item We propose a Continuous Cost Aggregation (CCA) scheme that results in an  efficient semi-global solution to extract disparity from DP images. 
    
    \item The proposed algorithm includes a multi-scale disparity fusion that is able to rectify errors from previous scale levels. 
        
    \item Our algorithm attains state-of-the-art (SOTA) performance in DP disparity estimation. Unlike the learning-based approaches, the algorithm does not depend on training data and attains a satisfactory performance across diverse data-sets.
\end{enumerate}

It is noteworthy that CCA has the potential to provide disparities from traditional stereo images. Despite our focus on DP stereo, we include an experiment and show that CCA performs similarly to SGM in standard large-baseline stereo matching, with lower space and time complexity.


\section{Related Work}
\label{sec:related_work}
\textbf{Depth from stereo}
Early works on depth estimation relied on two-view geometry and stereo matching~\cite{scharstein2002taxonomy, Hamzah-Sensors2016}, aiming to estimate the per-pixel disparity between two images. A large variety of methods has been proposed including local and global algorithms. While local algorithms work on a per-pixel basis and typically solve an optimization problem per pixel, global algorithms impose dependencies between neighboring pixels, e.g., Markov Random Fields (MRF)~\cite{boykov2001fast}, thus solving a single yet large optimization problem. As a result, global algorithms generally outperform local algorithms at the expense of high computational cost. To obtain continuous disparities, stereo algorithms either fit a parabola around the lowest-cost disparity~\cite{scharstein2002taxonomy,Hirschmuller-PAMI2008-sgm} or integrate interpolation kernels into the cost function~\cite{Psarakis-ICCV2005-ecc}, with parabola fitting as the dominant case. 


As mentioned, our algorithm is related to the SGM~\cite{Hirschmuller-PAMI2008-sgm} approach that mitigates the above mentioned performance limitations. 
SGM imposes a smoothness term along straight paths, thus resulting in an efficient algorithm. The algorithm has four main steps: cost-volume calculation, cost aggregation, winner-take-all optimization, and sub-pixel refinement step. Based on SGM, many variants have been proposed: incorporating DNN for cost-volume~\cite{Eigen-NeurIPS2014-multiscale, shaked2017improved, seki2017sgm}, updating aggregation step~\cite{facciolo2015mgm, Schonberger_2018_ECCV}, hardware implementation for edge devices~\cite{rahnama2018r3sgm,schumacher2014matching,banz2010real}, and memory efficient adaptations~\cite{hirschmuller2012memory, lee2017memory}. 

\textbf{Depth from single image sensor} 
Interestingly, depth estimation is not limited to stereo or multiple cameras and monocular depth estimation has also been investigated. Traditional and early methods mainly relied on inverting depth-dependent PSF \cite{Ens-PAMI1993-depthFromFocus,levin2007image,Haim2018}. More recent work, however, relies on deep learning and trains DNNs to estimate depth~\cite{Fu-CVPR2018-deepRegression, Maximov-CVPR2020-focusOnDefocus}. In this context, significant progress has been made in estimating depth from a single RGB image and plenty of methods have been proposed; for a detailed review, we refer the reader to~\cite{MING202114,Masoumian2022}.

More similar to the problem in question, DP images have been also used for depth estimation. 
In~\cite{Wadhwa_SIGGRAPH2018_syntheticDoF}, a classic local stereo matching method is combined with an up-sampling bilateral filter. Follow up efforts used DNNs for depth estimation; \cite{Garg-ICCV2019-learningDual} explicitly modeled the affine relationship between inverse depth and disparity and trained a network to estimate depth up to this ambiguity. In~\cite{Kim_2023_CVPR} a DNN used for stereo estimation is fine-tuned for DP using a self-supervised approach;~\cite{Yinda-ECCV2020-du2net} fed a DNN with fused data from stereo and DP cameras in order to exploit the complementary errors of depth estimation from each sensor alone. The authors of \cite{Punnappurath-ICCP2020-modelingDefocus} suggested modeling the different defocus blur in each of the left and right images to estimate depth. Despite the good performance, their method is slow as it requires solving an optimization problem for each patch in the image. The above methods have also emphasized that, although DP data is available from both smartphones and DSLR cameras, the difference in optical aberrations and depth-of-field (DOF) between devices leads to severe differences in the performance of the algorithms. 

It is noteworthy to mention that DP sensors have been also used for other applications, such as all-in-focus image generation~\cite{Yang_2023_CVPR, Abuolaim-ECCV2020-defocus, Abuolaim_2022_WACV, xin2021defocus}, reflection removal~\cite{Punnappurath-CVPR2019-reflection}, improved auto-focusing~\cite{Herrman-CVPR2020-learningAF}, 3D face reconstruction~\cite{kang2021facial} and multi-view motion synthesis~\cite{Abuolaim_2022_WACV2}. A large part of these algorithms rely on DNNs, thus requiring a large amount of data with ground-truth. To this end, DP image simulators have been proposed~\cite{Pan_CVPR2021_dpexploration,Abuolaim_2021_ICCV}. 

\section{Continuous Cost Aggregation}
\label{sec:csgm}
\begin{figure}[t]
    \setlength\abovecaptionskip{-0.7\baselineskip} 
    \setlength\belowcaptionskip{-20pt} 
   \begin{center}
     \includegraphics[width=0.98\linewidth]{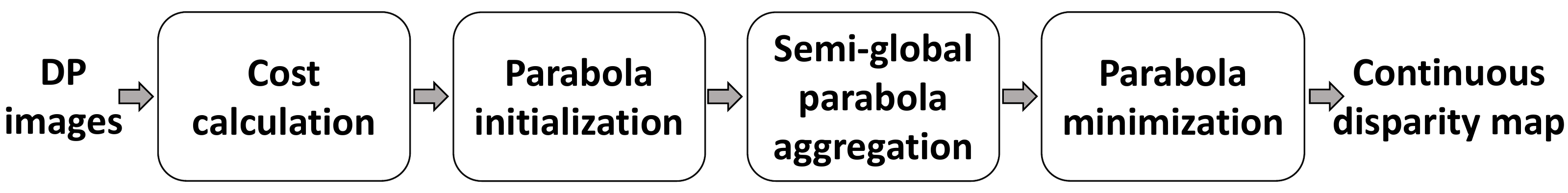} \\
    \end{center}
\caption{After the initial cost calculation, CCA initializes a parabola around the integer disparity that corresponds to the minimum cost, thus representing a continuous cost per pixel. Then, along 1D paths, parabola coefficients are aggregated subject to a smoothness constraint. The total cost of all the paths is finally minimized to provide a semi-globally smoothed disparity map. Note that the minimization has a simple closed-form solution since the total cost function per pixel remains a parabola.
}

 \label{fig:Algo_preview}
\end{figure}

The main pipeline of CCA is summarized in \figref{fig:Algo_preview}. In what follows, we describe the process along with implementation details such as the confidence score modification of the parabolas and the cost aggregation over multiple scales. 

\subsection{Continuous Cost Functions}

To start with, we compute the discrete cost $C_{int}(p,d)$ for all integer disparity values $d$ and pixels $p$. Note that in the case of DP, the disparity range is relatively small. Any standard matching cost that can be smoothly interpolated (e.g., sum of absolute differences (SAD), normalized cross-correlation (NCC)) can be used at this step of CCA. Then, for every pixel, we initialize a parabola around the integer disparity that corresponds to the minimum cost value. 

If we assume that $d^0$ is the integer disparity of the minimum cost for pixel $p$,  
we fit a parabola to the costs around $d^0$ to define a \emph{continuous} cost:
\BE  \label{local_parabola_cost}
C_{p,d^0}(\Delta d) = a_p \Delta d^2 + b_p\Delta d+c_p,
\EE
with $\Delta d$ being the continuous offset from the integer disparity $d^0$. The coefficients $\{a_p, b_p, c_p\}$ can be easily found from the costs of three adjacent disparities:
\BEAN
&a_p = \frac{C_{int}(p, d^0+1) + C_{int}(p, d^0-1) -2C_{int}(p,d^0)}{2}, \\
&b_p = \frac{C_{int}(p, d^0+1) - C_{int}(p, d^0-1)}{2}, \\
&c_p = C_{int}(p,d^0).
\EEAN

By considering the total continuous disparity $d:= d^0 + \Delta d$, we can rewrite Eq.\eqref{local_parabola_cost} as:
\begin{equation*}
C_{p,d^0}(d-d^0) = a_p (d - d^0)^2 + b_p (d - d^0) + c_p.
\end{equation*}
Recall that $d^0$ is a constant offset and the quadratic cost can be defined as:
\begin{equation}  \label{global_parabola_cost}
C_p (d) =\alpha_p  d^2 +\beta_p\ d+\gamma_p,
\end{equation}
 where the new coefficients  $\{\alpha_p, \beta_p, \gamma_p\}$ are defined as:
\BEAN
\alpha_p &= &a_p, \\
\beta_p &= &b_p - 2a_pd^0, \\
\gamma_p &= &c_p + a_p(d^0)^2 - b_pd^0.
\EEAN

Although these functions are low-order polynomials and cannot represent all the richness of the discrete cost volume, they have the advantage of inherently containing a non-integer estimation of the best local disparity, as well as its confidence. The optimal local disparity corresponds to the parabola minimizer:
\BEN 
d^{optimal}_{p} = -\frac{\beta_p}{2\alpha_p},
\EEN
and the confidence is represented by the value of $\alpha_p$, which is proportional to the curvature. Intuitively, pixels with a low curvature value have a rather flat cost function, and therefore, their confidence should be low. Although the parabolas are uniquely defined by three coefficients, neither the minimizer nor the confidence are affected by the the third coefficient and its computation can be skipped.


While we adopt a quadratic function for cost interpolation, different interpolants can be likewise used. 
However, the quadratic function is convex and attains a unique minimum. These properties together with its closure under addition simplify the cost aggregation step significantly (see Sec.~\ref{sec:Agg}). In addition, any local algorithm that provides sub-pixel disparity estimation~\cite{Miclea2015New, Psarakis-ICCV2005-ecc} can be used to calculate $\Delta d$ and re-center the \emph{initial} parabola at a better position before the aggregation step, \emph{e.g.}, by keeping $\alpha_p$ fixed and updating $\beta_p$ using a better value of $d^{optimal}_{p}$.

\subsection{Confidence Score }
\label{sec:Conf}

As mentioned, a confidence measure can be deduced from the curvature of the parabolas. However, it does not take into account other local minima. Therefore, while a parabola makes a good representation of the local cost function, it can only represent a single minimum. In some cases, especially in noisy regions or those with repetitive patterns, the real disparity value might not receive the minimum matching cost value. To mitigate this, we assume 
that the correct disparity value might not always get the minimum cost value, but instead it gets a different value that is close to the minimum. Thus, while computing the parabola coefficients, we reduce the confidence using the second-lowest cost. 
If this cost is attained at disparity $d^1$ that is not adjacent to $d^0$, $|d^1-d^0|>1$, we scale the parabola by the factor 
\BE 
S_{confidence} = \max \left(\min \left(\frac{1-q}{1-T_q}, 1 \right),\epsilon \right)^2,\label{eq_confidence}
\EE
where $q = C_{int}(p,d^0)/C_{int}(p,d^1)$ is the cost ratio, and $T_{q}$ is the ratio threshold below which $S_{confidence} = 1$ (no scaling). As can be seen in Eq.\eqref{eq_confidence}, the scale factor gets lower as the second-lowest cost value approaches the lowest one. Note that such a per-pixel scaling reduces the confidence of wide parabolas without changing their minimum.



\subsection{Cost Aggregation} \label{sec:Agg}
As in SGM approach, we aggregate the costs along multiple 1D paths through the image. For a single path, our cost function $L_p(d)$ includes both a local term and a quadratic regularizer (smoothness term):
\BE \label{eq_agg_cost}
    L_p(d)=C_p(d) +  P_{adapt} (d-m_{p-1})^2,
\EE
where $m_{p-1} = \argmin_d L_{p-1}(d)$ is the minimizer of the previous parabola in the path, and $P_{adapt}$ is an adaptive scale defined below. If we substitute Eq.\eqref{global_parabola_cost} into Eq.\eqref{eq_agg_cost}, the aggregated cost can be rewritten as a new parabola with coefficients $ \{\Alpha_p,\Beta_p,\Gamma_p \}$, that attains its minimum at $-\Beta_p/(2\Alpha_p)$ and has a confidence parameter $A_p$:

\BEN
L_p(d) =  \Alpha_p d^2 + \Beta_p d + \Gamma_p,
\EEN
where
\BEAN
&\Alpha_p = \alpha_p + P_{adapt}, \\
&\Beta_p = \beta_p + P_{adapt} \cdot \frac{\Beta_{p-1}}{\Alpha_{p-1}}, \\
&\Gamma_p = \gamma_p + P_{adapt} \cdot \left(\frac{\Beta_{p-1}}{ 2 \Alpha_{p-1}} \right)^2
\EEAN
and $m_{p-1} = -\Beta_{p-1}/(2\Alpha_{p-1})$. The adaptive weight of the smoothness  term in Eq.\eqref{eq_agg_cost} is defined by $P_{adapt} =  P \cdot \Alpha_{p-1} \cdot e^{\left(-(I_p - I_{p-1})^2/\sigma^2\right)}$, which includes the user-defined parameters $P$ and $\sigma$, the confidence of the previous aggregated parabola $\Alpha_{p-1}$, and a gradient based exponential factor - similar to the one used in ~\cite{Hirschmuller-PAMI2008-sgm} - that becomes smaller over edges of the input image $I$.

Recall that the constant term $\Gamma_p$ does not affect the minimizer, thus only A and B coefficients really need to be updated within a very simple propagation equation. 

Finally, to define the total cost for pixel $p$, we sum the aggregated costs of all the paths $r$ that end in pixel $p$.
\BEN
S_p(d) = \sum_r L_p(d),
\EEN
with the sum of parabolas $S_p(d)$ being in fact a parabola itself. Therefore the final continuous disparity value for pixel $p$ is obtained by the minimizer of $S_p(d)$:
\BE \label{eq:disp}
disparity =  - \frac{\sum_r \Beta_p}{2\sum_r \Alpha_p}.
\EE

As seen in Eq.~\equref{eq:disp}, an instability may occur if the denominator approaches $0$. Although the chances of this happening is low due to the summation over different paths, it is still important to mitigate this issue. 
To that end, for every pixel, we invalidate the parabola if the coefficient $\alpha_p$ is below a small threshold $T_a$: 
\[
    \{\alpha_p, \beta_p, \gamma_p\} = 
\begin{cases}
    \{\epsilon, 0, 0\},& \text{if } \alpha_p<T_a\\
    \{\alpha_p, \beta_p, \gamma_p\}, & \text{otherwise}.
\end{cases}
\]
Using $\epsilon$ and $0$ values means that the aggregated parabola will receive its minimum from the previous pixels in the path.

It is important to clarify here that the parabola aggregation can significantly change the initial integer disparity around which the parabola is initialized. As such, CCA should not be seen as a sub-pixel refinement of the initial integer disparity.

As discussed, the cost aggregation is performed along straight paths, which may result in artifacts known as streaks
~\cite{facciolo2015mgm}. To alleviate this issue, one may optionally apply several iterations of aggregation to make the smoothness requirement stronger. 
An iterative cost aggregation step is discussed in the appendix.

\subsection{Cost Aggregation Over Scales}
A multi-scale approach is beneficial to stabilizing the disparity estimation, as well as to the potential acceleration due to the reduced number of pixels and the shrinkage of the disparity range. In the specific case of DP cameras, a multi-scale approach is also beneficial to resolve depth in the presence of blur that may destabilize the disparity in fine levels of the pyramid. Given a continuous disparity result at a coarse scale $s$, the question is how to use it to compute the disparity at finer scale $s-1$. 
A simple solution is to upscale the coarse-level disparity and compute the nearest integer disparity at the finer level. 
Then, the fine-level parabolas can fit around the up-scaled values. Although this approach is very efficient, errors in coarse levels seep through to finer scales without any way to correct them. 

Instead, we use a less strict approach where the fine scale solution is more affected by coarse scale solution in less confident regions and has only a minor effect in high-confidence regions. For this, we bring the aggregated parabolas 
$\{\Alpha_{p,s},\Beta_{p,s}\}$
from the coarse scale $s$ to the finer scale $s-1$ by upscaling and updating the coefficient maps of scale $s$ to match the map size of scale $s-1$, using bilinear interpolation: 
\BEAN
&\Alpha_{p,s-1}^{prior} = upsample\left(\Alpha_{p,s} \right), \\
&\Beta_{p,s-1}^{prior} = upsample\left(\Beta_{p,s} \cdot F\right),
\EEAN
where  $F$ is the scale factor between scale $s-1$ and $s$. Finally, these maps are weighted by a factor $w$ and added to the per-pixel coefficient maps of scale $s-1$, in order to calculate the coefficients of Eq.~\eqref{global_parabola_cost}:
\BEAN
&\alpha_{p,s}^{with \; prior} = \alpha_{p,s} + w \cdot \Alpha_{p,s-1}^{prior}, \\
&\beta_{p,s}^{with \; prior} = \beta_{p,s} + w \cdot \Beta_{p,s-1}^{prior}.
\EEAN
Such a weighting has the potential to shift and stabilize “weak” parabolas in weakly textured or blurred regions.

Interestingly, any kind of prior might be added in a similar way. For instance, it is possible to aggregate the parabolas over time where $w$ can be controlled by the similarity of the current pixel to the corresponding pixel from a previous frame.
In order to accelerate the computation, our multi-scale solution works by finding the range $[d_{MIN},d_{MAX}]$ of the expected disparity of the current scale according to the previous scale. Then, the matching costs for the current scale is computed only for the expected integer disparity values in the range $[d_{MIN}-1,d_{MAX}+1]$. This approach assumes that the globally detected disparity range is a good representation of the possible disparity range, while still avoiding making early decisions at pixel level.



\subsection{Complexity Analysis}\label{sec:complexity_analysis}
In order to derive the complexity of CCA, we consider a single-scale of disparity computation. Computing the integer disparity corresponding to the lowest cost and then fitting the initial parabolas requires $O(WHD)$ operations, where $W$ and $H$ are the width and height of the image and $D$ is the number of disparity levels. In the cost aggregation step, we need to propagate two coefficients over the entire image and accumulate the results, which requires $O(WH)$ operations for every path. Therefore, assuming the number of paths $R$, the total complexity of the aggregation is $O(WHR)$. This is in contrast to SGM where the cost aggregation complexity is $O(WHDR)$, which is considerably higher. The final computation of the optimal disparity can be done as part of the last path computation and therefore additional computational complexity is not added. Therefore, the time complexity of CCA is $O(WHD+WHR)$. Since CCA does not need to revisit the entire cost volume and maintains only two scalars per pixel, the space complexity does not depend on the number of disparities and is $O(WH)$.




\section{Experiments}

To evaluate the performance of the proposed algorithm and compare it to SOTA in DP disparity estimation, we conduct experiments using DP data from both DSLR and phone cameras. In addition, we compare the performance of CCA and SGM on standard stereo data-sets.

\subsection{Dual Pixel Cameras}
\label{sec:experiments DP}

Openly accessible DP data exists for Canon DSLR cameras~\cite{Punnappurath-ICCP2020-modelingDefocus} and Google Pixel phones~\cite{Garg-ICCV2019-learningDual}, with available data-sets for both cameras. As DP sensors capture both views in a single shot, the two captured images share the same exposure time and gain, which should result in images with similar intensities. However, while DSLR left-right DP images have similar intensity levels, left-right sub-pixels of phone cameras have different properties (i.e. due to a lower lens/sensor quality). This leads us to use different pre-processing steps in order to compute the disparity maps. In the next sections, we evaluate our algorithm on these two different data-sets and compare to SOTA algorithms. The hyper-parameters used for the experiments below and additional experimental results with different parameter configurations such as number of iterations are provided in the appendix. 



\subsubsection{DSLR Captured Images}

In DSLR cameras the two resulting DP images are color images as each pixel on the sensor is divided into two adjacent pixels. In addition, the high quality sensor and lens system result in images with similar intensity levels and few aberrations. The large aperture results in images with rather large disparities between images. The good quality of images means that additional pre-processing or robust cost function is not needed. We calculate costs using the SAD and we shift the initial parabola using the subpixel estimation of ENCC~\cite{Psarakis-ICCV2005-ecc} since it better initializes the aggregation step (see supplementary material). As out-of-focus regions cause an increase in both disparity and blur of the image, we used three levels in multi-scale aggregation. In all the experiments, we evaluate the performance of our algorithm with and without post-processing filter, by adopting the filtering pipeline of~\cite{Punnappurath-ICCP2020-modelingDefocus}. 

\begin{figure}[t]
    \setlength\abovecaptionskip{-0.6\baselineskip} 
    \setlength\belowcaptionskip{-15pt} 
  \begin{center}
   \begin{tabular}{c@{~}c@{~}c@{~}c}
       {\raisebox{0.6cm}{\rotatebox[origin=c]{90}{~\scriptsize{(a)Image}}}}&
       \includegraphics[width=0.3\linewidth]{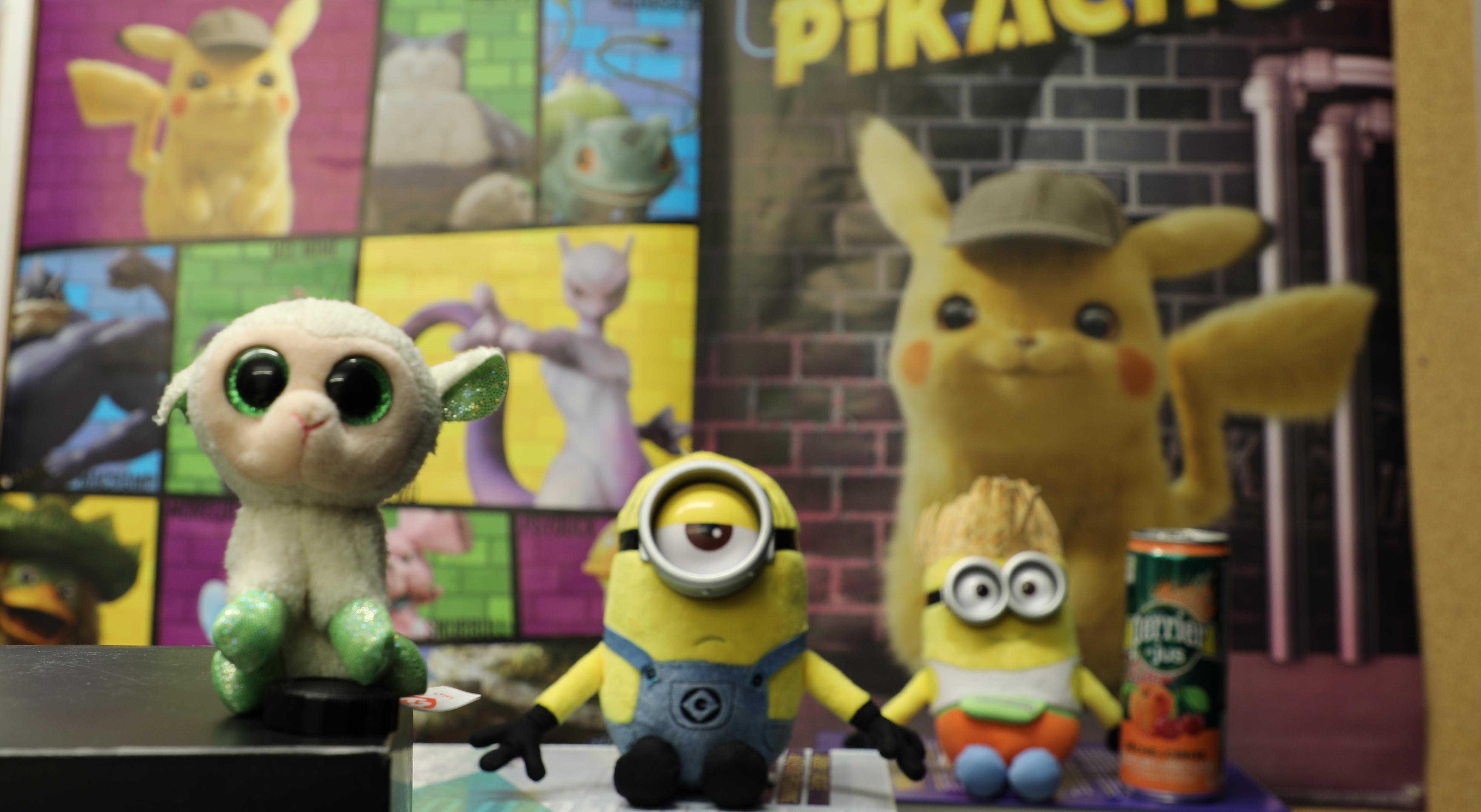} &
      \includegraphics[width=0.3\linewidth]{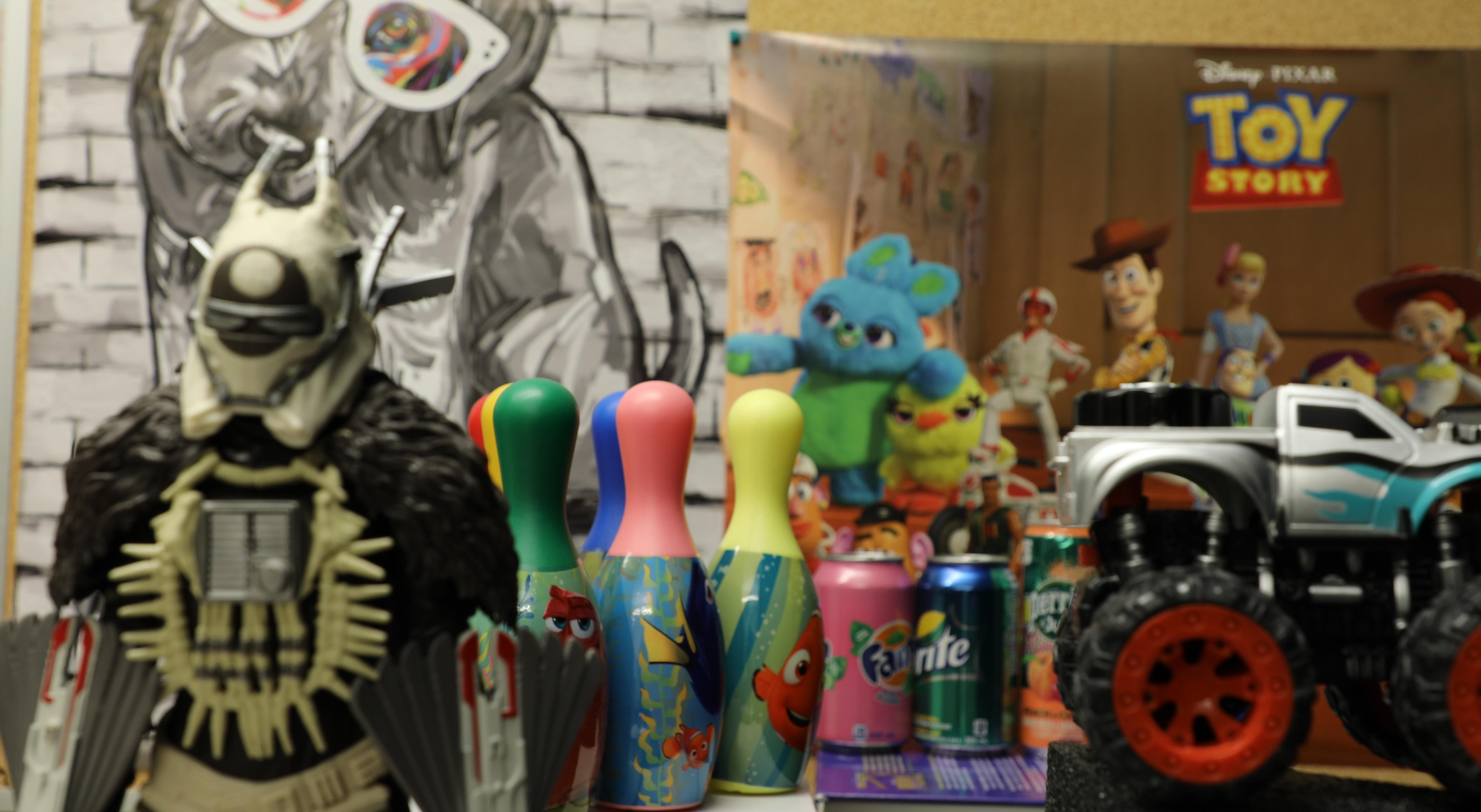} &
       \includegraphics[width=0.3\linewidth]{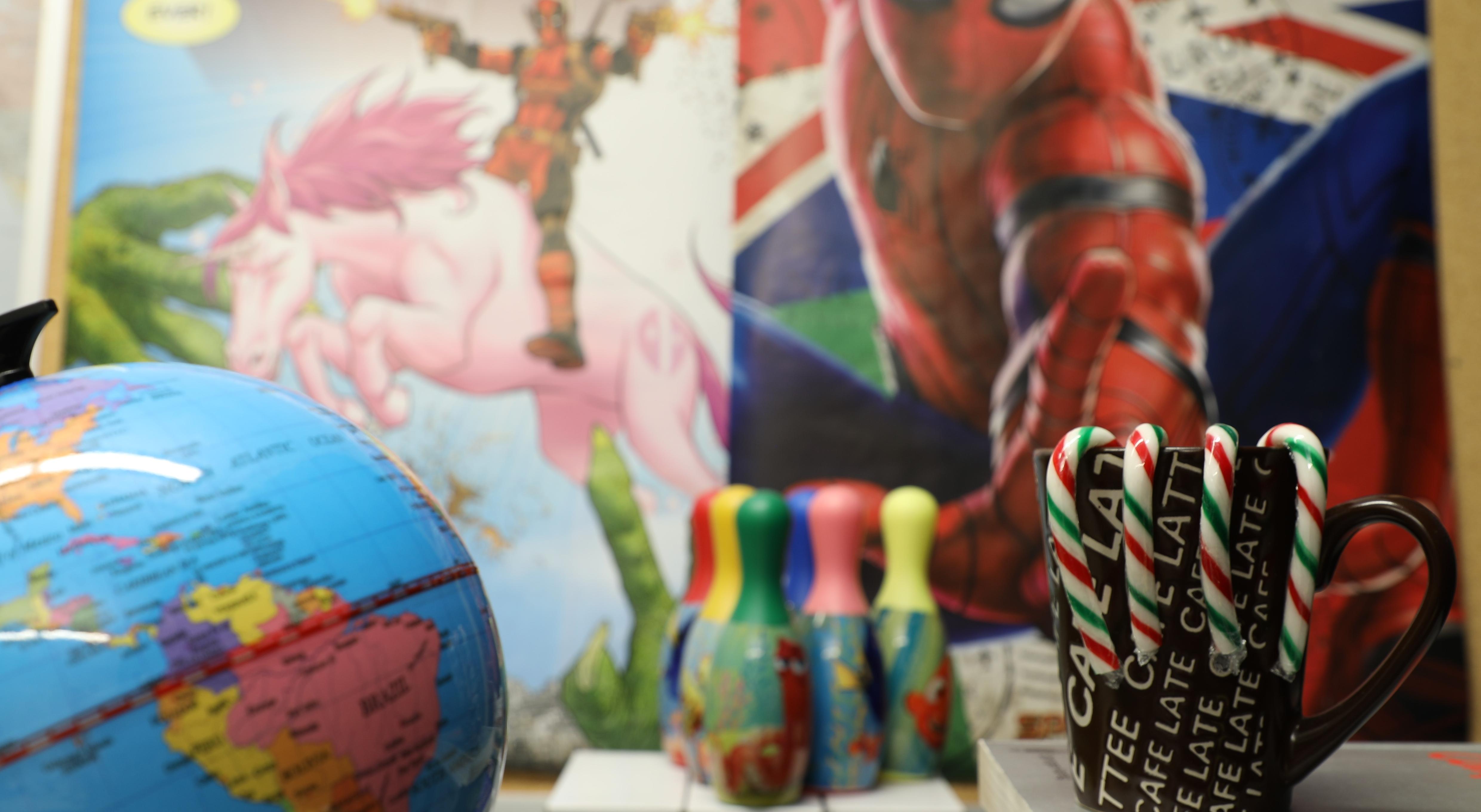} \\

       {\raisebox{0.6cm}{\rotatebox[origin=c]{90}{~\scriptsize{(b)GT}}}}&
       \includegraphics[width=0.3\linewidth]{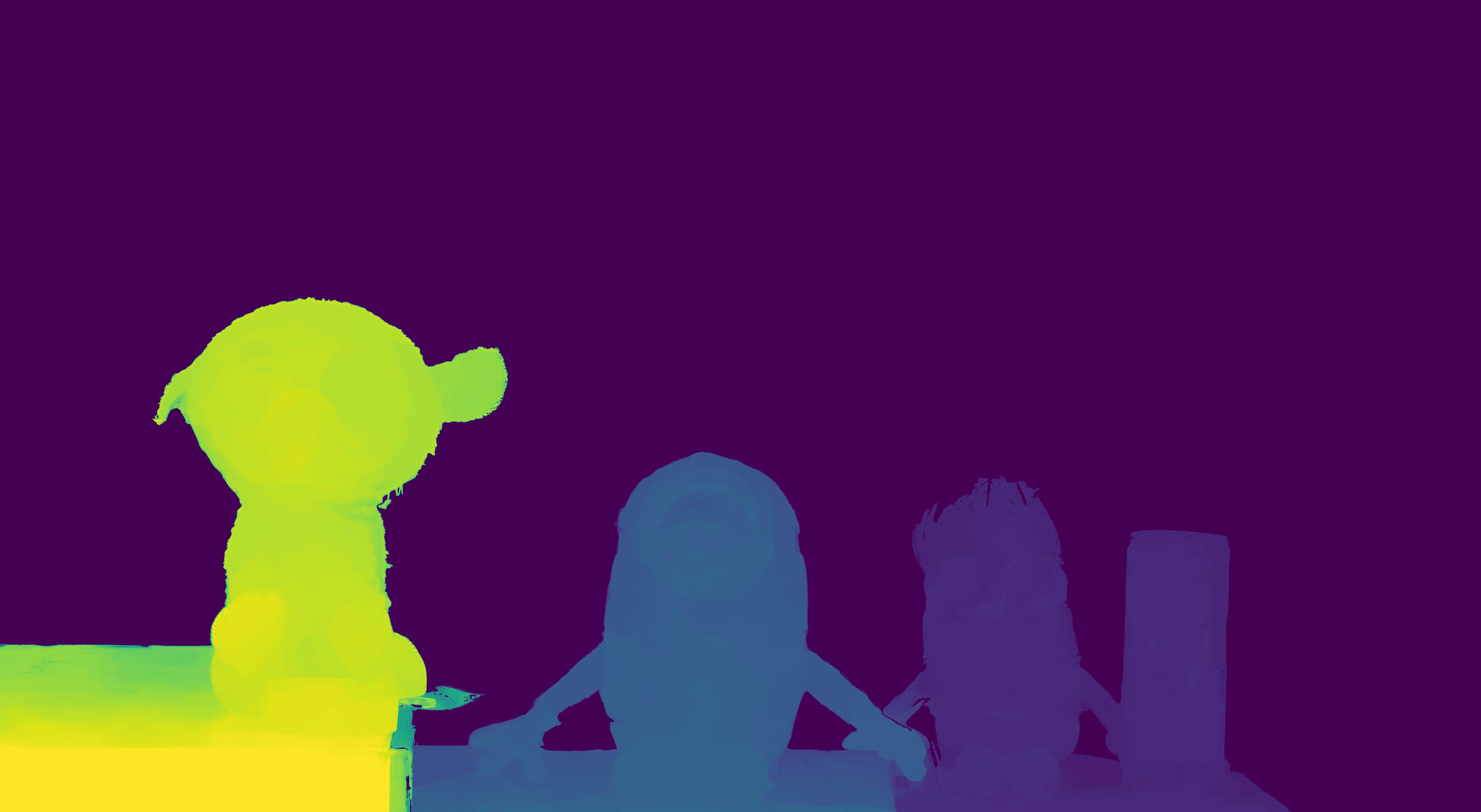} &
      \includegraphics[width=0.3\linewidth]{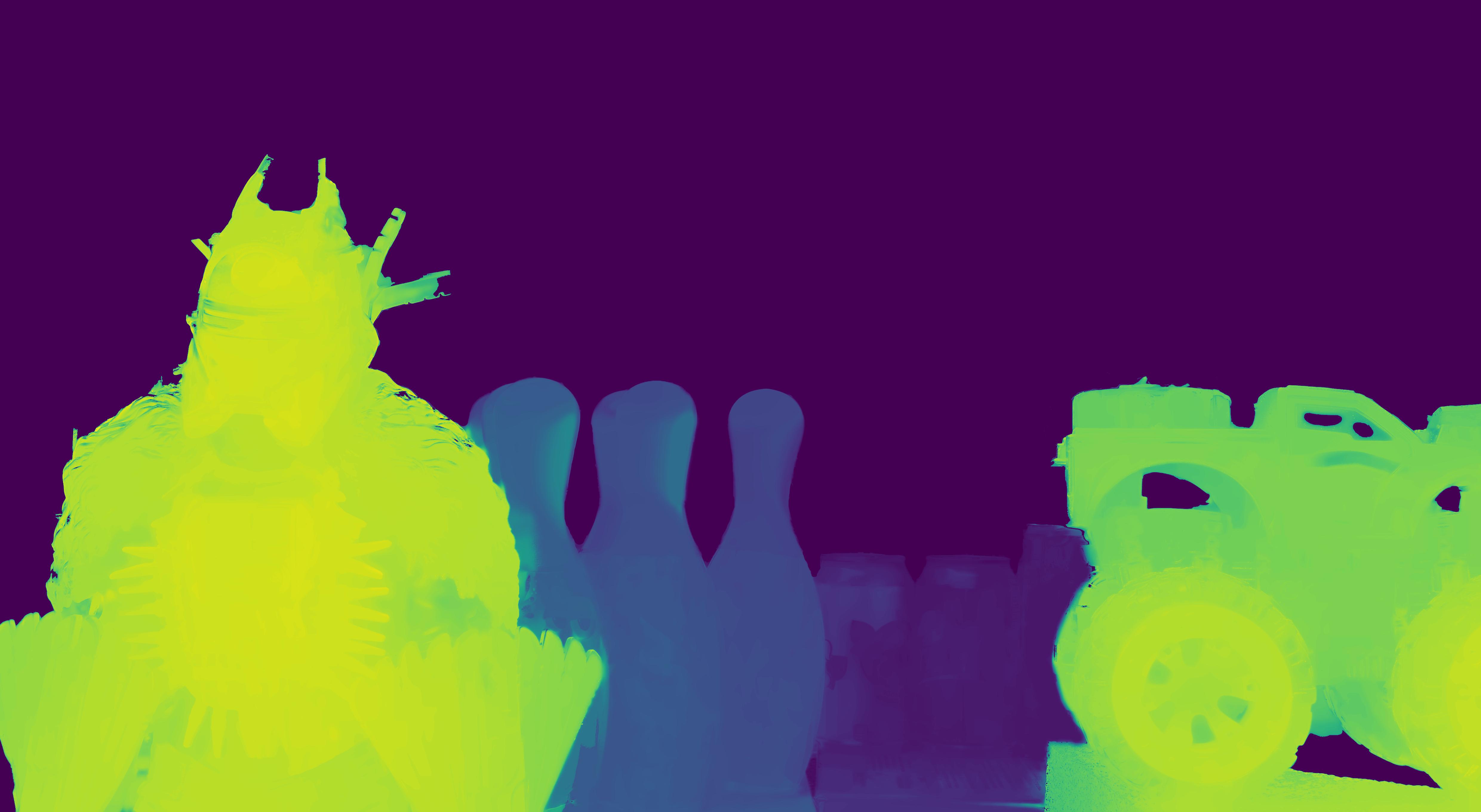} &
       \includegraphics[width=0.3\linewidth]{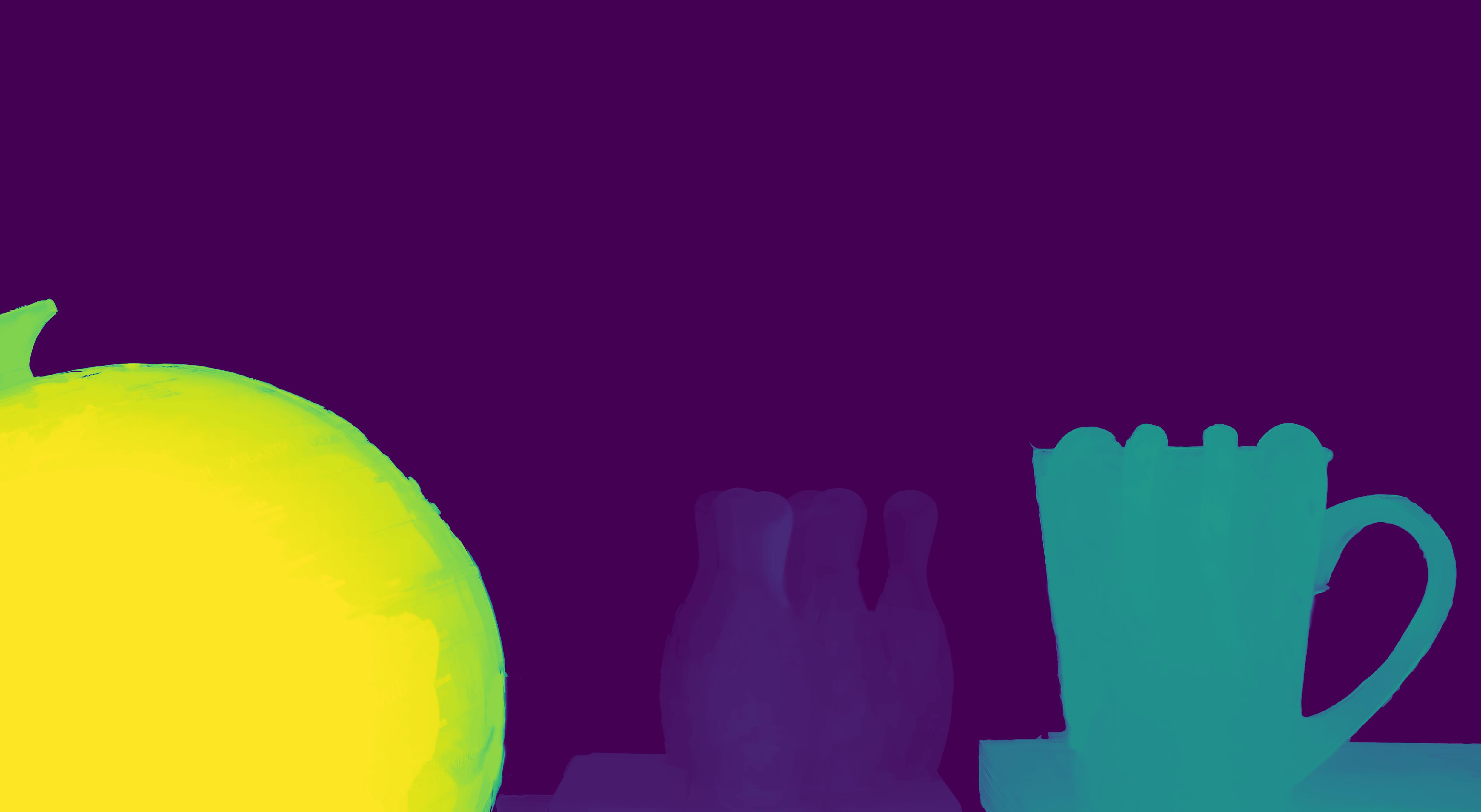} \\
       

       {\raisebox{0.6cm}{\rotatebox[origin=c]{90}{~\scriptsize{(c)SDoF~\cite{Wadhwa_SIGGRAPH2018_syntheticDoF}}}}}&
       \includegraphics[width=0.3\linewidth]{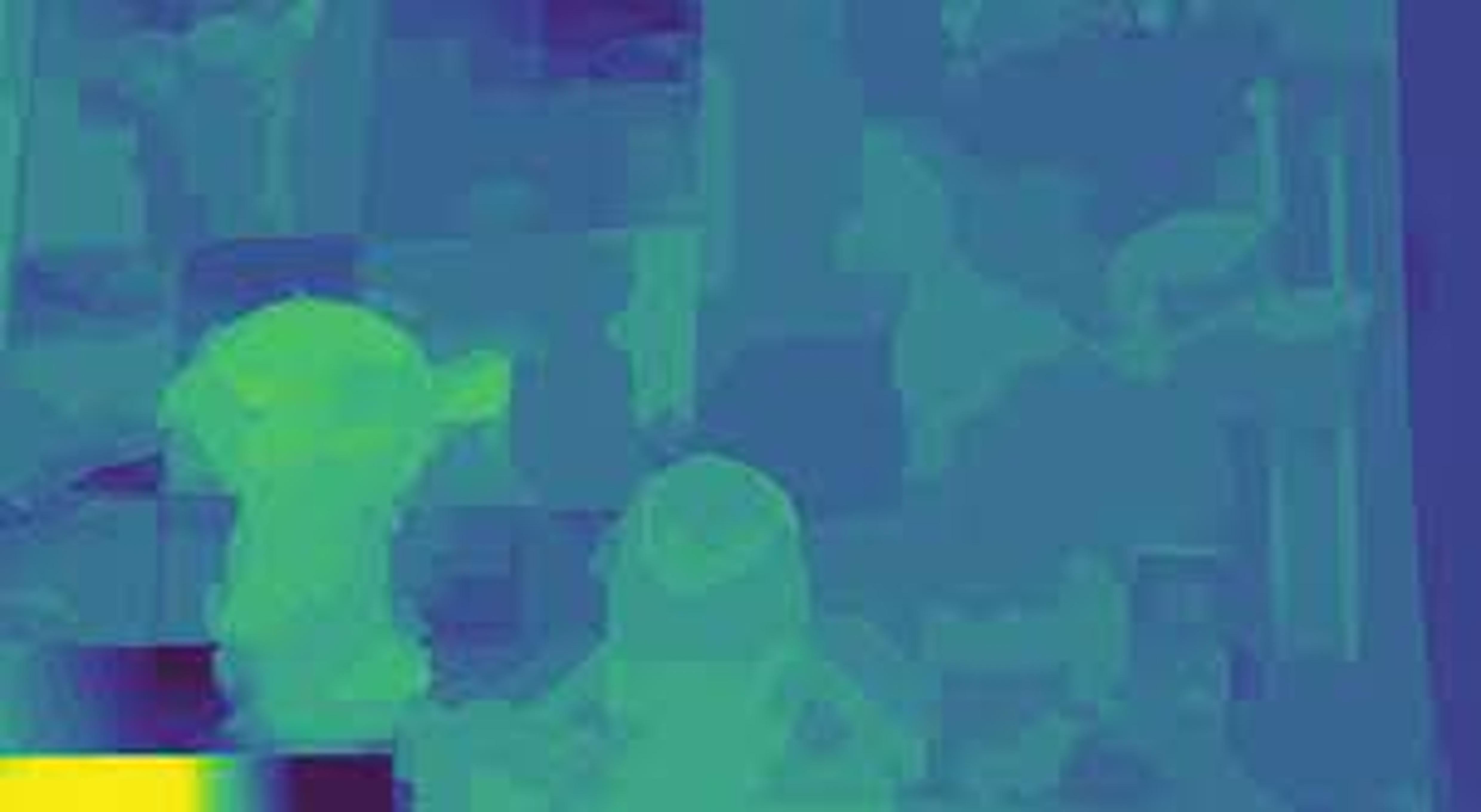} &
      \includegraphics[width=0.3\linewidth]{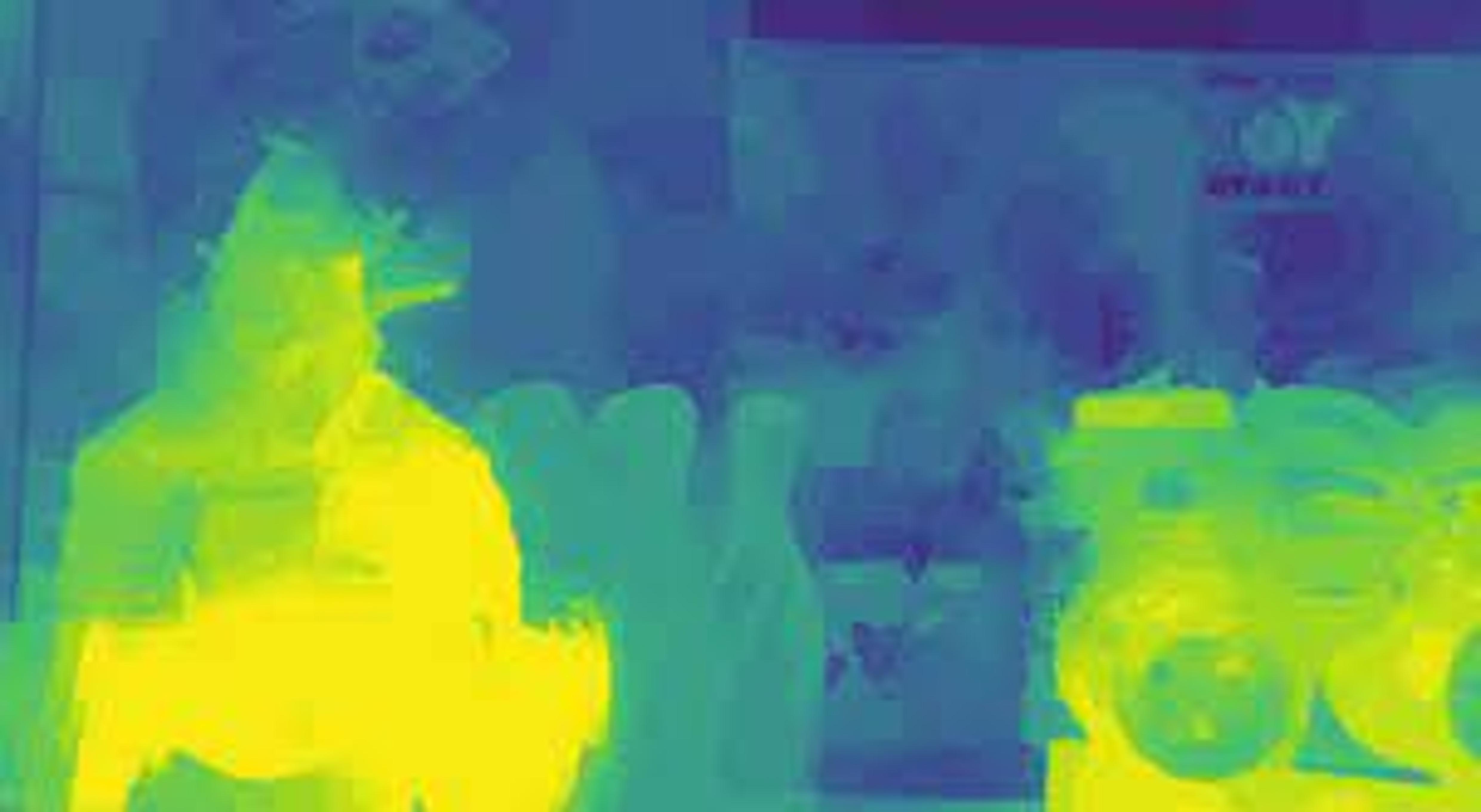} &
       \includegraphics[width=0.3\linewidth]{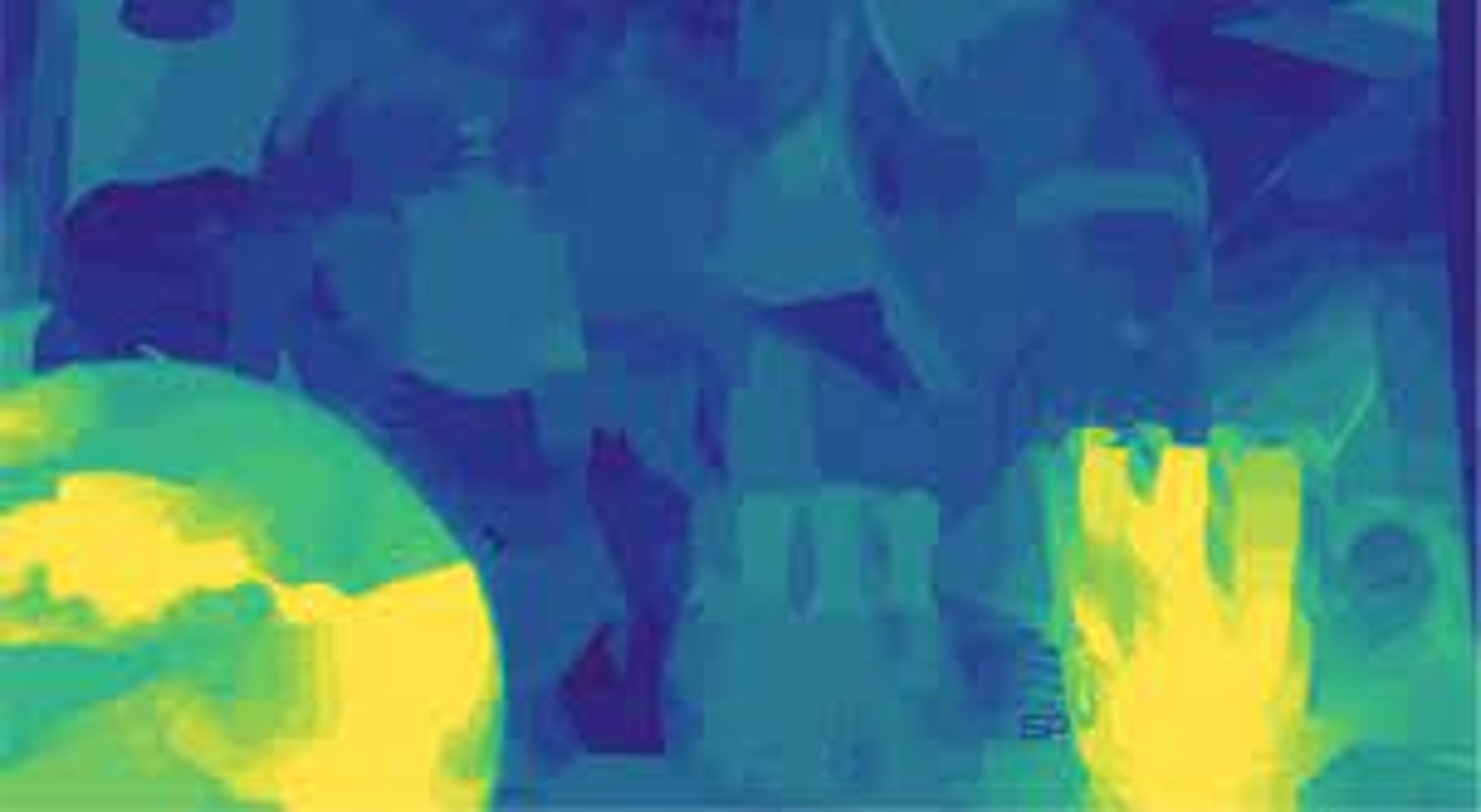} \\
      
       {\raisebox{0.5cm}{\rotatebox[origin=c]{90}{~\scriptsize{(d)DPdisp~\cite{Punnappurath-ICCP2020-modelingDefocus}}}}}&
       \includegraphics[width=0.3\linewidth]{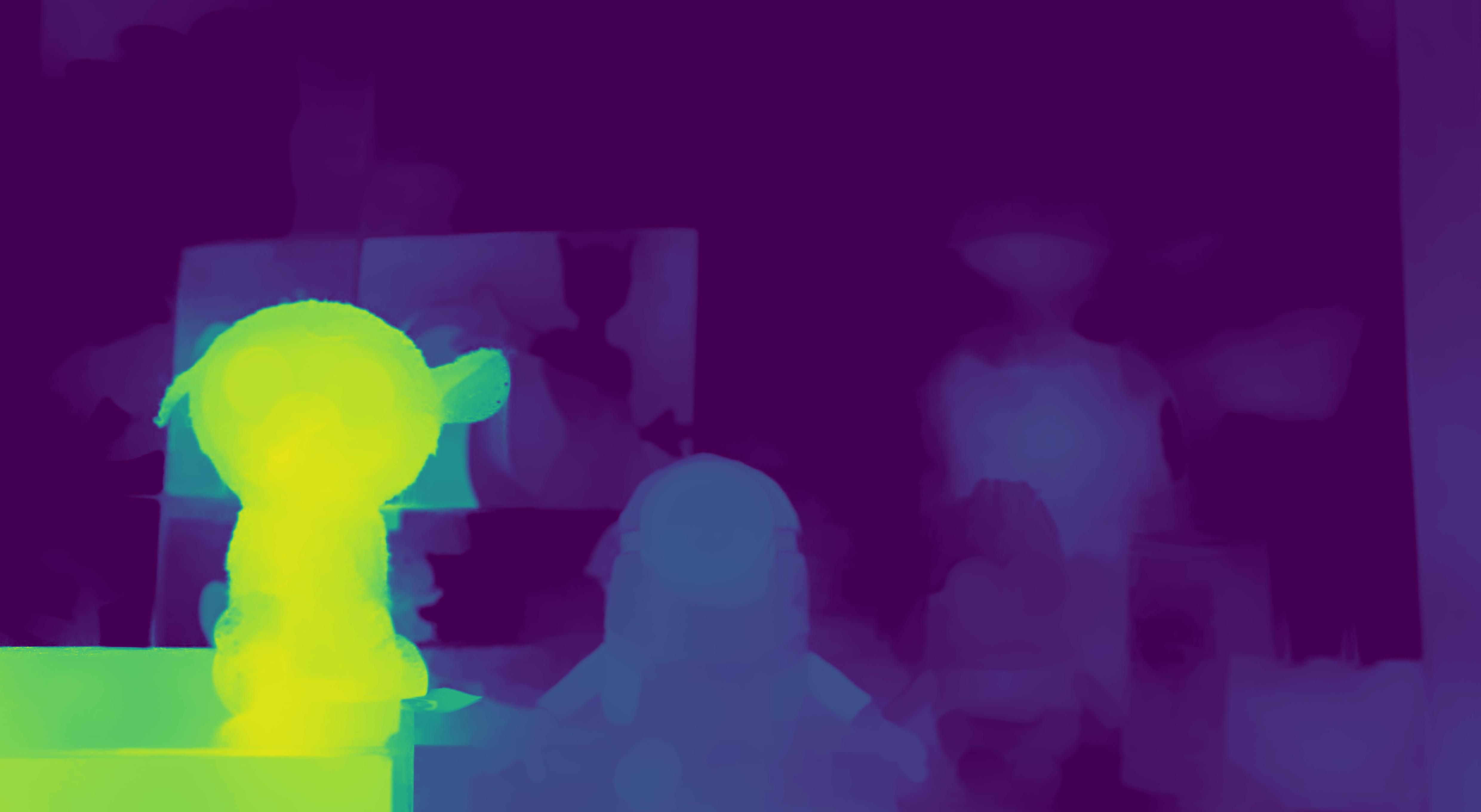} &
      \includegraphics[width=0.3\linewidth]{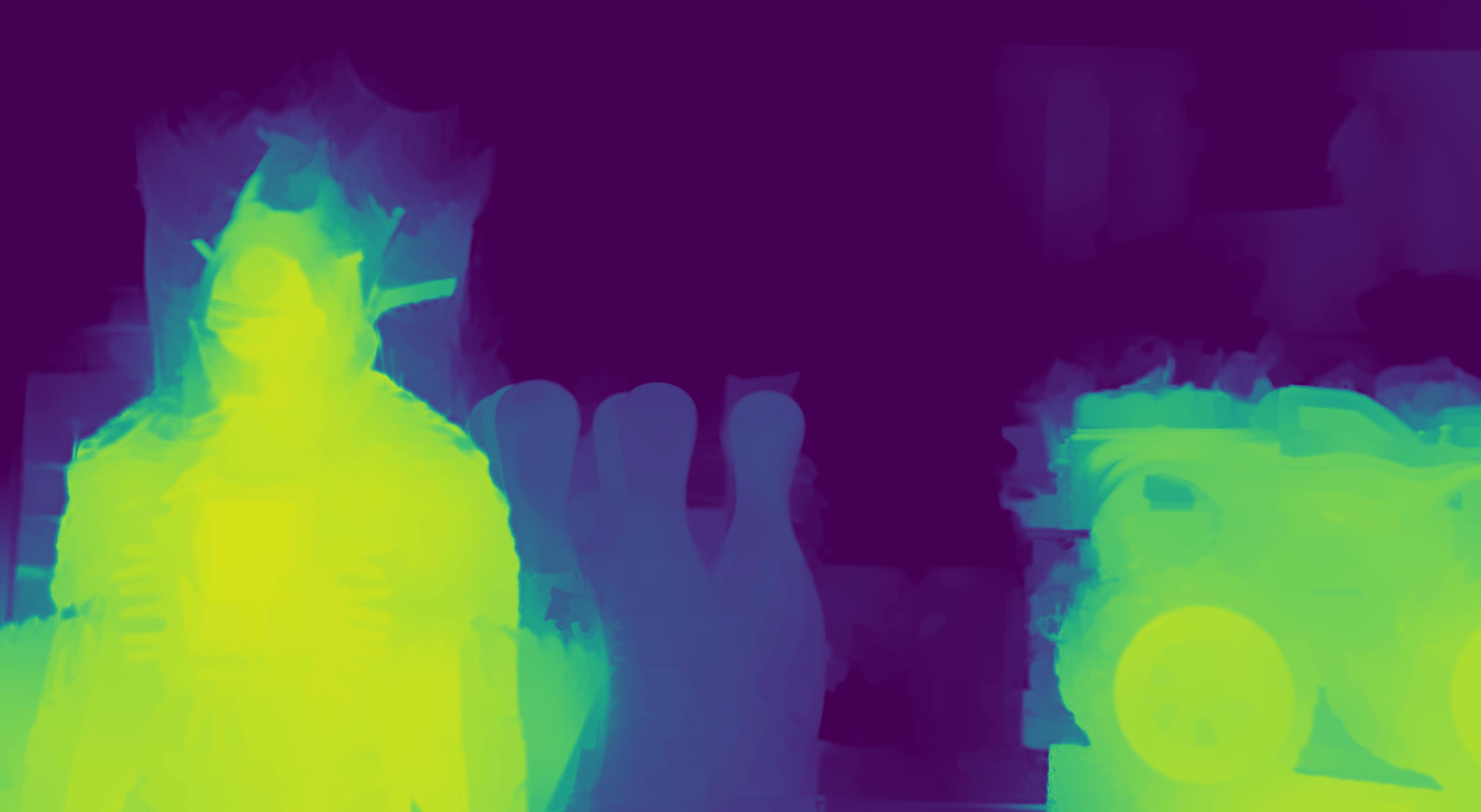} &
       \includegraphics[width=0.3\linewidth]{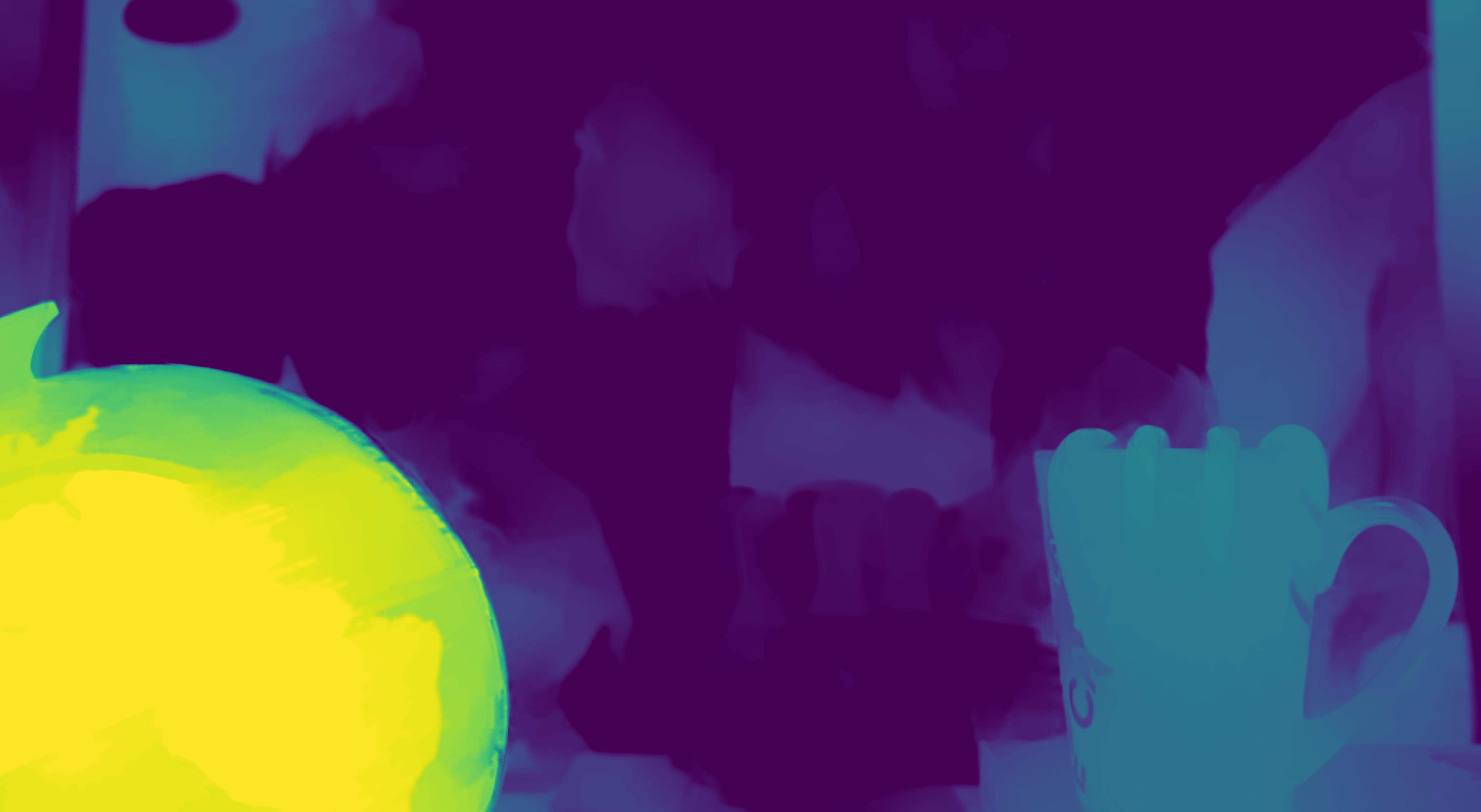}\\

       {\raisebox{0.6cm}{\rotatebox[origin=c]{90}{~\scriptsize{(e)DPE~\cite{Pan_CVPR2021_dpexploration}}}}}&
       \includegraphics[width=0.3\linewidth]{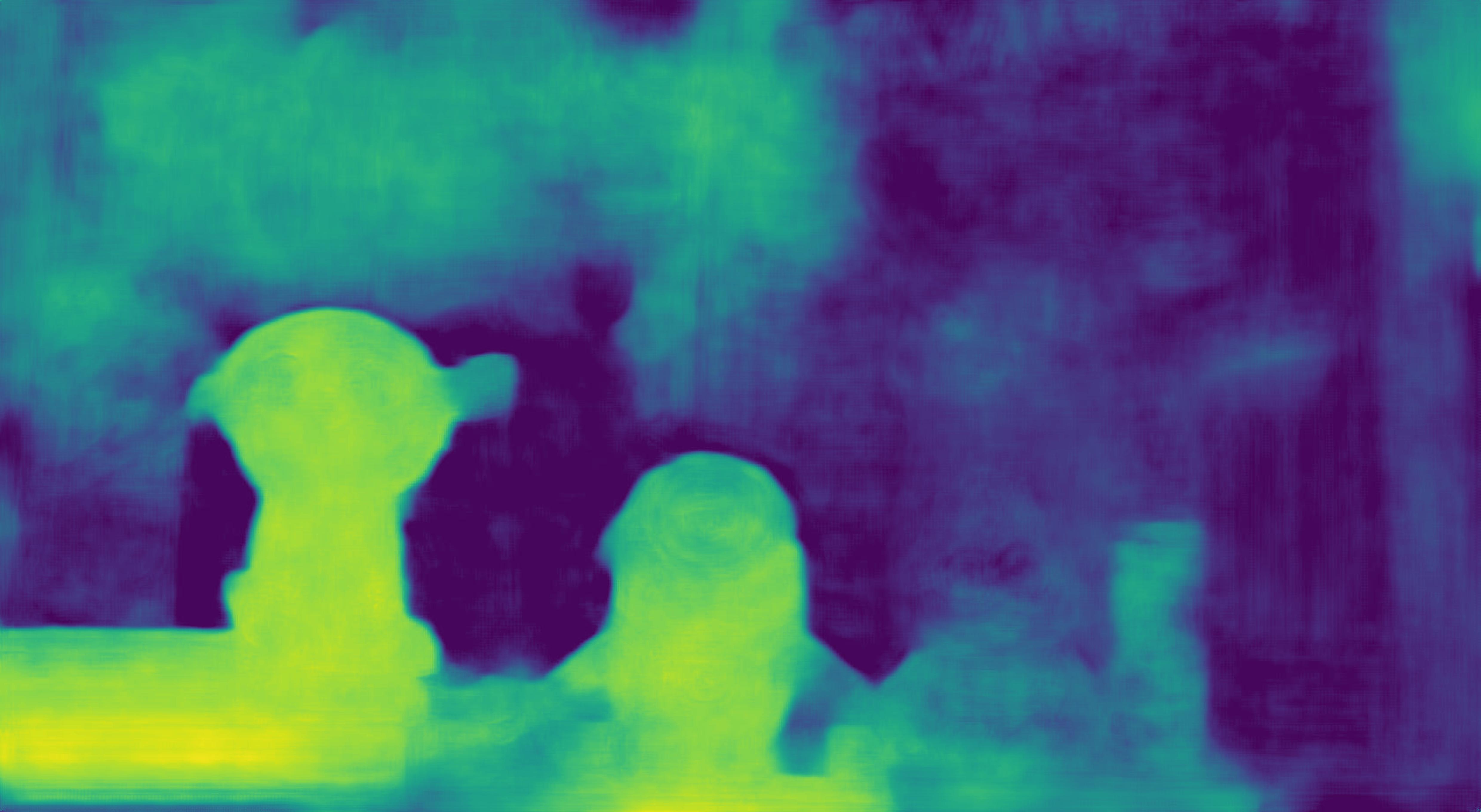} &
      \includegraphics[width=0.3\linewidth]{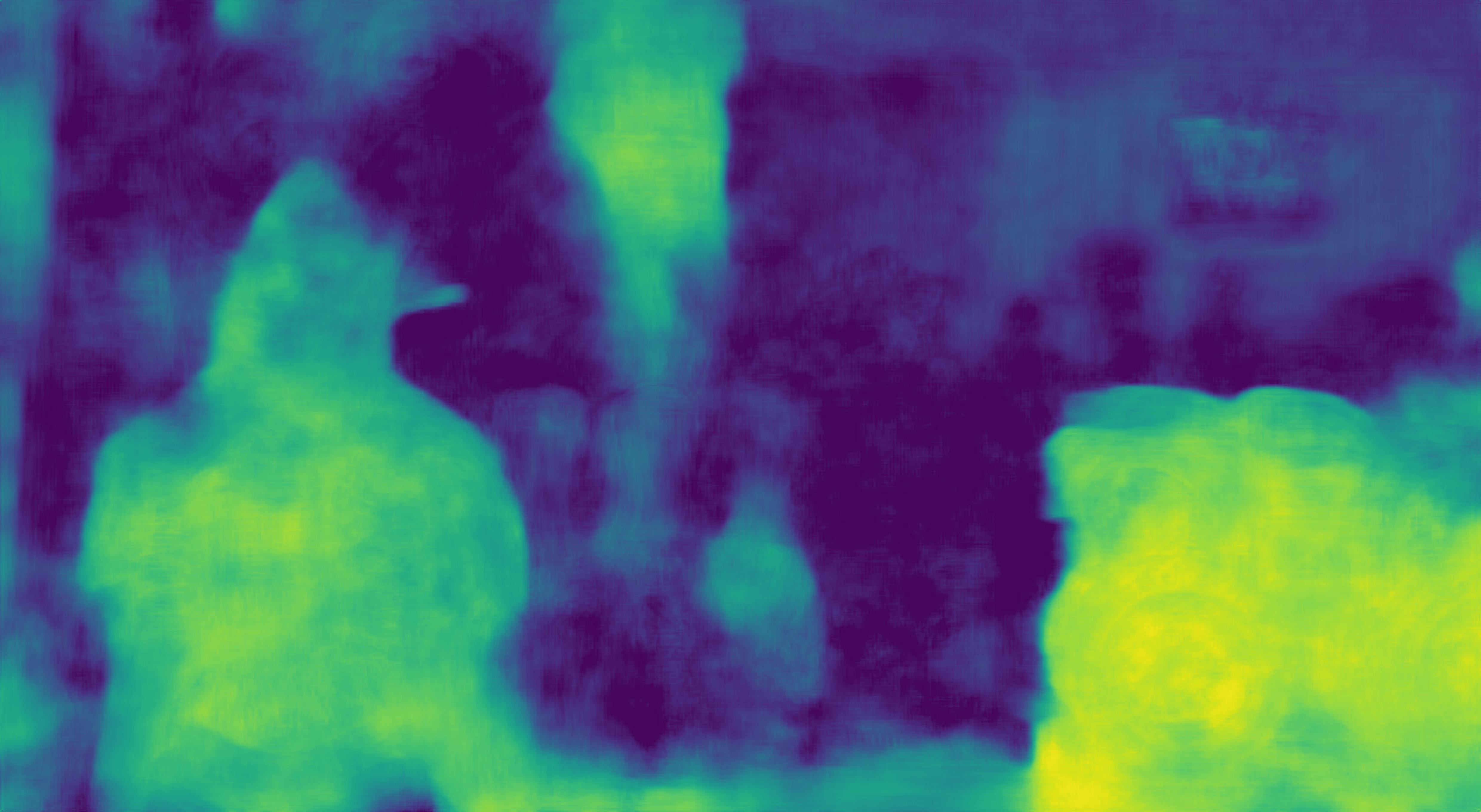} &
       \includegraphics[width=0.3\linewidth]{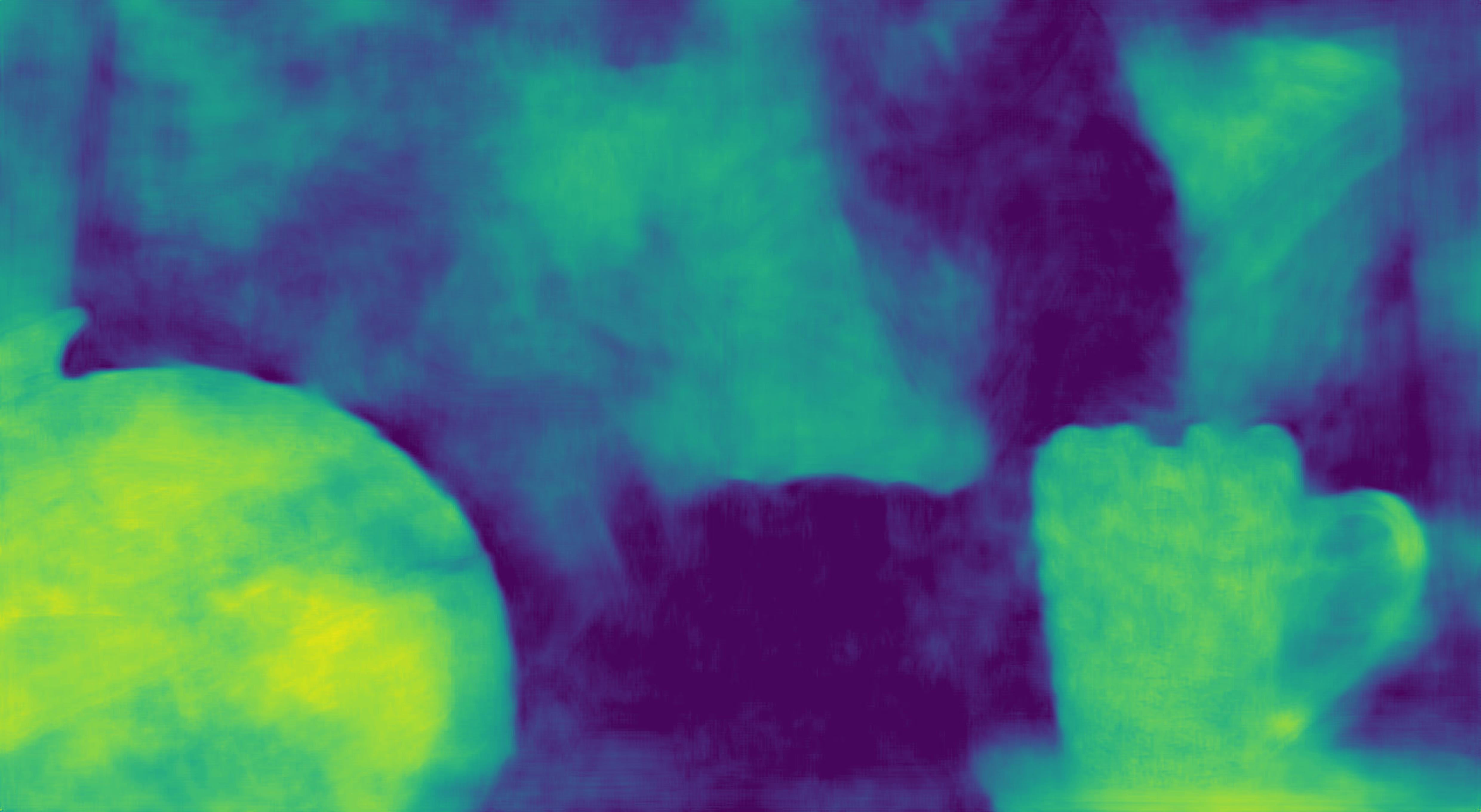} \\

       {\raisebox{0.5cm}{\rotatebox[origin=c]{90}{~\scriptsize{(f)CCA}}}}&
       \includegraphics[width=0.3\linewidth]{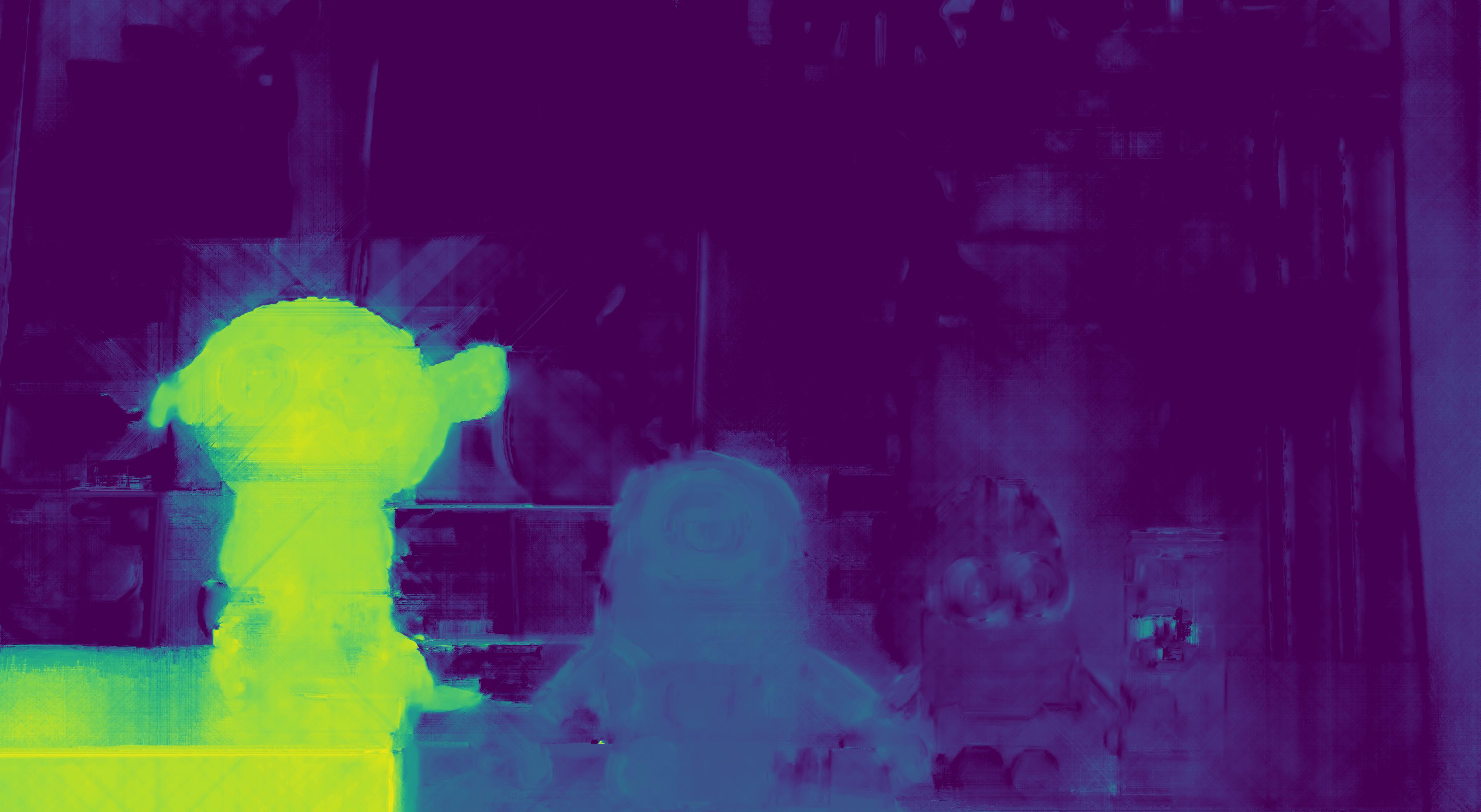} &
      \includegraphics[width=0.3\linewidth]{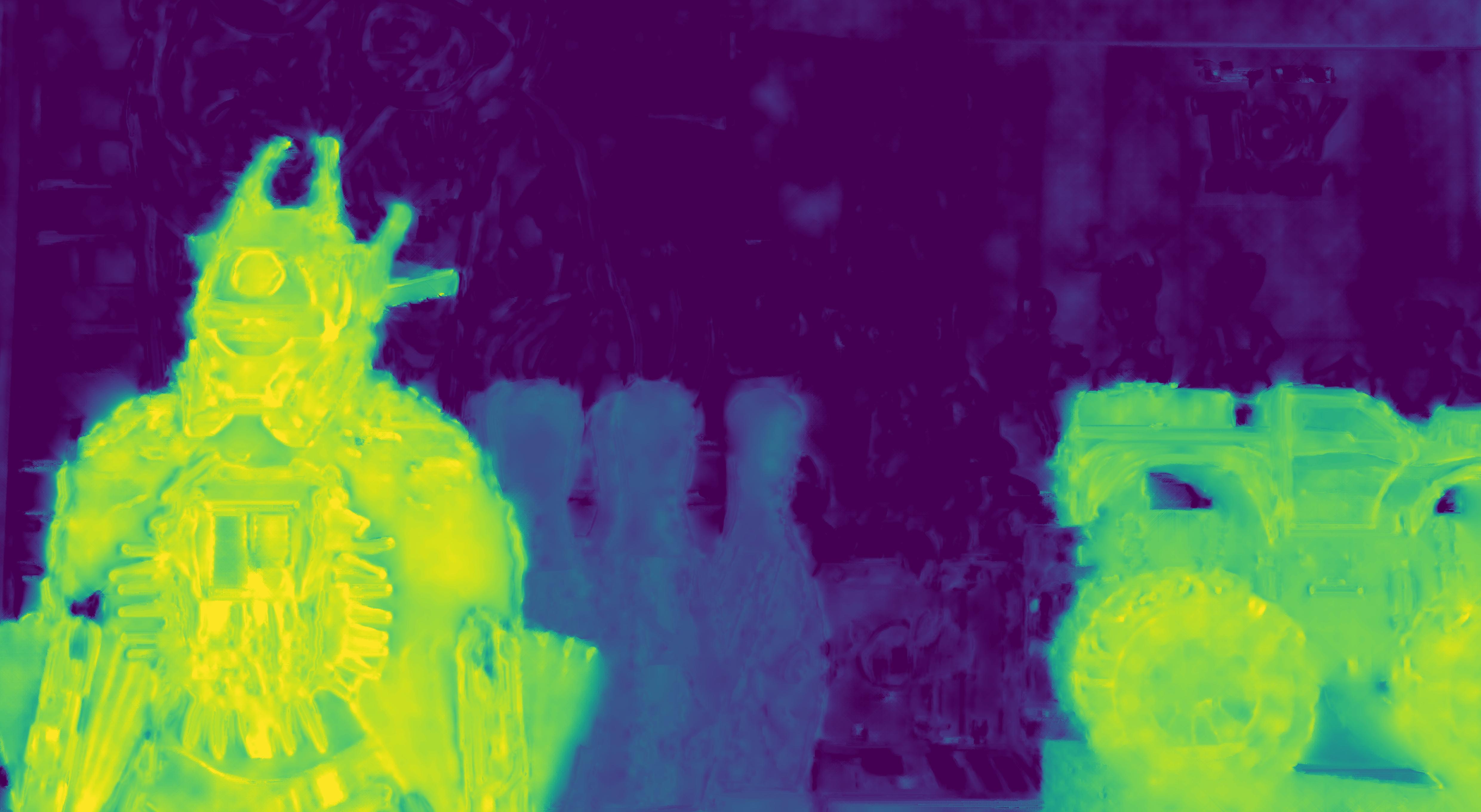} &
       \includegraphics[width=0.3\linewidth]{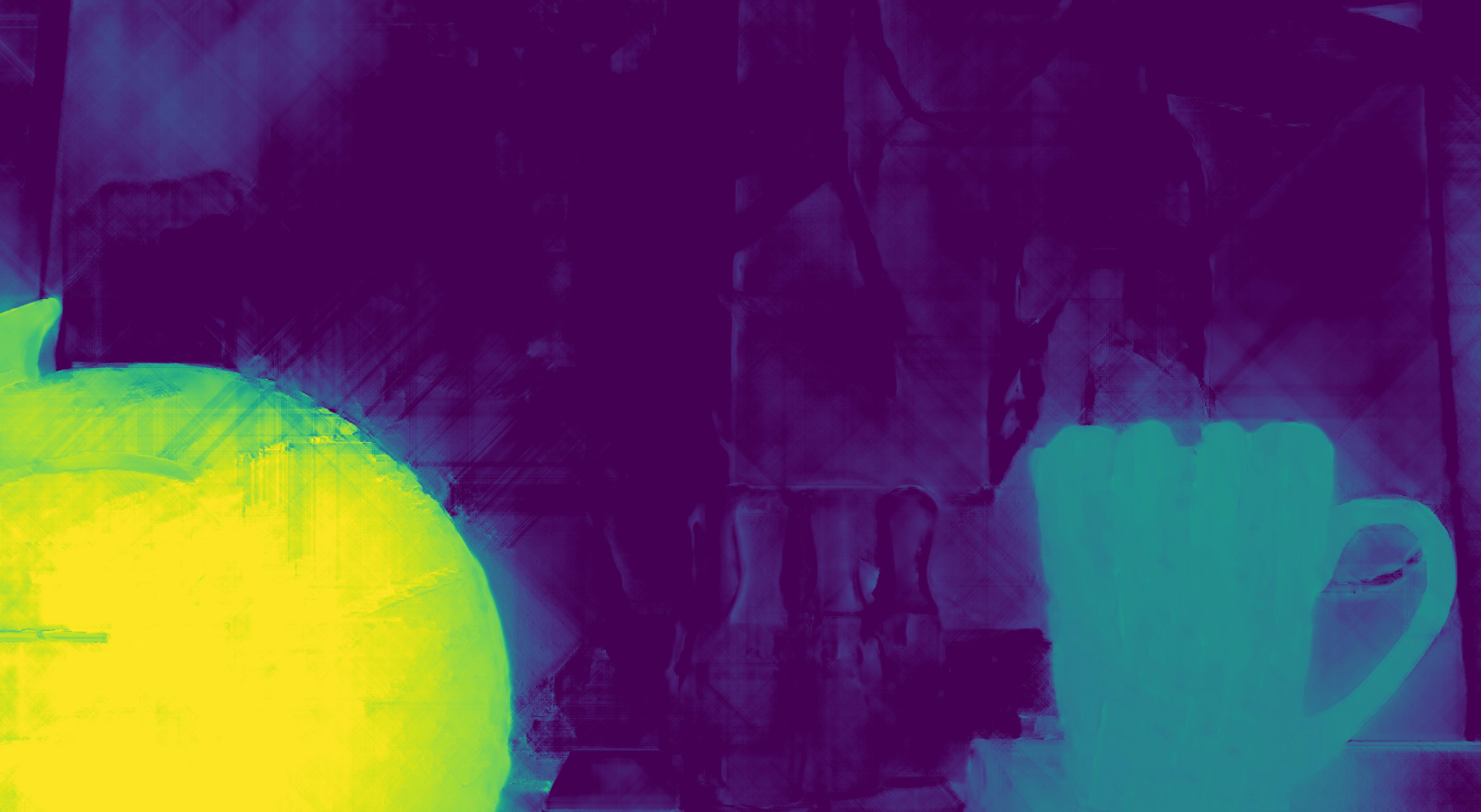} \\

       {\raisebox{0.4cm}{\rotatebox[origin=c]{90}{~\scriptsize{(g)CCA $+$ filter}}}}&
       \includegraphics[width=0.3\linewidth]{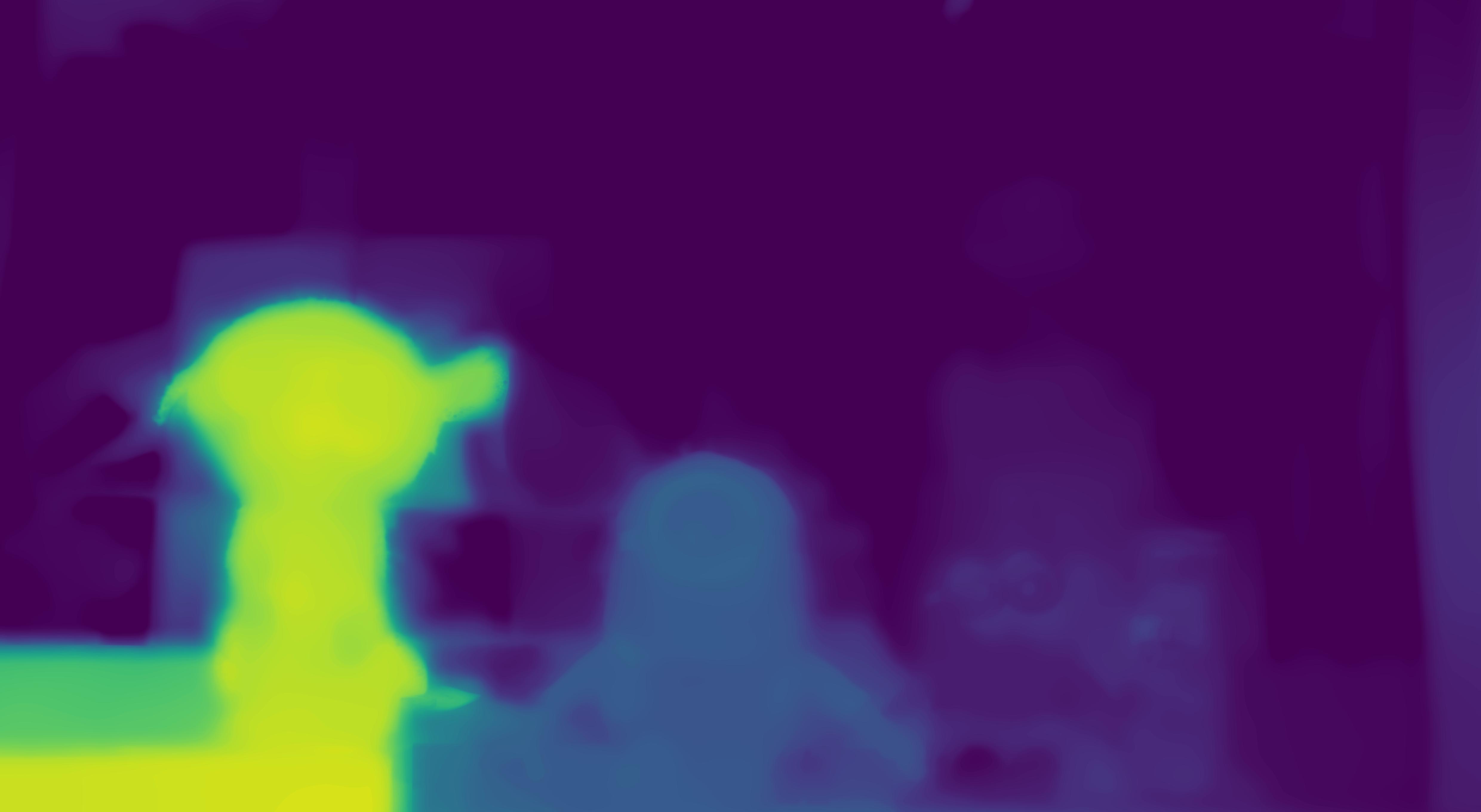} &
      \includegraphics[width=0.3\linewidth]{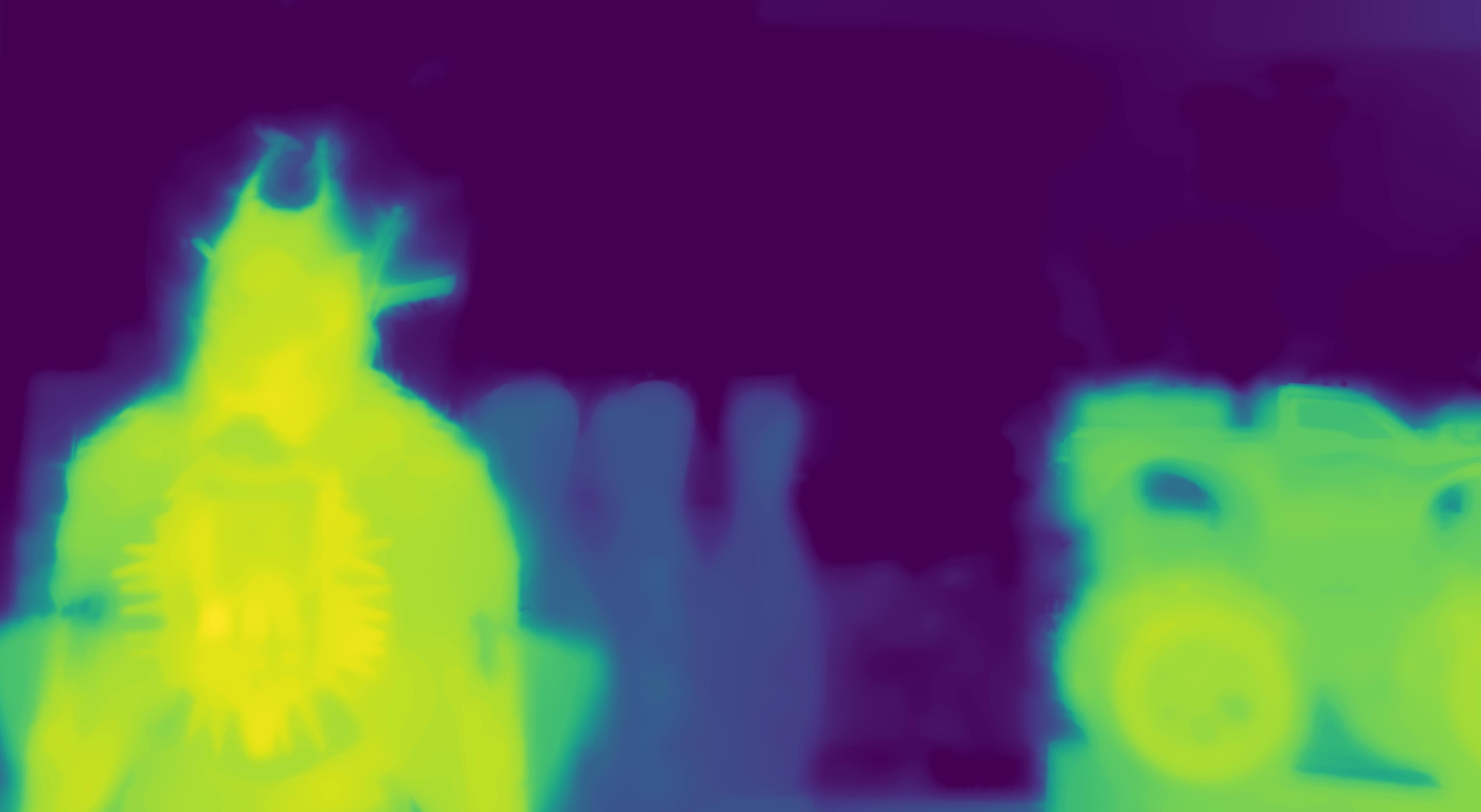} &
       \includegraphics[width=0.3\linewidth]{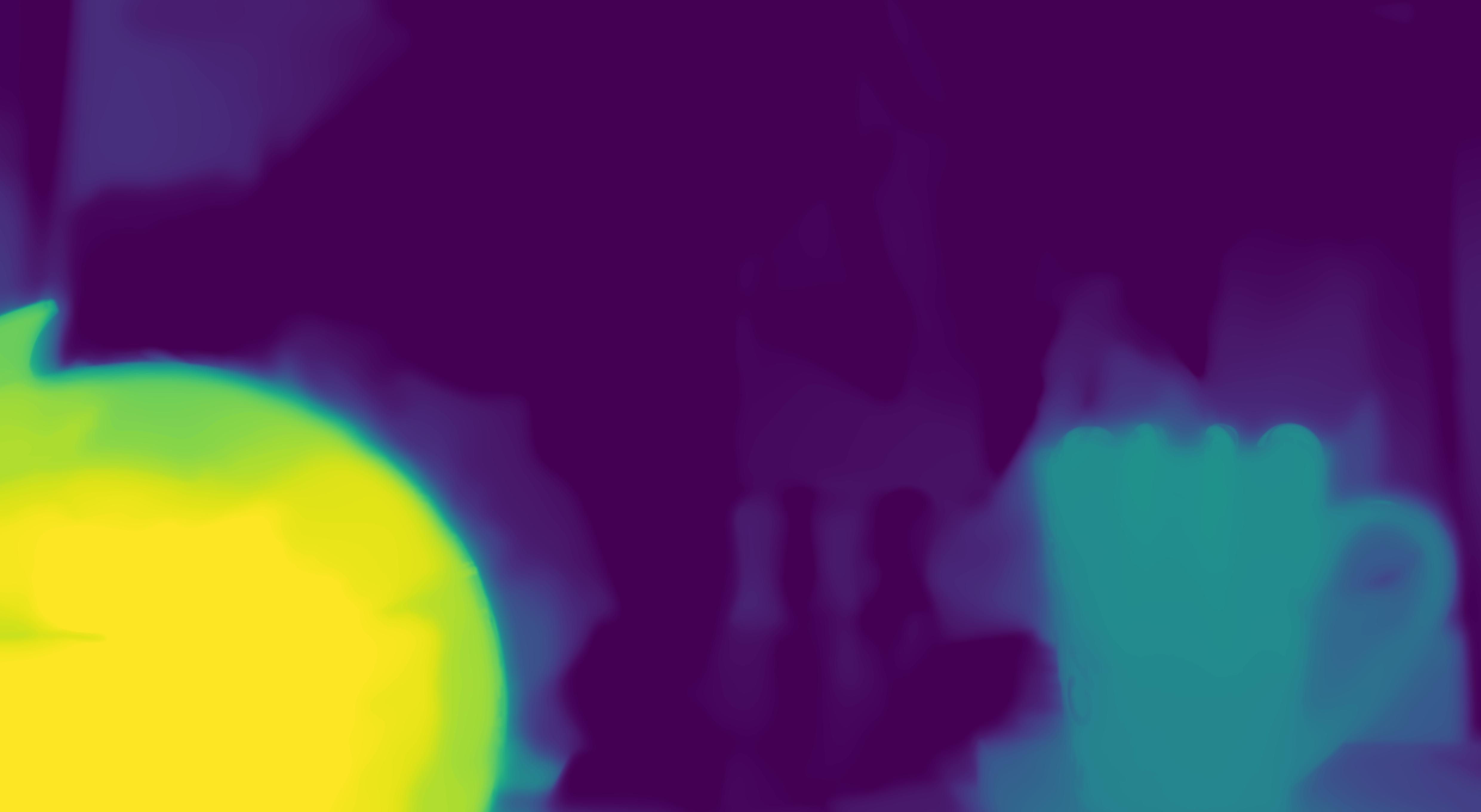}

    \end{tabular}
    \end{center}
 \caption{Results from DSLR examples:(a) Input image, (b) ground truth (GT), (c) SDoF ~\cite{ Wadhwa_SIGGRAPH2018_syntheticDoF} (as implemented in~\cite{Punnappurath-ICCP2020-modelingDefocus}), (d) DPdisp~\cite{Punnappurath-ICCP2020-modelingDefocus} (e) DPE~\cite{Pan_CVPR2021_dpexploration} (improved model, images provided by the author of~\cite{Punnappurath-ICCP2020-modelingDefocus}), (f) our method, (h) our method with filter. All the images have undergone an affine transform to best fit the GT.}
 \label{fig:DSLR_Quantitive}
\end{figure}

\begin{table}[t!]   
\begin{tabular}{|c|c|c|c|c|}
    \hline
	Method & AI(1) & AI(2) &  $ 1-|\rho_s|$ & Geometric \\ 
        &&&& Mean \\ 
    \hline
    SDoF~\cite{Wadhwa_SIGGRAPH2018_syntheticDoF} & 0.087 & 0.129 & 0.291 & 0.144 \\
    \hline
    DPdisp~\cite{Punnappurath-ICCP2020-modelingDefocus} & 0.047 & 0.074 & 0.082 & 0.065 \\
    \hline
    DPE~\cite{Pan_CVPR2021_dpexploration} & 0.061 & 0.098 & 0.103 & 0.110 \\
    \hline 
    CCA & $\mathbf{0.041}$ & $\mathbf{0.068}$ & $\mathbf{0.061}$ & $\mathbf{0.053}$ \\
    \hline
    CCA $+$ filter & $\mathbf{0.036}$ & $\mathbf{0.061}$ & $\mathbf{0.0490}$ & $\mathbf{0.048}$ \\
    \hline

\end{tabular}
\caption{Quantitative evaluation on DSLR data-set \cite{Punnappurath-ICCP2020-modelingDefocus} using the error metrics  \cite{Punnappurath-ICCP2020-modelingDefocus}; the right-most column shows the geometric mean of all the metrics.
}
\label{tab:DSLR_res}
\end{table}

We evaluate our method on a data-set from~\cite{Punnappurath-ICCP2020-modelingDefocus}, being the data captured with a Canon DSLR camera; two sub-sets are available, with and without ground-truth (GT), respectively; we refer to these as DLSR-A and DSLR-B data-sets. We compare our results to the following DP disparity estimation algorithms: DPdisp~\cite{Punnappurath-ICCP2020-modelingDefocus} which is an optimization method that models the different PSF of both views, SDoF~\cite{ Wadhwa_SIGGRAPH2018_syntheticDoF} that uses a classic local disparity estimation (results are adopted from its implementation in~\cite{Punnappurath-ICCP2020-modelingDefocus}), and DPE~\cite{Pan_CVPR2021_dpexploration} which is a learning based approach. The results of DPE on DSLR-A data-set are provided by the authors of~\cite{Pan_CVPR2021_dpexploration} using an improved fine-tuned model, compared to the results on DSLR-B  data-set that are obtained by the publicly available model.

\begin{figure}[t]
    \setlength\abovecaptionskip{-0.6\baselineskip} 
    \setlength\belowcaptionskip{-15pt} 
  \begin{center}
   \begin{tabular}{c@{~}c@{~}c@{~}c}
       {\raisebox{0.8cm}{\rotatebox[origin=c]{90}{~\footnotesize{(a)Image}}}}&
      \includegraphics[width=0.3\linewidth]{\Figs/DSLR_Qualtitive/Images/imagesOrig1.jpg} &
       \includegraphics[width=0.3\linewidth]{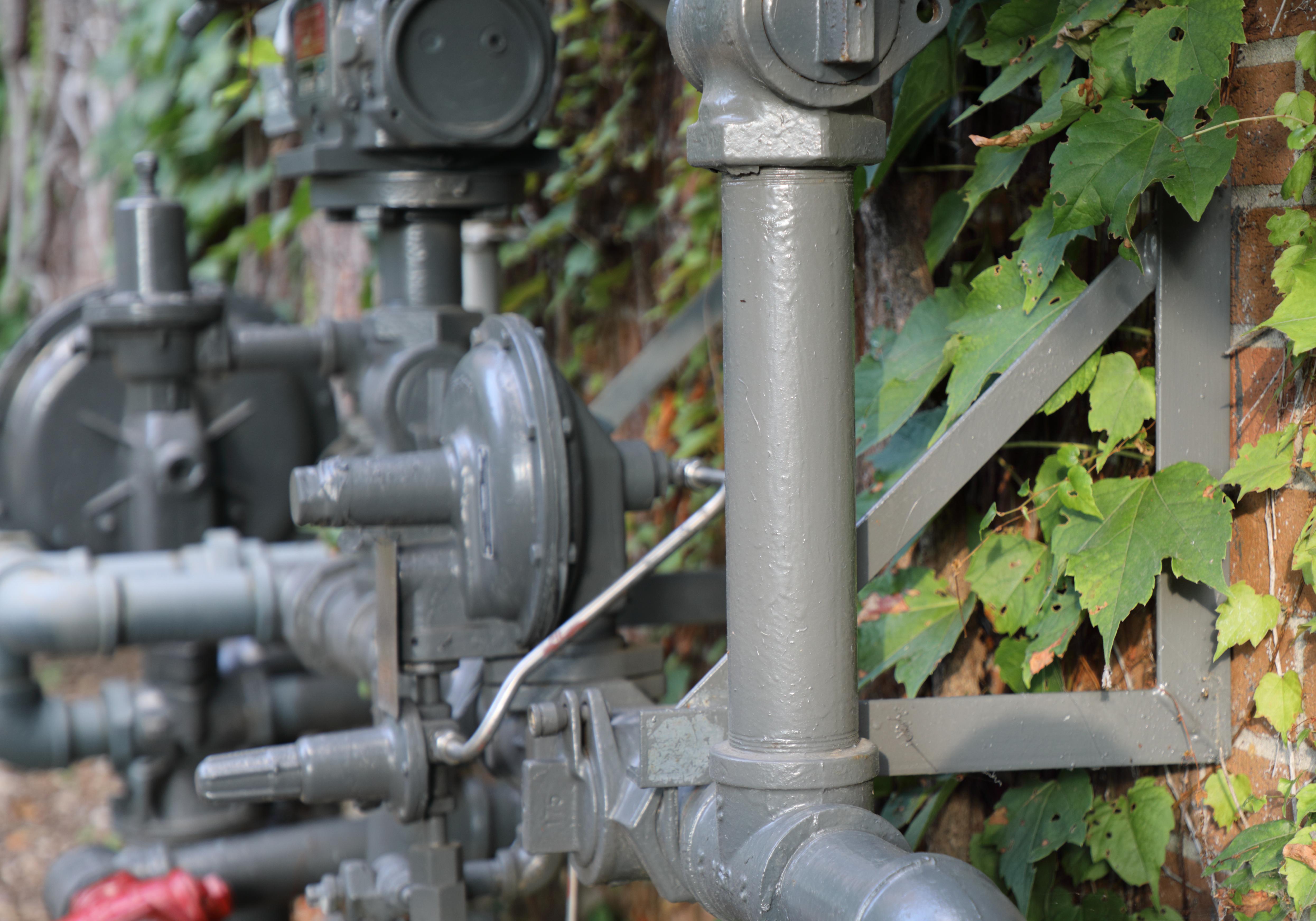} &
       \includegraphics[width=0.3\linewidth]{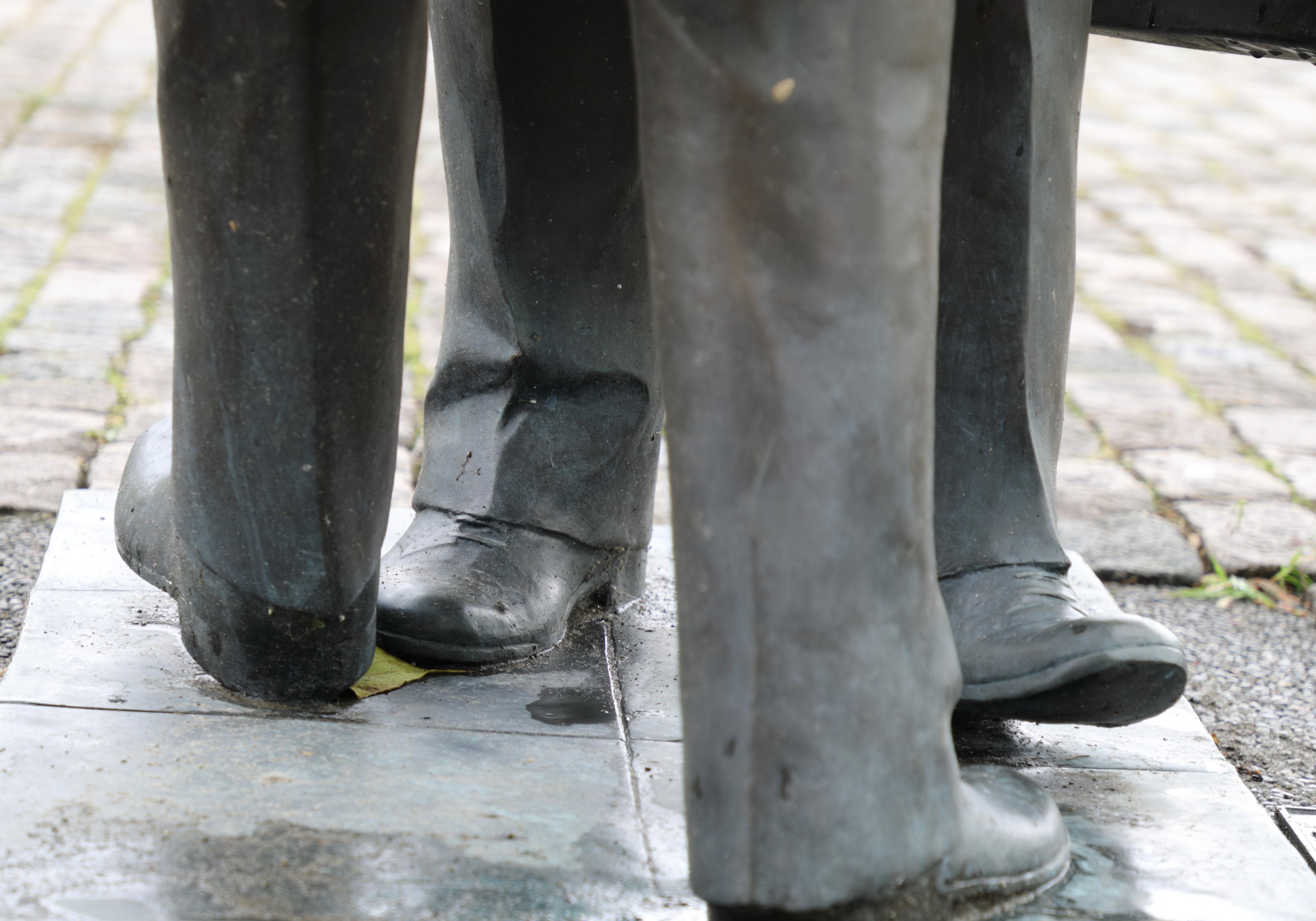} \\
       

       {\raisebox{0.8cm}{\rotatebox[origin=c]{90}{~\footnotesize{(b)SDoF~\cite{Wadhwa_SIGGRAPH2018_syntheticDoF}}}}}&
      \includegraphics[width=0.3\linewidth]{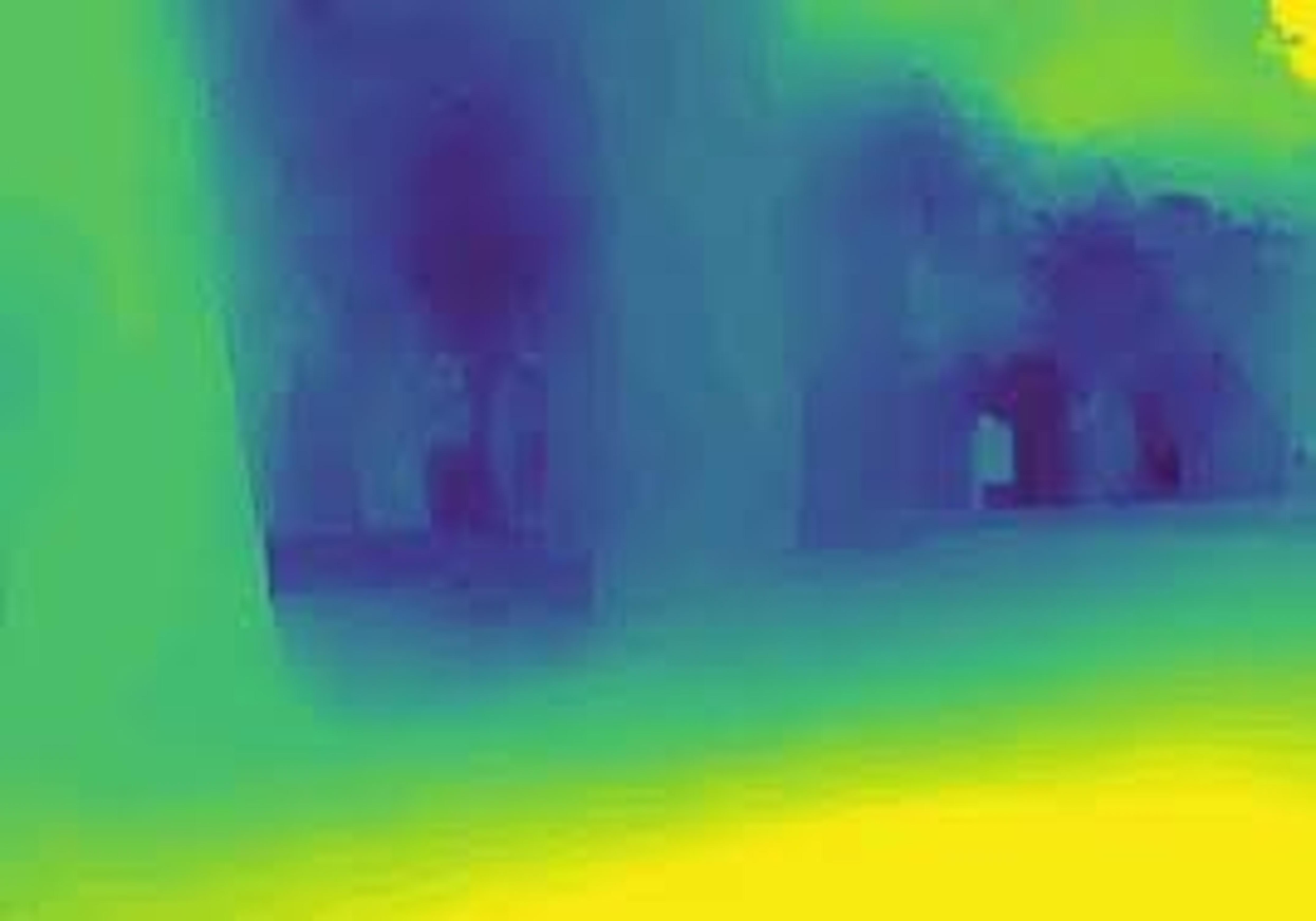} &
       \includegraphics[width=0.3\linewidth]{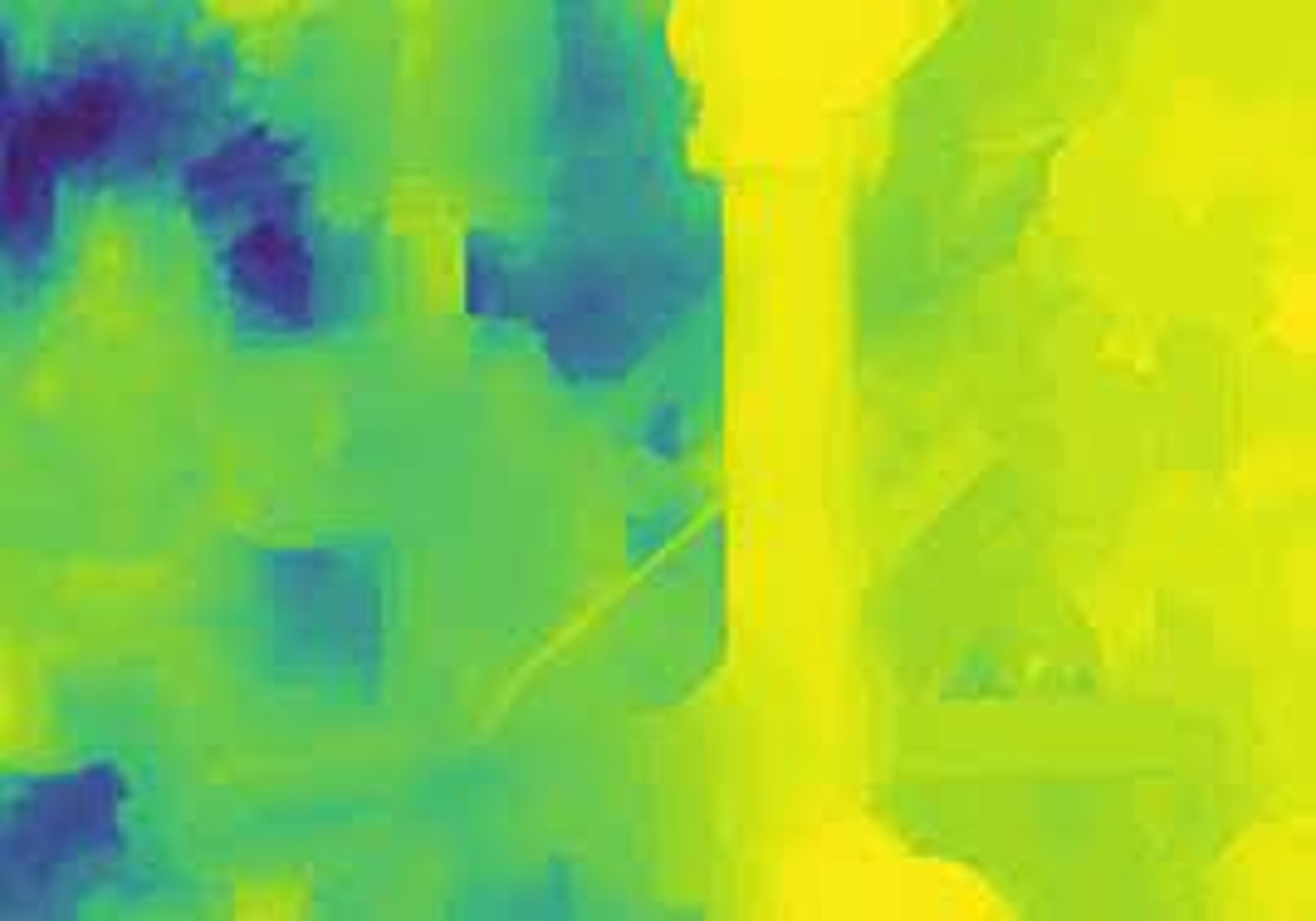} &
       \includegraphics[width=0.3\linewidth]{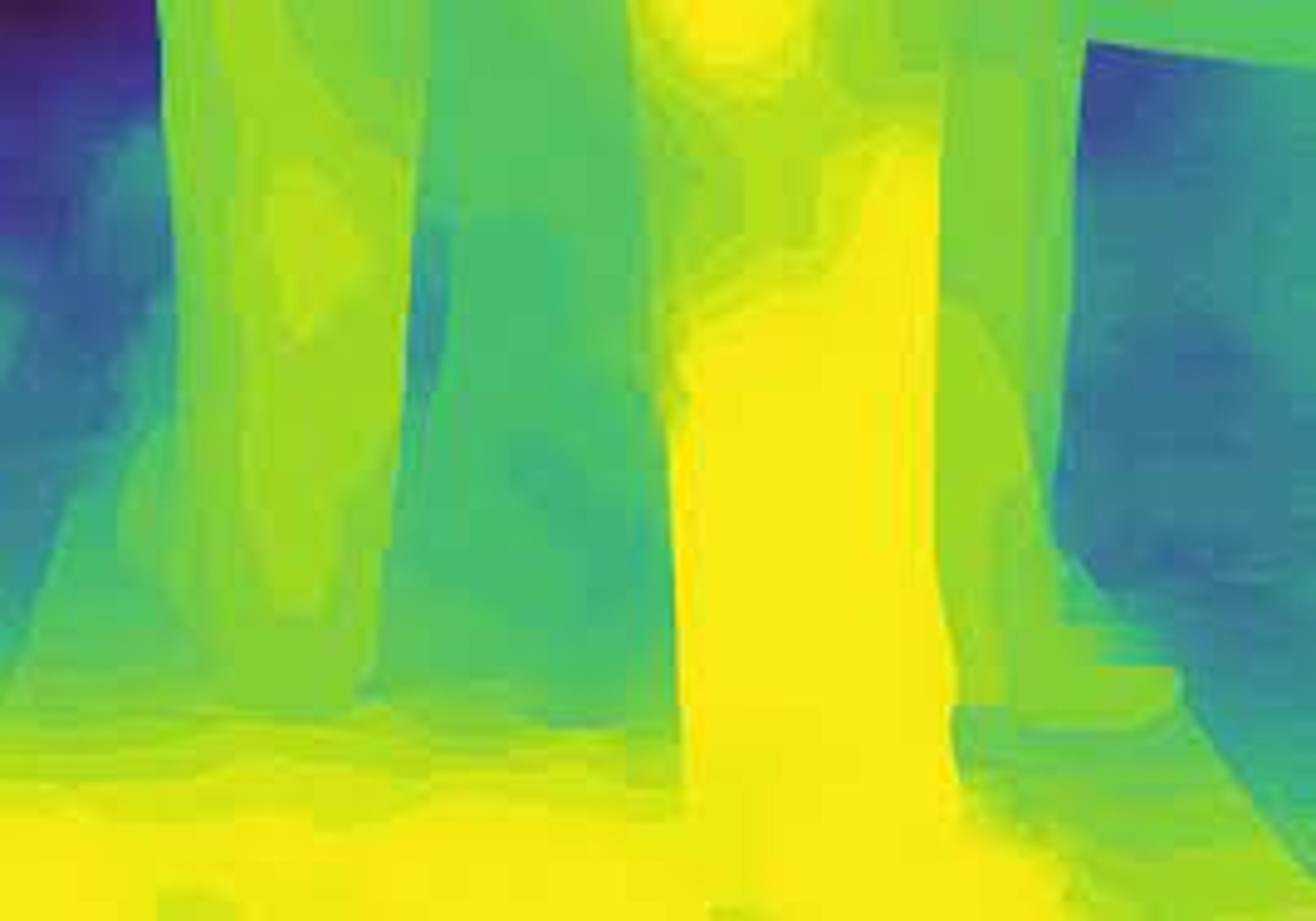} \\
      
       {\raisebox{0.8cm}{\rotatebox[origin=c]{90}{~\footnotesize{(c)DPdisp~\cite{Punnappurath-ICCP2020-modelingDefocus}}}}}&
      \includegraphics[width=0.3\linewidth]{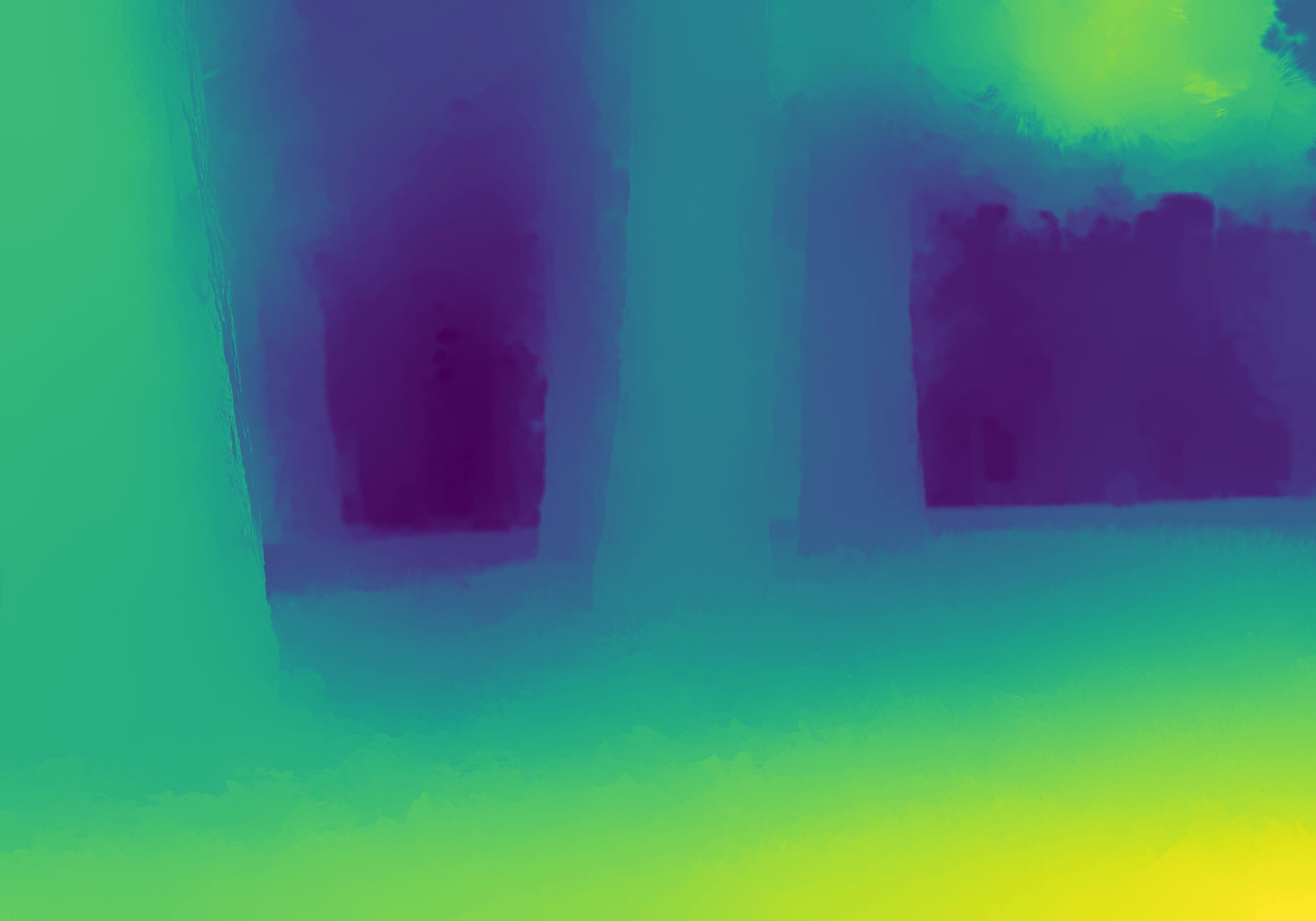} &
       \includegraphics[width=0.3\linewidth]{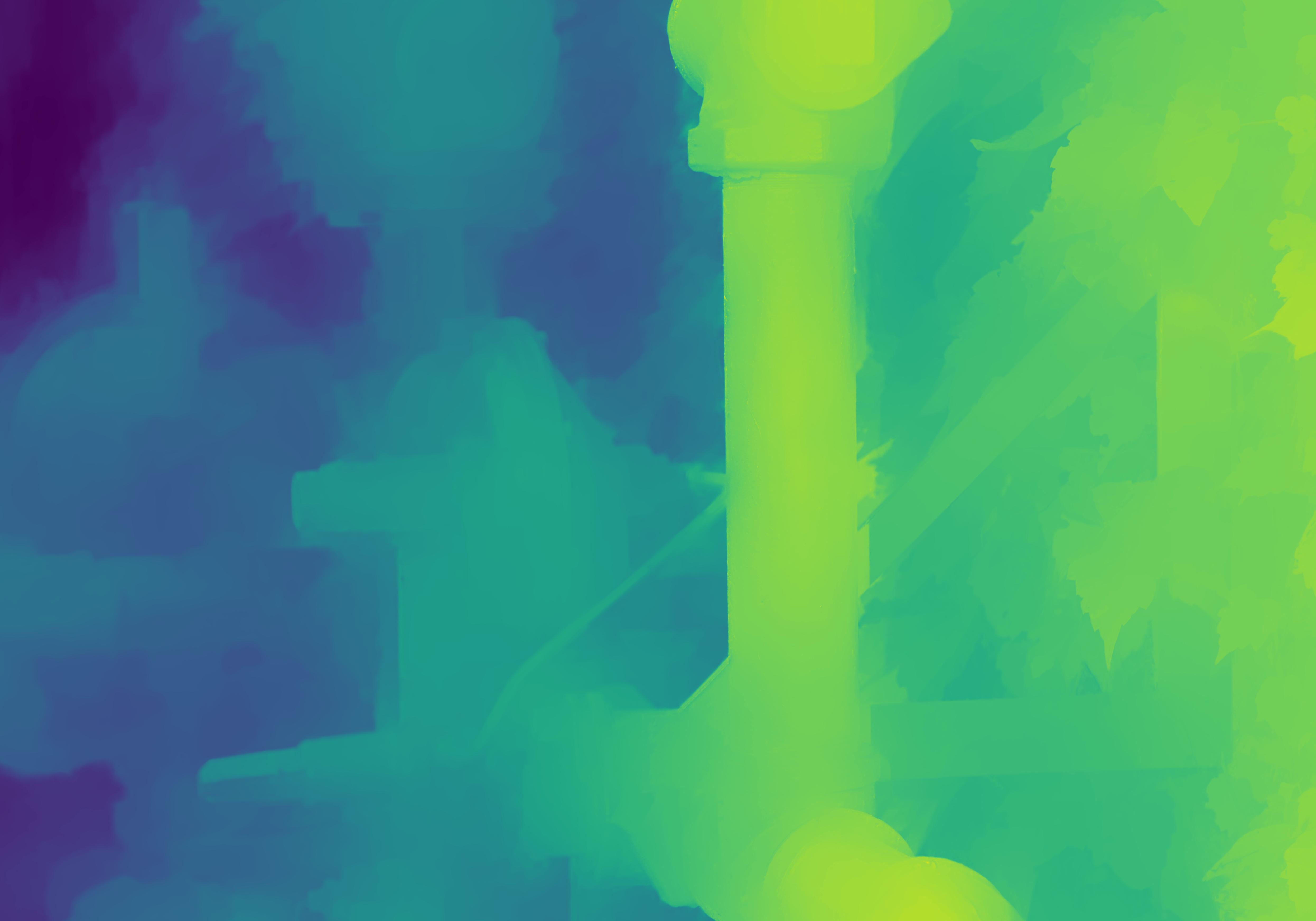} &
       \includegraphics[width=0.3\linewidth]{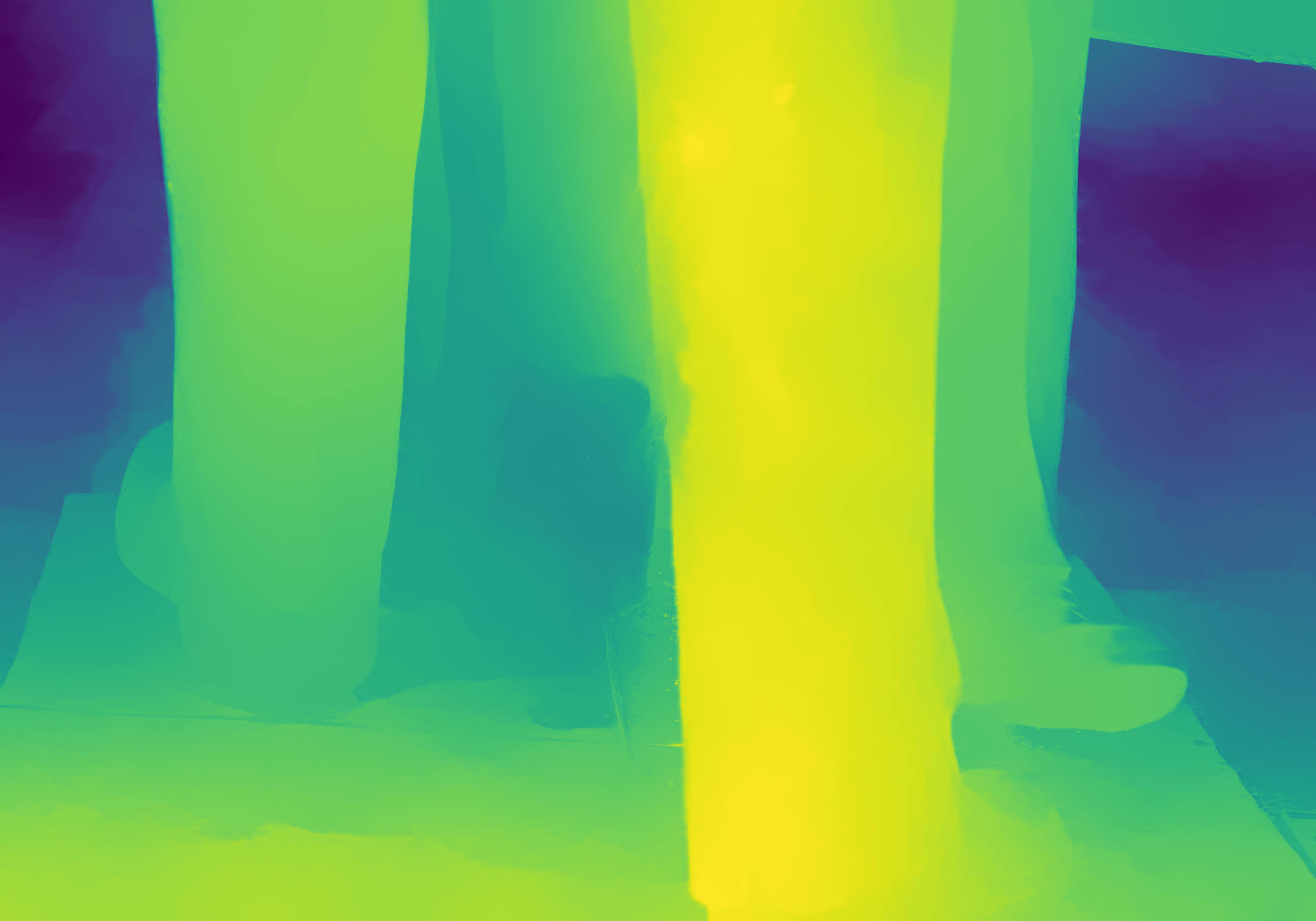} \\

       {\raisebox{0.7cm}{\rotatebox[origin=c]{90}{~\footnotesize{(d)DPE~\cite{Pan_CVPR2021_dpexploration}}}}}&
      \includegraphics[width=0.3\linewidth]{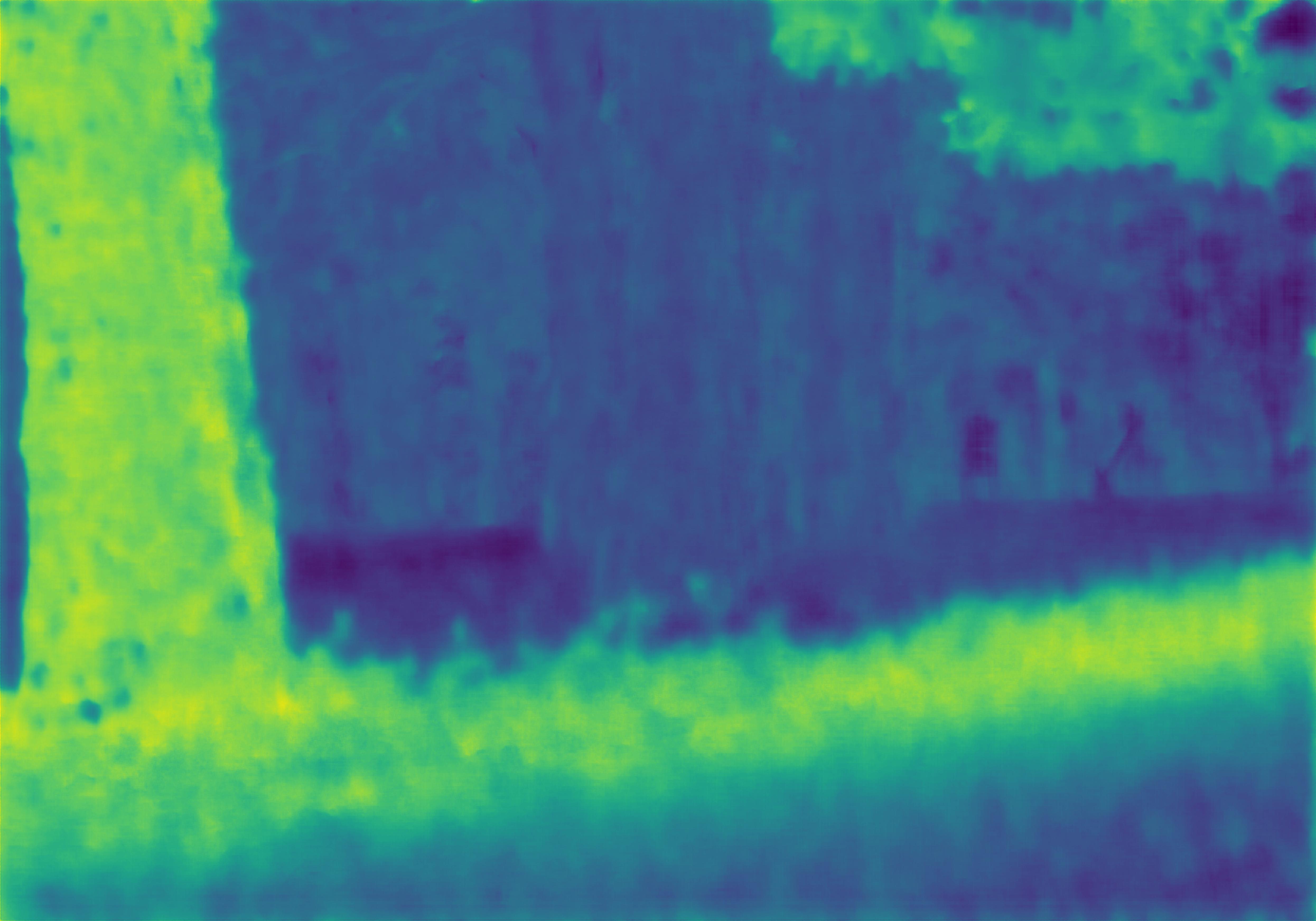} &
       \includegraphics[width=0.3\linewidth]{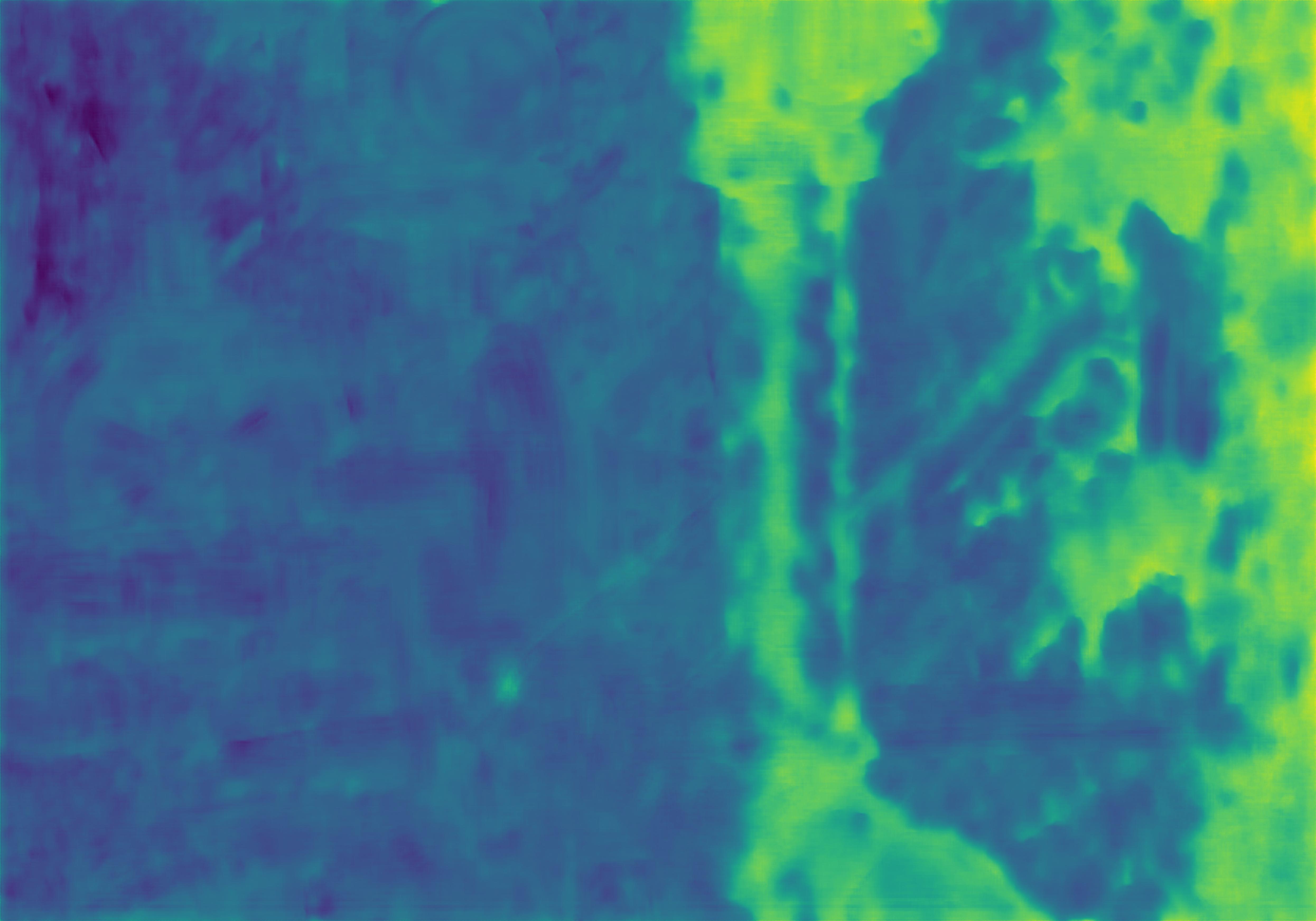} &
       \includegraphics[width=0.3\linewidth]{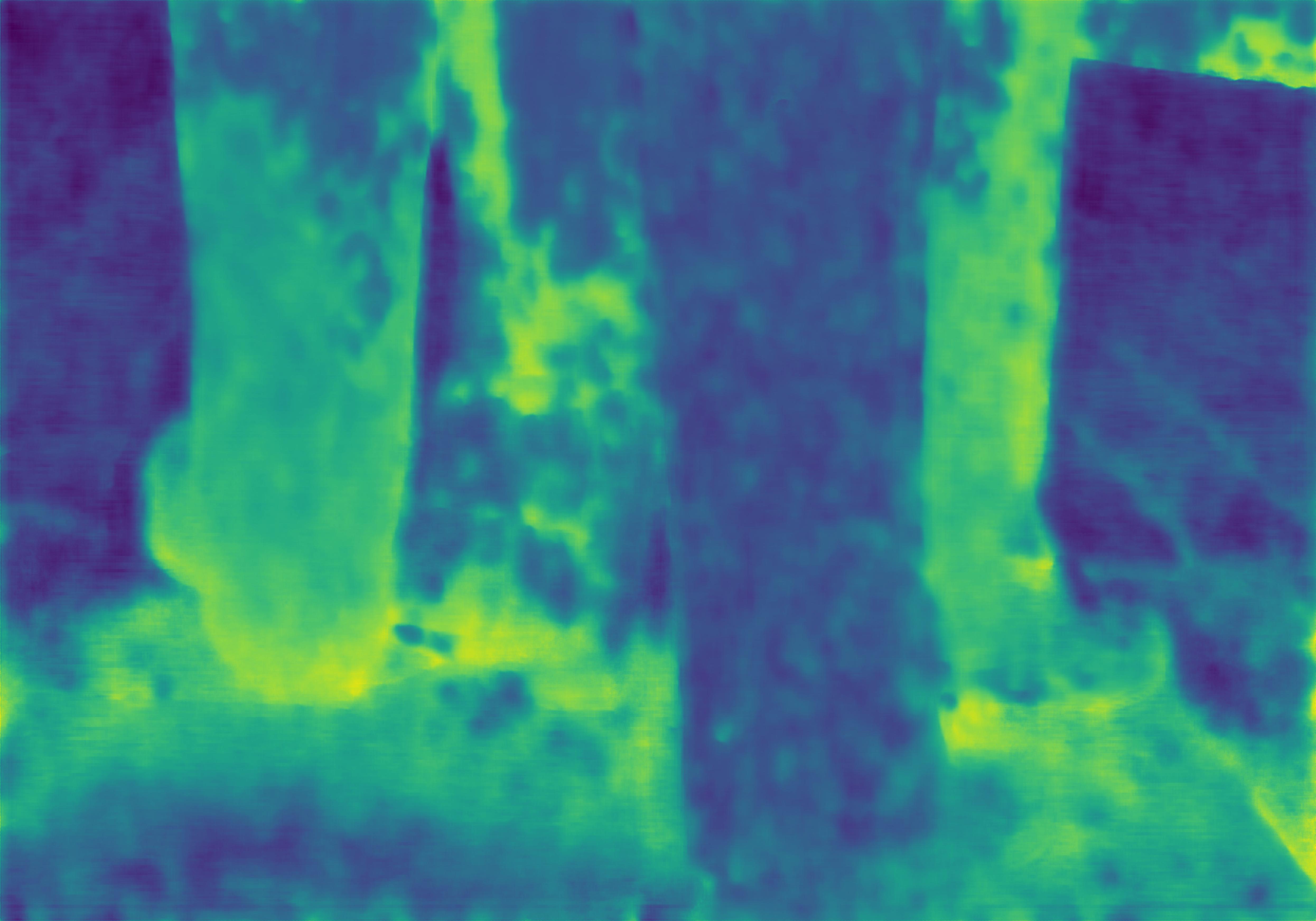} \\

       {\raisebox{0.8cm}{\rotatebox[origin=c]{90}{~\footnotesize{(e)CCA}}}}&
      \includegraphics[width=0.3\linewidth]{\Figs/DSLR_Qualtitive/CSGM/Disparity_MAP_Qualtitve_1.jpg} &
       \includegraphics[width=0.3\linewidth]{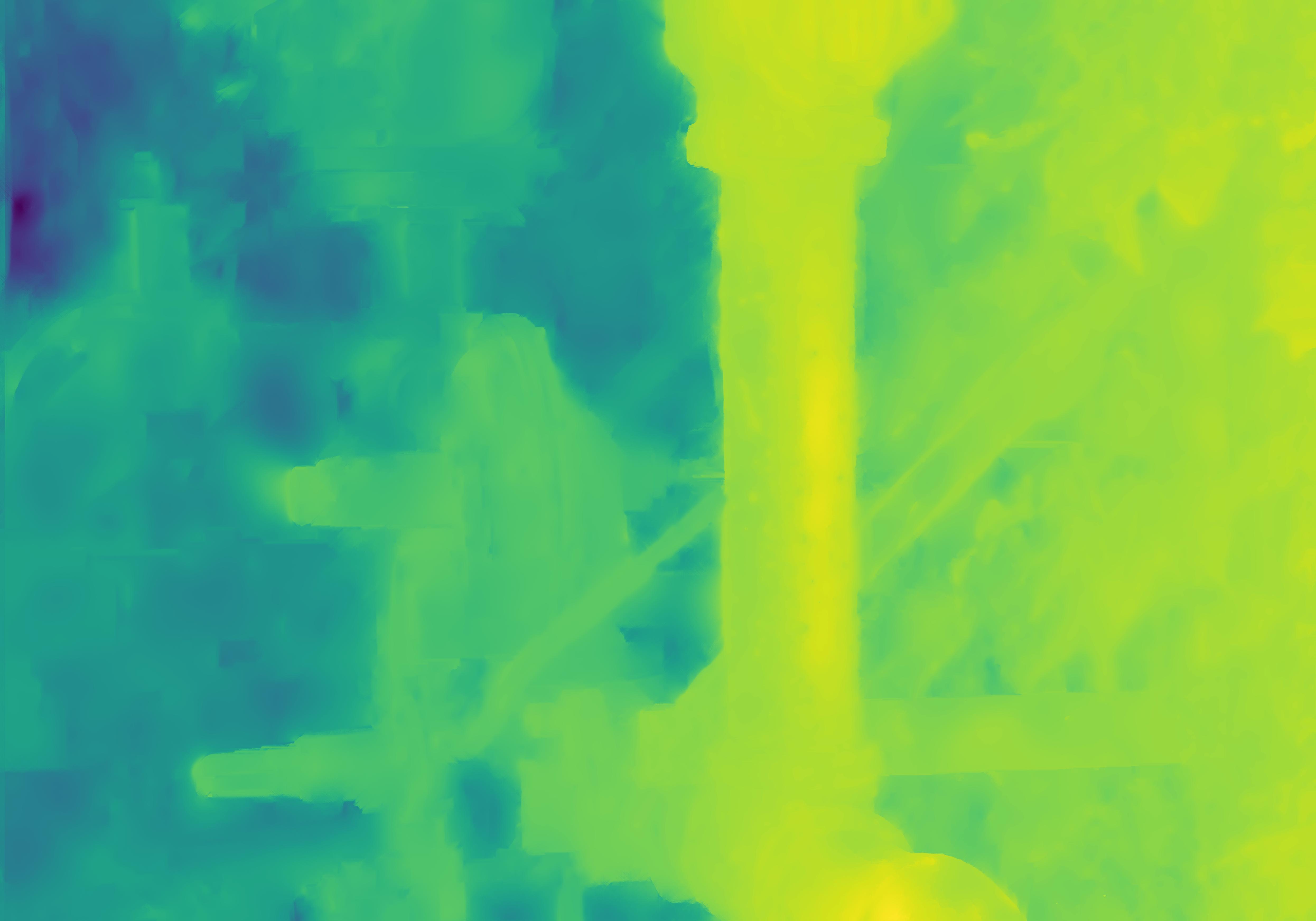} &
       \includegraphics[width=0.3\linewidth]{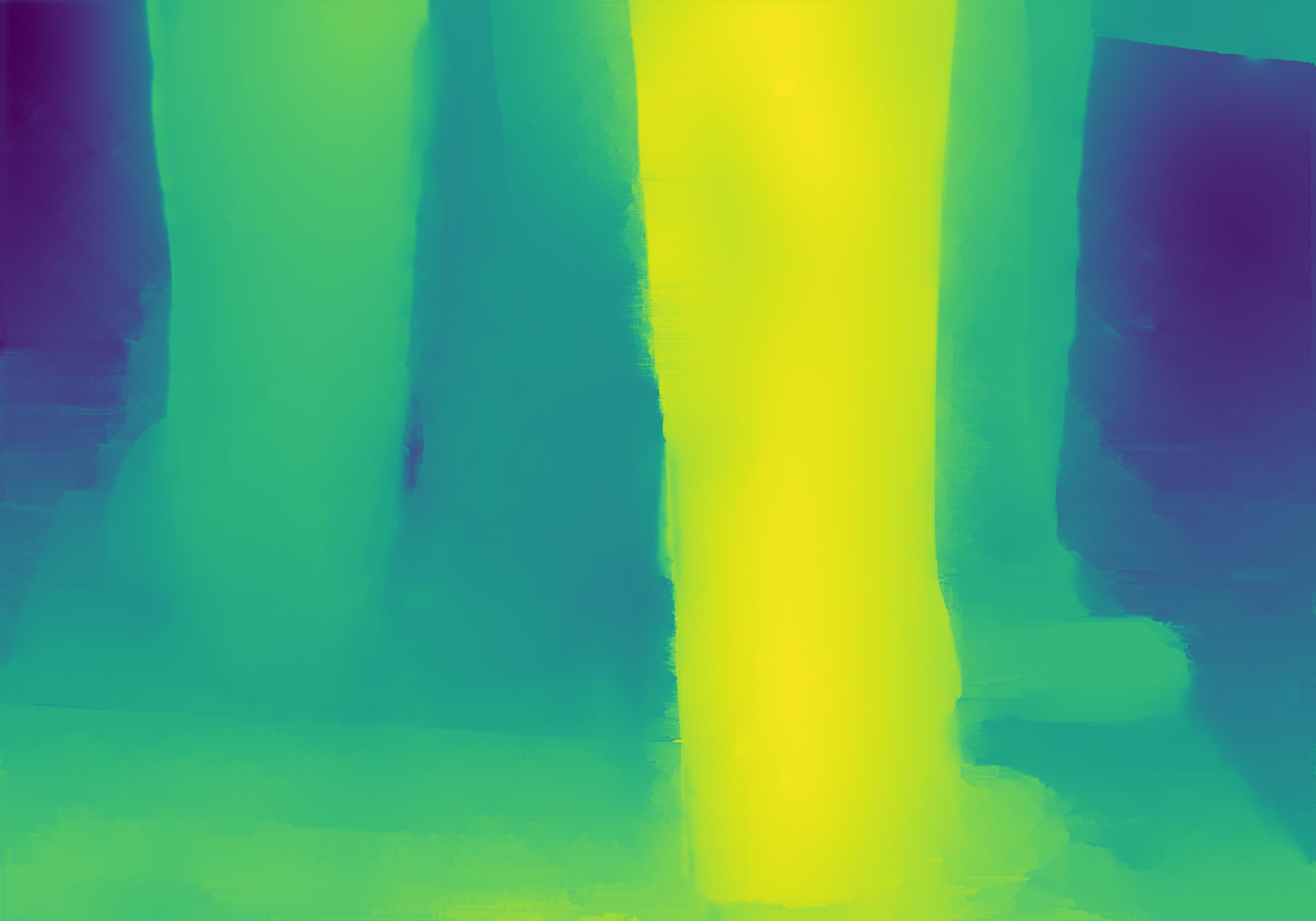} \\

       {\raisebox{0.8cm}{\rotatebox[origin=c]{90}{~\footnotesize{(f)CCA $+$ filter}}}}&
      \includegraphics[width=0.3\linewidth]{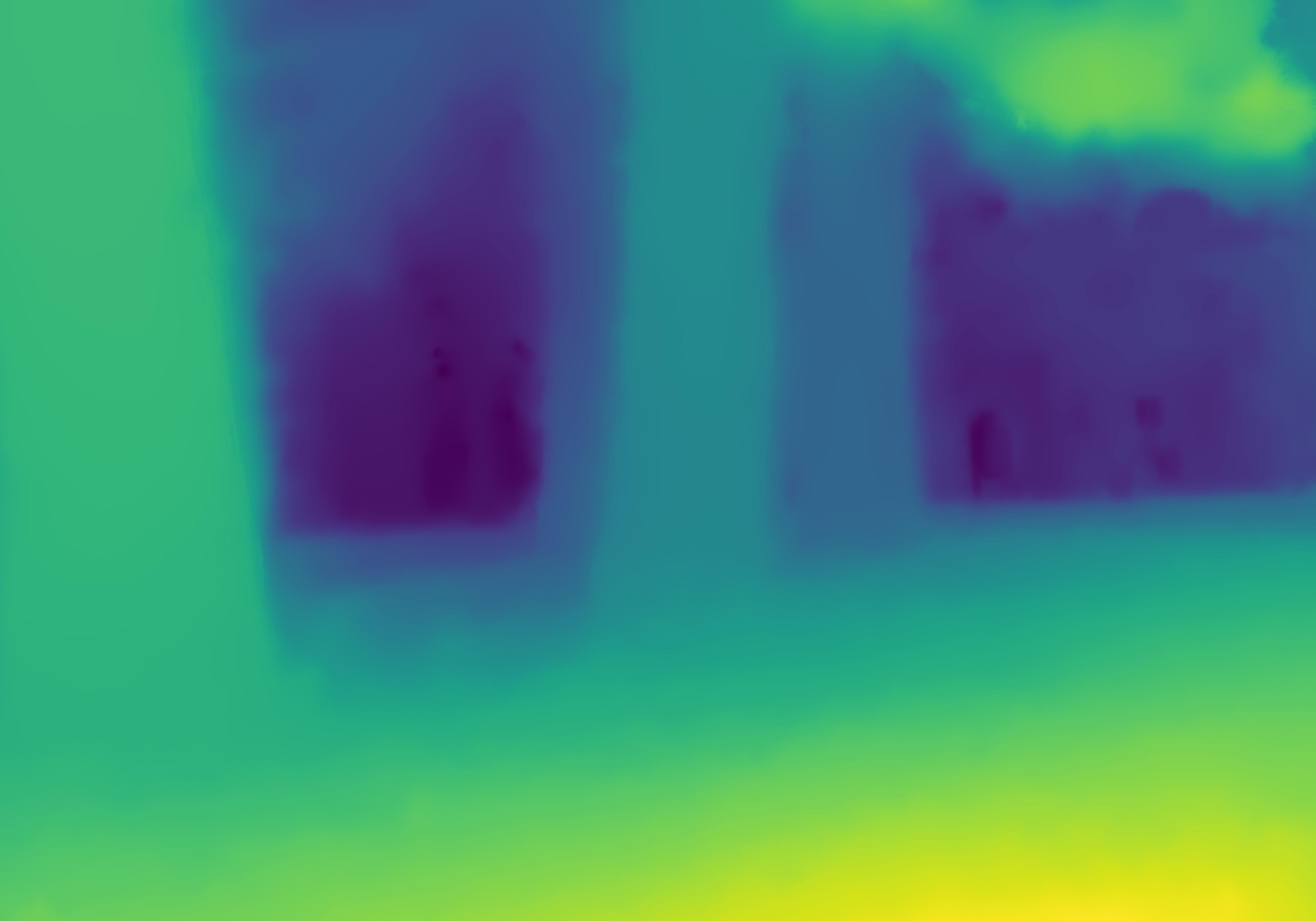} &
       \includegraphics[width=0.3\linewidth]{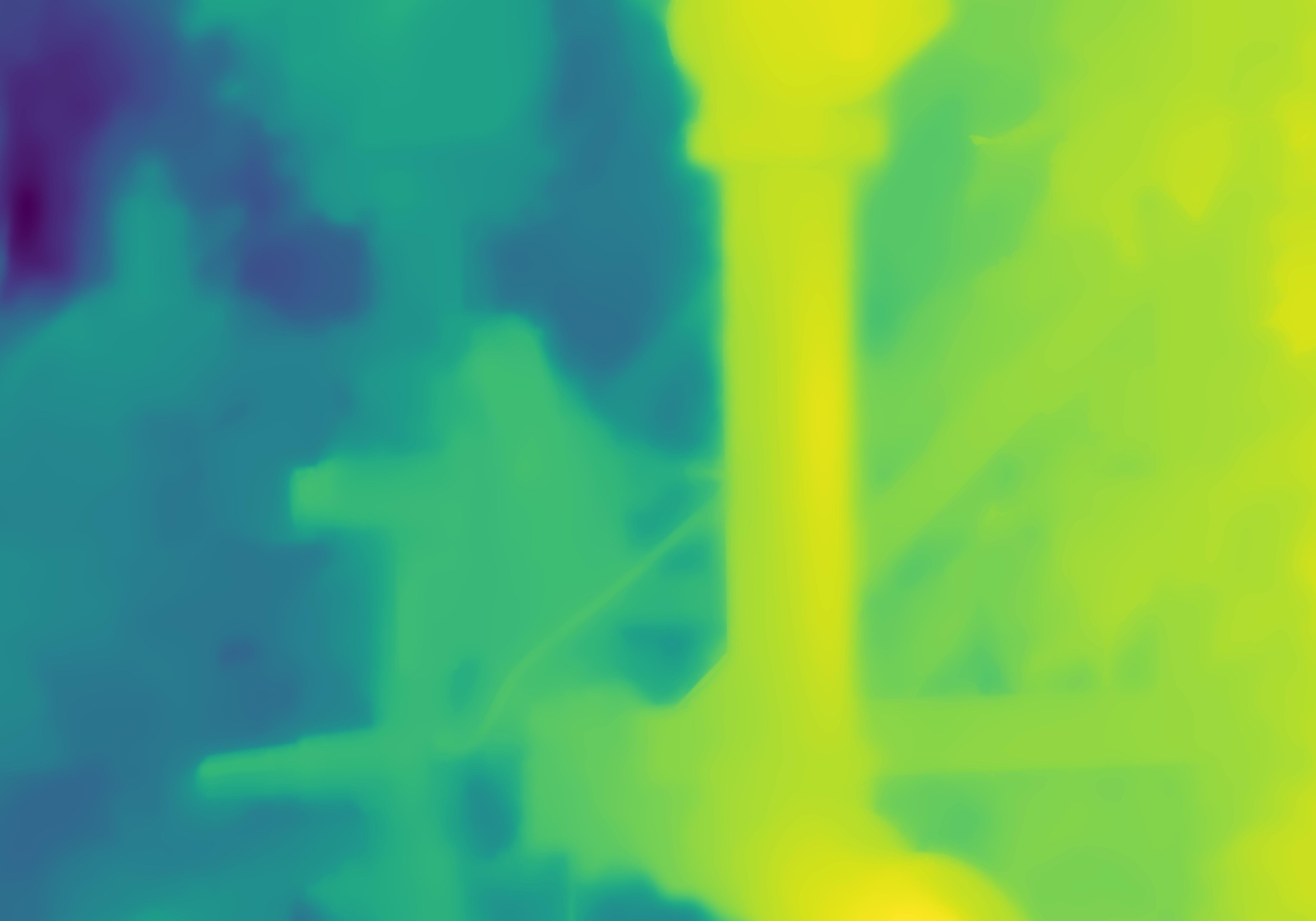} &
       \includegraphics[width=0.3\linewidth]{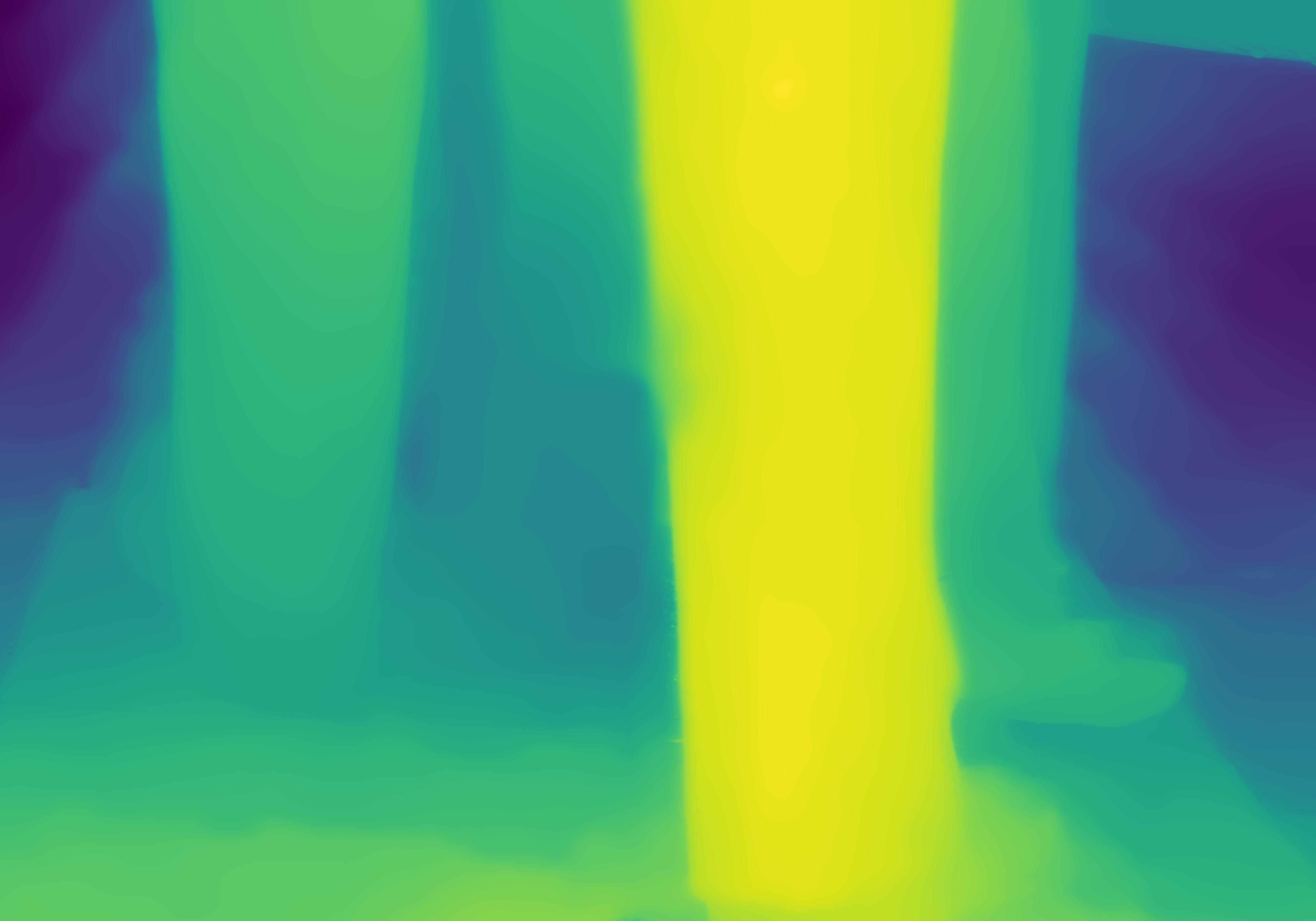}
       
    \end{tabular}
    \end{center}
 \caption{Qualitative results from examples of DSLR data-set:(a) Input image, (b) SDoF ~\cite{ Wadhwa_SIGGRAPH2018_syntheticDoF} (as implemented in~\cite{Punnappurath-ICCP2020-modelingDefocus}), (c) DPdisp~\cite{Punnappurath-ICCP2020-modelingDefocus}, (d) DPE~\cite{Pan_CVPR2021_dpexploration} (publicly available model)
 (e) our method, (f) our method with filter.}
 \label{fig:DSLR_Qualtitive}
 \vskip -0.3cm
\end{figure}

As pointed out in~\cite{Garg-ICCV2019-learningDual}, disparity or inverse depth results can be recovered up to an unknown affine ambiguity. Therefore, we consider the affine-invariant error metrics of~\cite{Punnappurath-ICCP2020-modelingDefocus} to conduct a quantitative evaluation. Table~\ref{tab:DSLR_res} compares the performance of algorithms on the full DSLR-A data-set. As seen, CCA outperforms the other algorithms, even without post-processing. \figref{fig:DSLR_Quantitive} and \figref{fig:DSLR_Qualtitive} qualitatively compare the algorithms by showing results of images examples from DSLR-A and DSLR-B data-sets, respectively. As seen in these figures, CCA significantly outperforms SDoF~\cite{Wadhwa_SIGGRAPH2018_syntheticDoF} and DPE~\cite{Pan_CVPR2021_dpexploration} that suffer from strong artifacts. In comparison to DPdisp~\cite{Punnappurath-ICCP2020-modelingDefocus}, CCA performs better on the boundary of objects (e.g. horns of the action figure, toy car and mug handle in \figref{fig:DSLR_Quantitive}). Note that all the images in \figref{fig:DSLR_Quantitive} have undergone an affine transform to best fit GT.

\subsubsection{Phone Captured Images}

The second data-set is from~\cite{Garg-ICCV2019-learningDual}, which provides data captured by Google Pixel 2 and 3 with GT inverse depth-maps. 
Unlike DSLR images, this data-set suffers from more aberrations, noise and only captures green-channel data. In addition, phone cameras have large depth of field and small aperture, resulting in small sub-pixel disparities. 
Therefore, we pre-process the images. Since we do not have access to calibration data or devices for a proper vignetting calibration~\cite{xin2021defocus}, we compensate for vignetting by using a low-pass filter (LPF), i.e., $I_L^{calib} = I_L \cdot (LPF(I_R)/LPF(I_L))$. 
 We then remove local photometric distortions using a subtraction bilateral filter~\cite{Heiko2009Eval}. Because of the low image quality, we do not use ENCC for the initial subpixel offset. Instead, we use the histogram-equalized kernel of~\cite{Miclea2015New} that provides a less-biased initial continuous disparity (see supplementary material). 
In addition to vignetting, the images suffer from field-curvature aberration, resulting in radial varying disparity. Unlike~\cite{Wadhwa_SIGGRAPH2018_syntheticDoF}, we are unable to calibrate these aberrations since we do not have calibration data.

\begin{figure}[t]
    \setlength\abovecaptionskip{-0.6\baselineskip} 
    \setlength\belowcaptionskip{-15pt} 
  \begin{center}
  \begin{tabular}{c@{~}c@{~}c@{~}c}

      {\raisebox{1.4cm}{\rotatebox[origin=c]{90}{~{(a)Image}}}}&
      \includegraphics[width=0.3\linewidth]{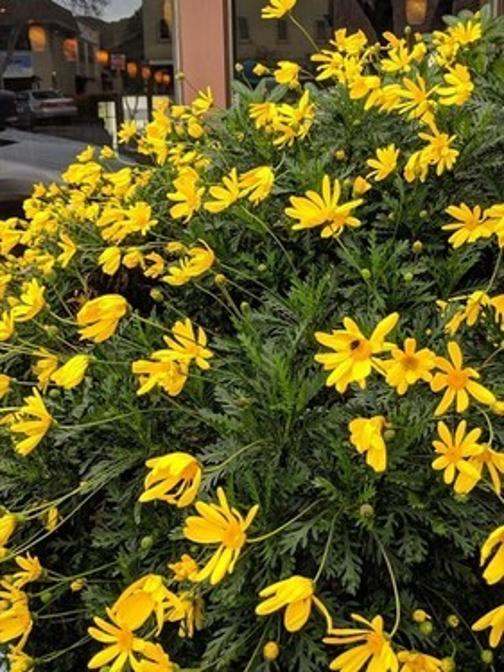} &
      \includegraphics[width=0.3\linewidth]{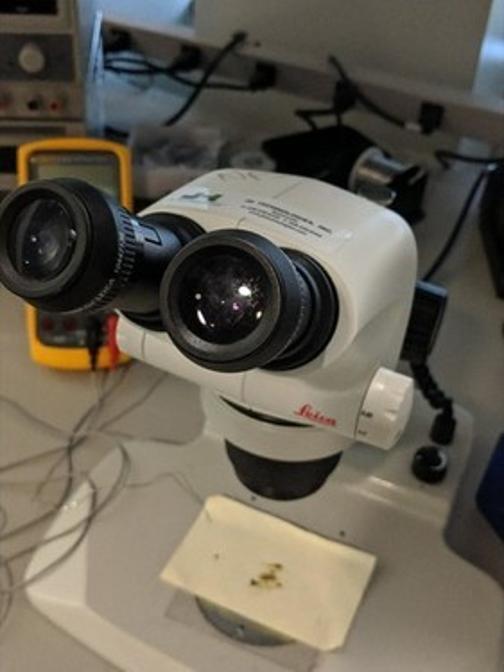} &
      \includegraphics[width=0.3\linewidth]{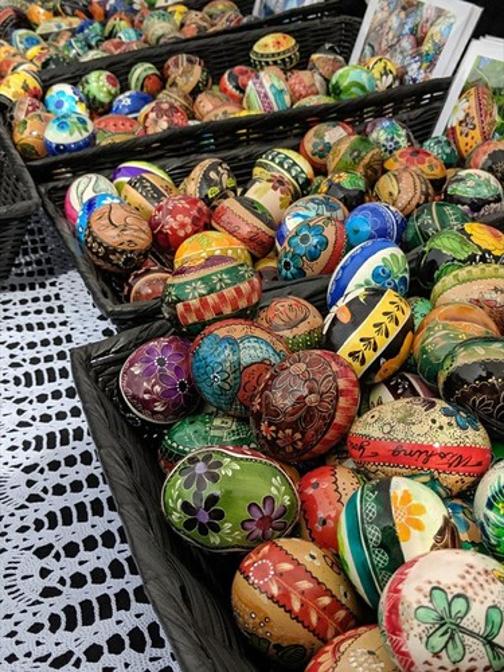} \\

      {\raisebox{1.4cm}{\rotatebox[origin=c]{90}{~{(b)GT}}}}&
      \includegraphics[width=0.3\linewidth]{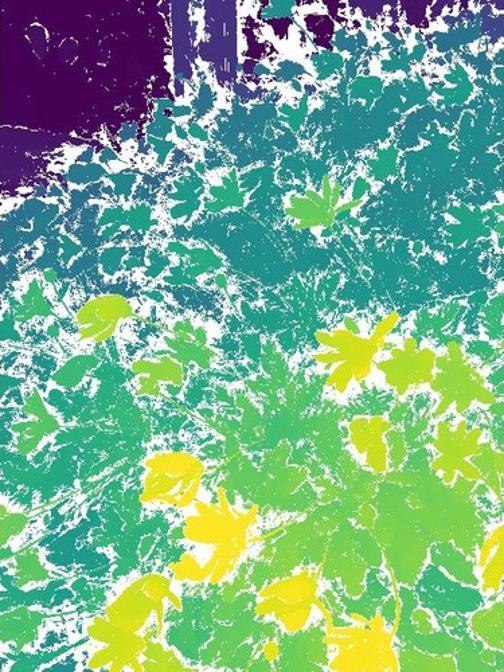} &
      \includegraphics[width=0.3\linewidth]{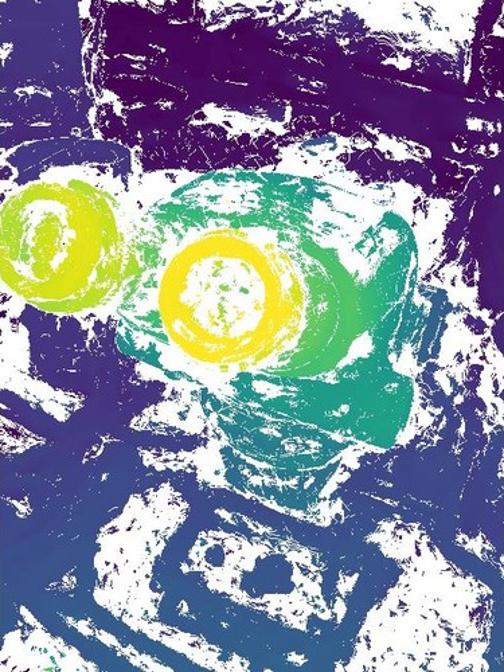} &
      \includegraphics[width=0.3\linewidth]{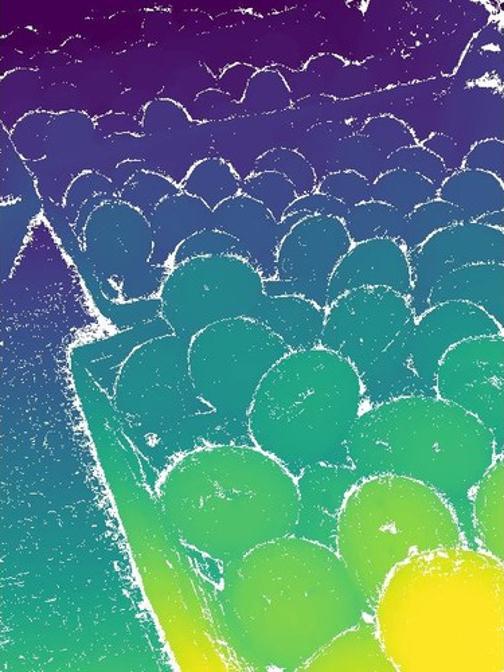} \\

      {\raisebox{1.2cm}{\rotatebox[origin=c]{90}{~{(c)SDoF~\cite{Wadhwa_SIGGRAPH2018_syntheticDoF}}}}}&
      \includegraphics[width=0.3\linewidth]{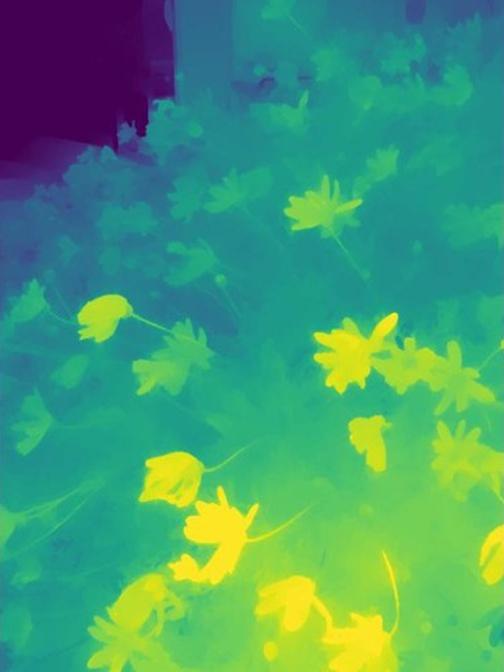} &
      \includegraphics[width=0.3\linewidth]{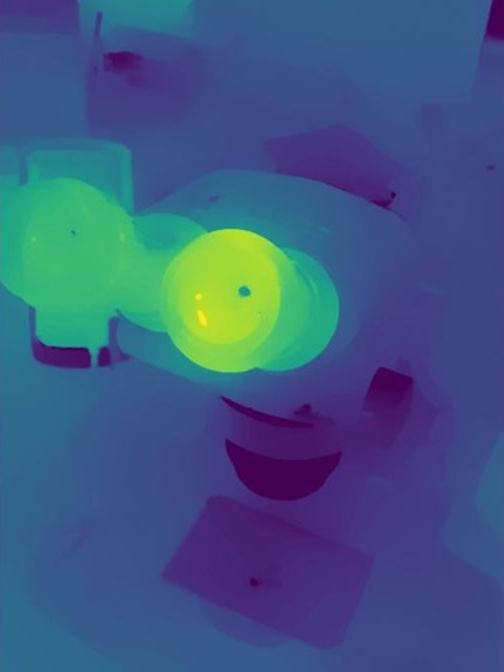} &
      \includegraphics[width=0.3\linewidth]{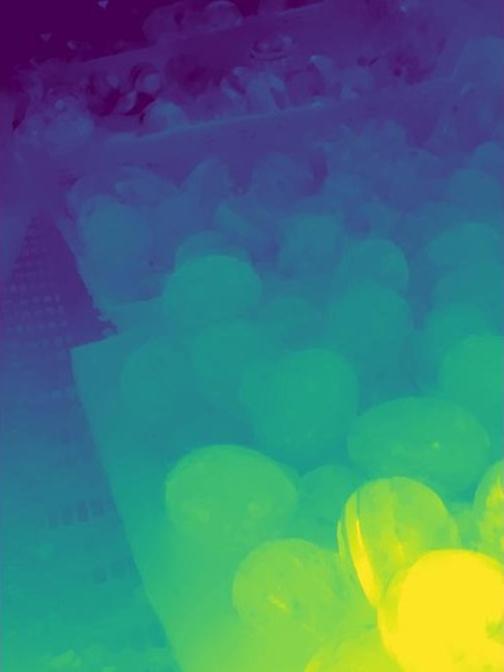} \\

      {\raisebox{1.4cm}{\rotatebox[origin=c]{90}{~{(d)CCA}}}}&
      \includegraphics[width=0.3\linewidth]{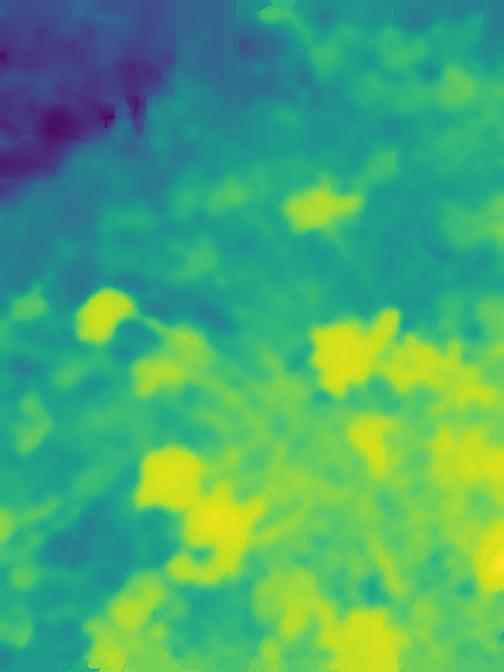} &
      \includegraphics[width=0.3\linewidth]{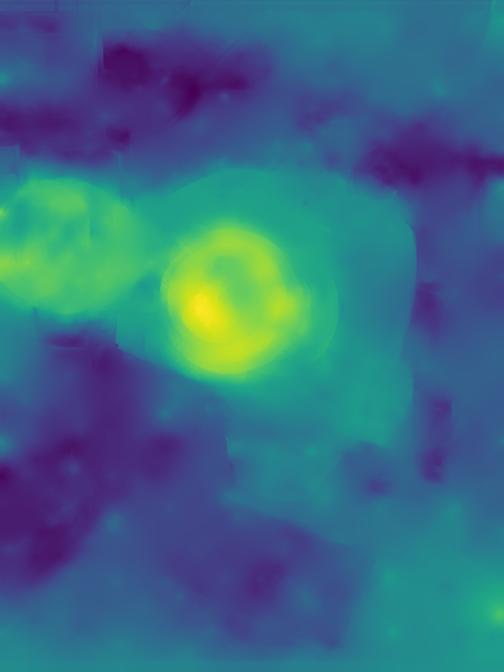} &
      \includegraphics[width=0.3\linewidth]{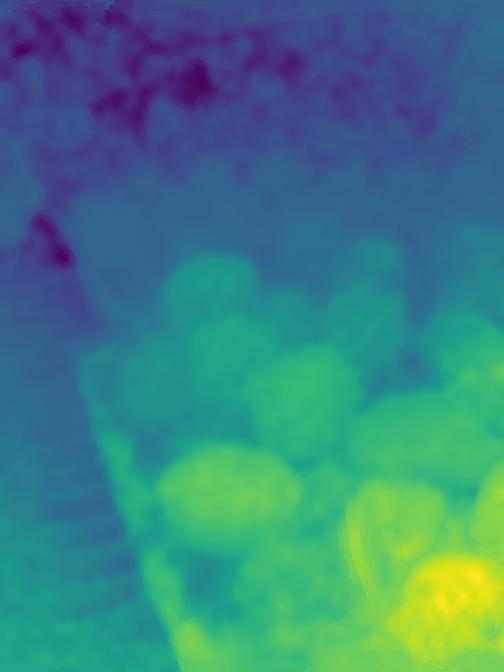} \\

      {\raisebox{1.4cm}{\rotatebox[origin=c]{90}{~{(e)CCA $+$ filter}}}}&
      \includegraphics[width=0.3\linewidth]{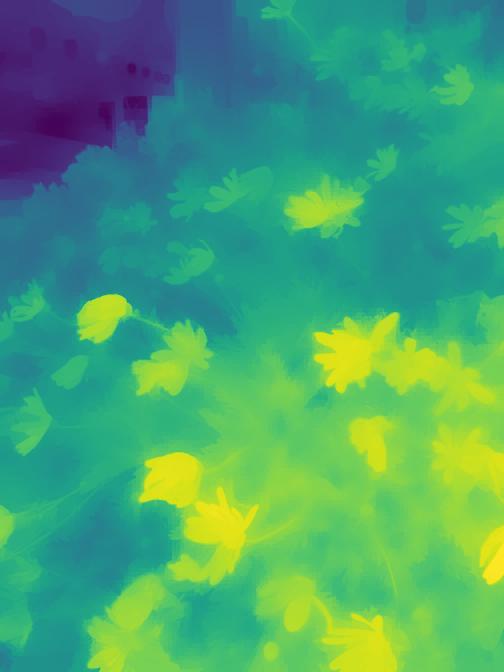} &
      \includegraphics[width=0.3\linewidth]{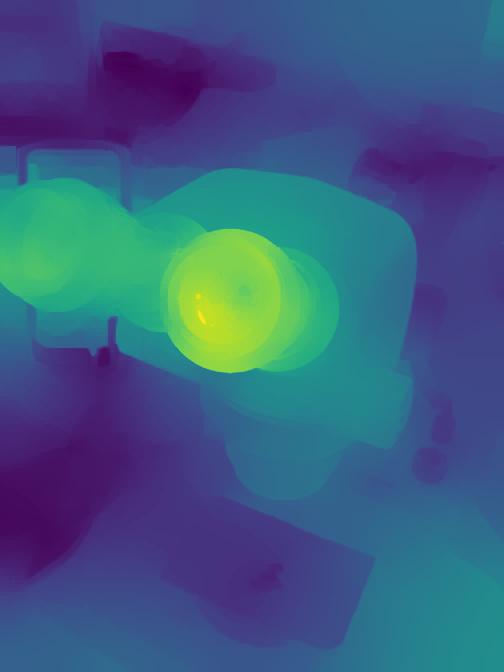} &
      \includegraphics[width=0.3\linewidth]{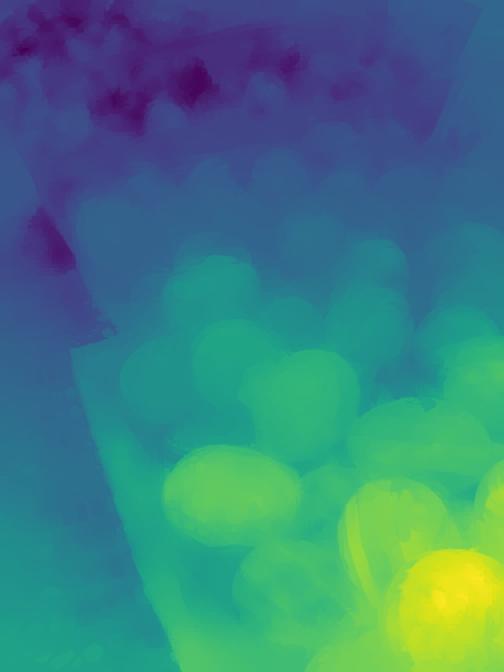}

    \end{tabular}
    \end{center}
\caption{Disparity examples from Google Pixel DP data-set. CCA performs similarly to SDoF~\cite{Wadhwa_SIGGRAPH2018_syntheticDoF} on small disparities. }
 \label{fig:PIXEL_result_main}
\end{figure}

\begin{table}[b!]   
\begin{tabular}{|c|c|c|c|c|}
    \hline
	Method & AI(1) & AI(2) &  $ 1-|\rho_s|$ & Geometric \\ 
        &&&& Mean \\ 
    \hline
    SDoF~\cite{Wadhwa_SIGGRAPH2018_syntheticDoF} & 0.027 & 0.037 & 0.236 & 0.063 \\
    \hline
    CCA & $\bold{0.026}$ & $\bold{0.036}$ & $\bold{0.225}$ & $\bold{0.059}$ \\
    \hline
    CCA $+$ filter & $\bold{0.025}$ & $\bold{0.035}$ & $\bold{0.217}$ & $\bold{0.057}$ \\
    \hline
\end{tabular}
\caption{Accuracy of different methods on Google-Pixel dataset \cite{Garg-ICCV2019-learningDual}. Lower is better. The right-most column shows the geometric mean of all the metrics.}
\label{tab:Pixel_res}
\end{table}


\begin{figure*}[t!]
    \setlength\abovecaptionskip{-0.6\baselineskip} 
    \setlength\belowcaptionskip{-15pt} 
  \begin{center}
   \begin{tabular}{c@{\hspace{1mm}}c@{\hspace{1mm}}c@{\hspace{1mm}}c@{\hspace{1mm}}c@{\hspace{1mm}}c}

       \includegraphics[width=0.15\linewidth]{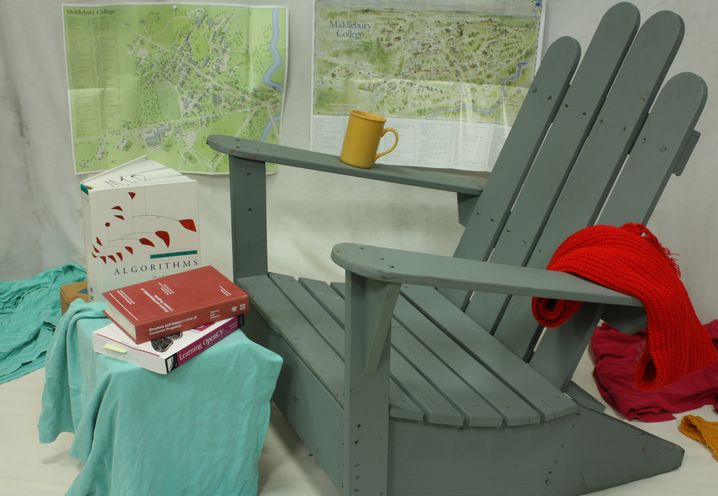} &
       \includegraphics[width=0.15\linewidth]{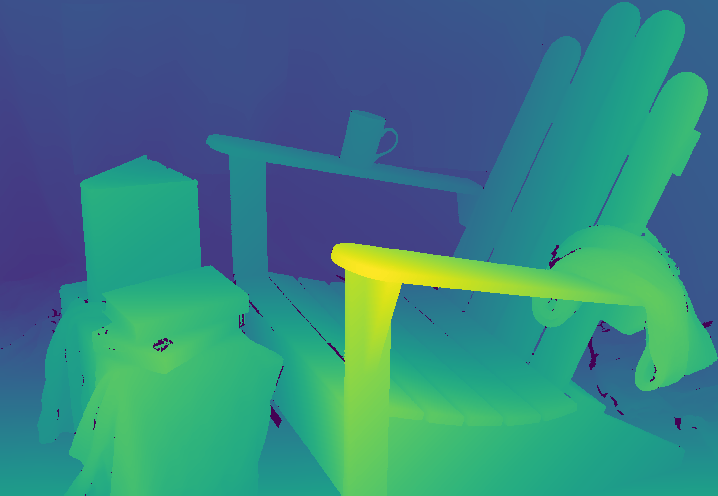} &
       \includegraphics[width=0.15\linewidth]{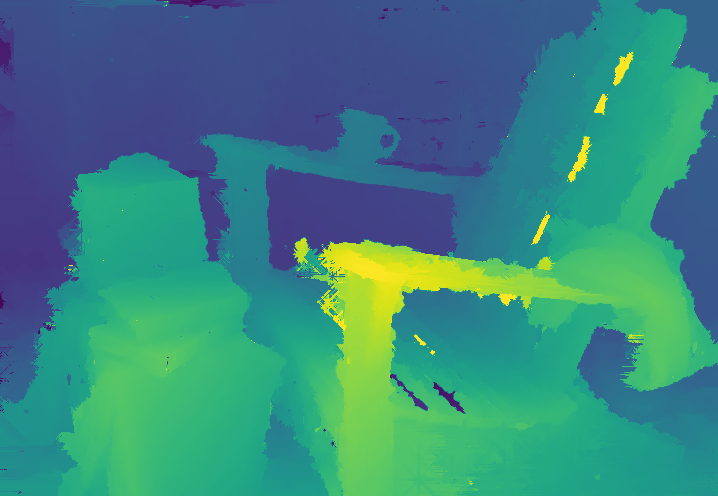} &
       \includegraphics[width=0.15\linewidth]{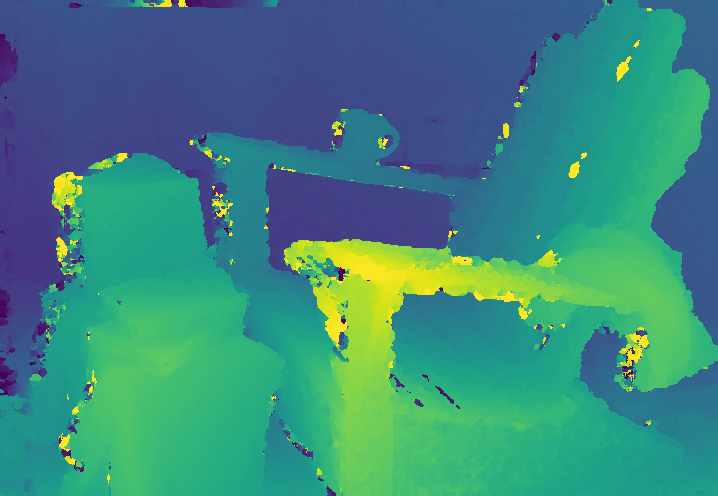} &
       \includegraphics[width=0.15\linewidth]{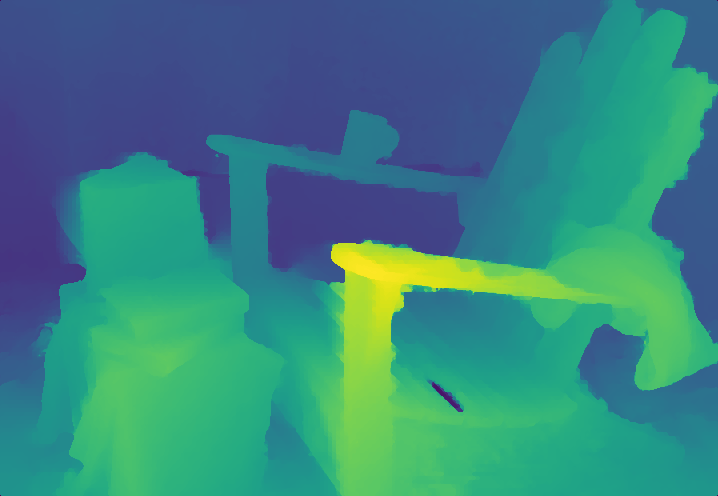} &
       \includegraphics[width=0.15\linewidth]{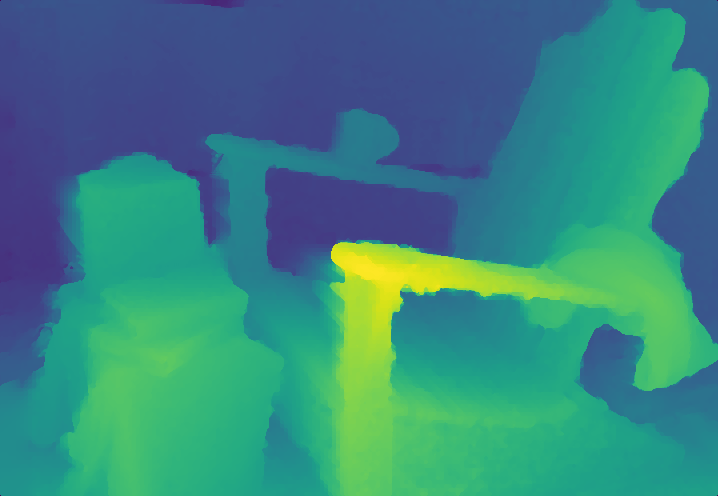} \\

       \includegraphics[width=0.15\linewidth]{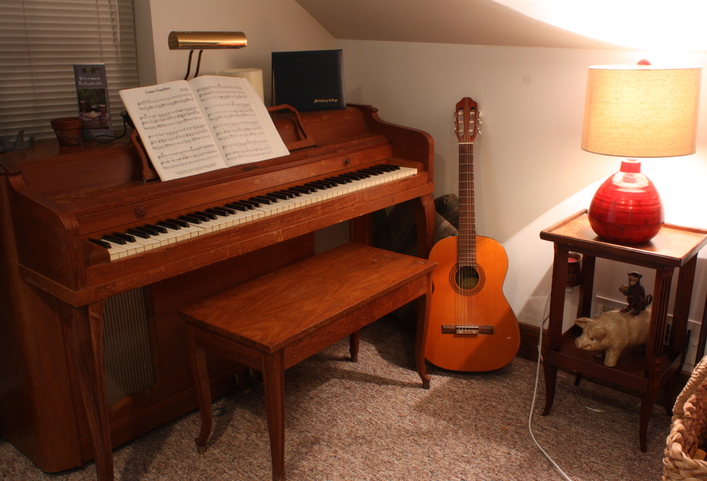} &
       \includegraphics[width=0.15\linewidth]{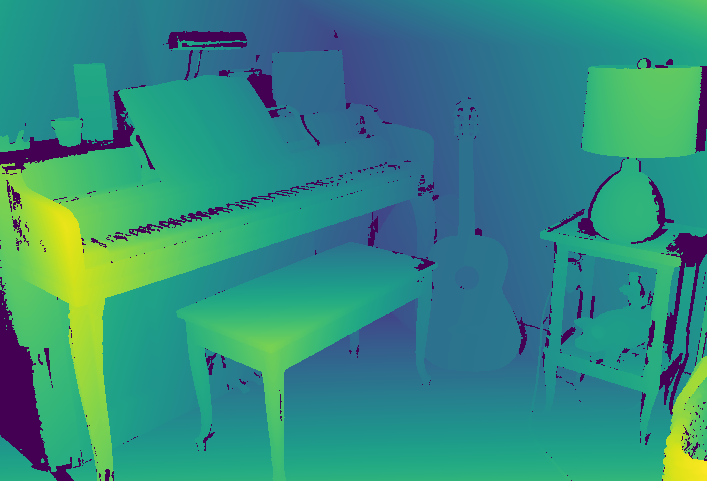} &
       \includegraphics[width=0.15\linewidth]{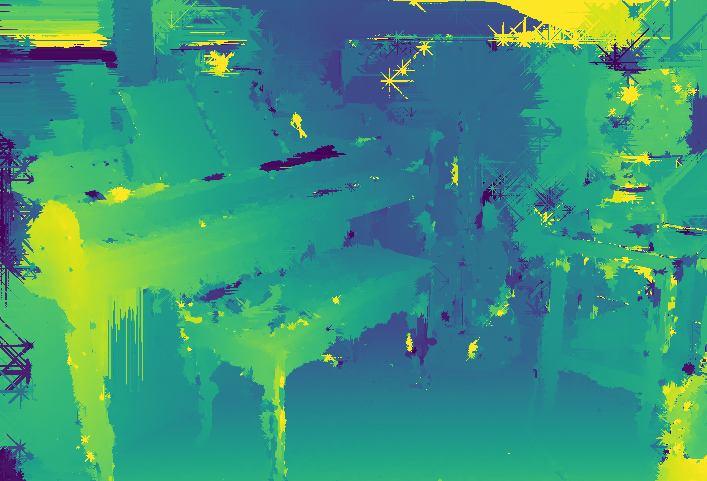} &
       \includegraphics[width=0.15\linewidth]{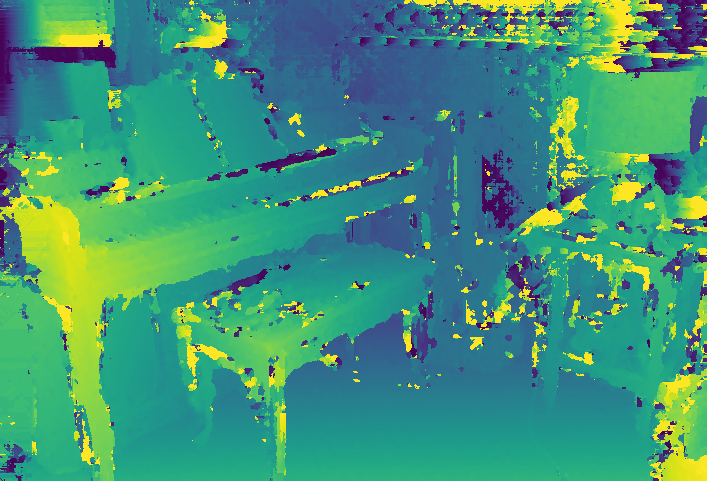} &
       \includegraphics[width=0.15\linewidth]{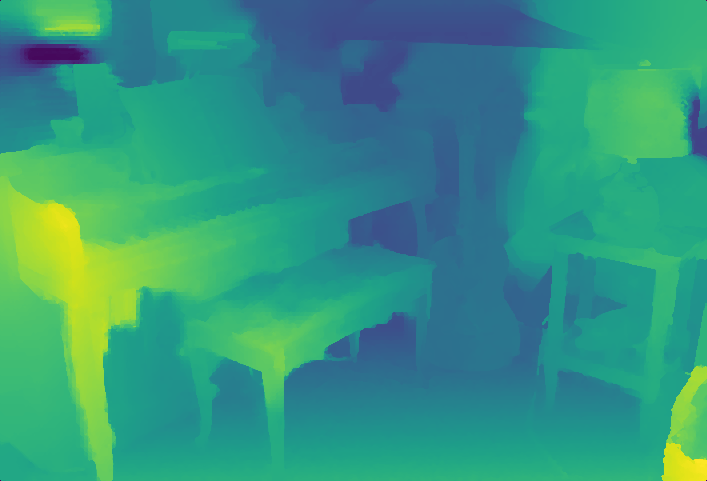} &
       \includegraphics[width=0.15\linewidth]{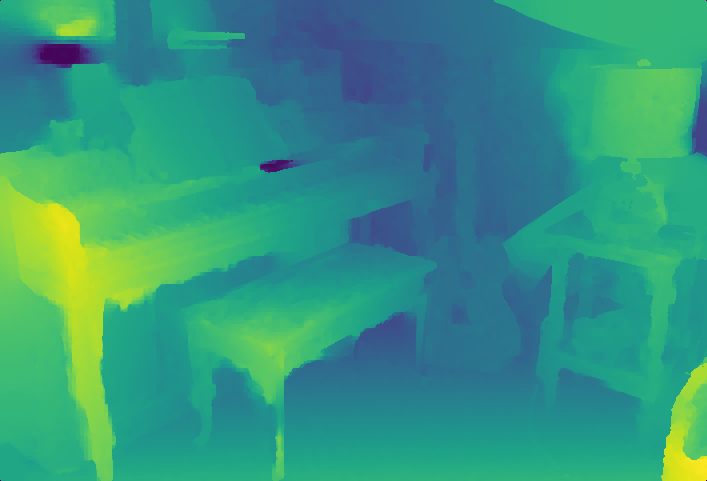} \\

      (a) \small{Image} & (b) \small{GT} & (c) \small{CCA}  & (d) \small{SGM}  & (e) \small{CCA $+$ filter} & (f) \small{SGM $+$ filter}
    \end{tabular}
    \end{center}
    \caption{SGM and CCA results on Middlebury dataset. Both algorithms perform similarly, however, CCA has lower space- and time-complexity as it does not require storing and aggregating the entire calculated cost-volume.}
 \label{fig:Stereo_result_main}
\end{figure*}

We compare our results to the original implementation of SDoF~\cite{Wadhwa_SIGGRAPH2018_syntheticDoF} (results and images are taken from~\cite{Garg-ICCV2019-learningDual}) in \figref{fig:PIXEL_result_main}. We also tested DPdisp~\cite{Punnappurath-ICCP2020-modelingDefocus} on this dataset; however, it under-performs on mobile devices as it is tailored for DSLR data as discussed in~\cite{Punnappurath-ICCP2020-modelingDefocus}. Hence, we do not include DPdisp~\cite{Punnappurath-ICCP2020-modelingDefocus} in our comparison. One can observe the above mentioned radial distortion in the results of our algorithm in the second-column. We use the error metrics suggested in~\cite{Garg-ICCV2019-learningDual}, which consider confidence in the GT inverse depth. In Table~\ref{tab:Pixel_res} we show a quantitative comparison between the algorithms. Even without radial disparity calibration, our algorithm slightly outperforms SDoF. Note that~\cite{Garg-ICCV2019-learningDual} provides better results than our algorithm. We do not include this algorithm in the comparison since its neural network is able to compensate for severe aberrations without additional calibration. This network, however, learns from data collected from a specific phone and is very unlikely to perform well on quite different devices, such as Canon DSLR. The low generalizabilty, in particular between DSLR and mobile devices is discussed in~\cite{Punnappurath-ICCP2020-modelingDefocus}. Adapting data from different DP devices and synthetic data is an ongoing research~\cite{Abuolaim_2021_ICCV, Pan_CVPR2021_dpexploration}; however, it has not been demonstrated successfully for disparity estimation. One could argue that per-device training is applicable; however, acquiring GT depth data for DP devices requires complicated setups~\cite{Garg-ICCV2019-learningDual, kang2021facial} that may not be available. Instead, our method is adaptable to any DP device. We provide additional results in the supplementary material, including results from~\cite{Garg-ICCV2019-learningDual}.

Our implementation of CCA algorithm includes non-optimized Matlab code. However, as discussed in Sec.~\ref{sec:complexity_analysis},
CCA has a lower complexity than SGM, which real-time execution has been demonstrated by several implementations~\cite{opencv_library,Michael2013}. A time comparison here is not fair because only \cite{Punnappurath-ICCP2020-modelingDefocus} and \cite{Pan_CVPR2021_dpexploration} provide their code and both infer disparities on CPU in minutes.



\subsection{Standard Stereo Cameras}\label{res:stereo}

To complete the evaluation of our algorithm, we test its performance on typical stereo data-sets (no DP). To adapt CCA to a standard stereo algorithm we modify two steps of the algorithm, the confidence score (Sec.~\ref{sec:Conf}) and the cost aggregation step (Sec.~\ref{sec:Agg}). Our original confidence score reduces confidence in regions with more than one local minima. For DP images the disparity range is rather small so considering a single minimum is enough. For wide baseline stereo, as the disparity range increases we need to consider several minima, so we adapt our confidence score to consider up to five minima. In addition, we use left-right consistency check to reduce confidence in non-consistent regions (e.g. occlusions) and speckle detection to reduce confidence in regions with small blocks of disparity. As for the aggregation step, we use a threshold $T$ which depends on the difference between the current pixel disparity $d$ and the minimizer of the previous parabola in the path $m_{p-1}$. If $|d - m_{p-1}| < T$ we use the original aggregation, as presented in Sec.~\ref{sec:Agg}. If $|d - m_{p-1}| \geq T$, we then assign $d = m_{p-1}$ if the difference between the curvature is large enough and no edge is detected between pixels (see supplementary for more details of the algorithm adaptation). 

\begin{table}[b]
\centering
\begin{tabular}{|c|c|c|c|c|}
    \hline
	Method & Bad px & Bad px &  Bad px  & RMSE \\ 
 	 & 0.5 $[\%]$ &  1 $[\%]$ &   2 $[\%]$ &  \\ 

    \hline
    SGM & 26.1 & 17.2 & 12.2 & 9.9 \\
    \hline 
    CCA & 26.2 & 18.3 & 13.2 & 5.2 \\ 
    \hline
    SGM $+$ filter& $\bold{23.5}$ & $\bold{15.2}$ & $\bold{10.5}$ & $\bold{4.04}$  \\
    \hline 
    CCA $+$ filter& 24.6 & 16.7 & 11.6 & $\bold{4.04}$  \\ 
    \hline
\end{tabular}
\caption{Comparison of SGM and CCA performance on Middlebury dataset (non-occluded areas). } 
\label{tab:Stereo_res}
\end{table}

We compare our algorithm to SGM with sub-pixel refinement using parabola fitting\footnote{SGM implementation from: \url{https://github.com/kobybibas/semi_global_matching}}. To ensure unbiased comparison we use the same cost function for both algorithms and compare both methods with and without a common post processing filter. We test both algorithms on Middlebury data-set~\cite{scharstein2014high} at $1/4$ of the original resolution. In Table~\ref{tab:Stereo_res} we report errors of both algorithms on non-occluded areas over the full data-set. Two representative examples are shown in \figref{fig:Stereo_result_main}. Both algorithms perform roughly the same, with SGM slightly out-performing CCA, mainly in smooth textureless areas such as the wall in the second example of \figref{fig:Stereo_result_main}. The large RMSE value of SGM before filtering is due to the many speckles of large disparities at the boundary of occluded pixels, which are usually filtered out after post-processing (see the first example of \figref{fig:Stereo_result_main}). 



\section{Conclusions}
\label{sec:conclusions}



This work proposes a continuous disparity estimation algorithm for DP stereo. 
The proposed algorithm (CCA) uses parabolas to represent continuous costs and computes the total cost by aggregating the parabola coefficients. This results in a quadratic total cost, which minimization is simple, fast and directly provides continues disparities. The multi-scale extension of this aggregation makes the solution robust to the varying PSF of DP images. 

Our evaluation shows that CCA attains SOTA performance on DP data from both DSLR and phone cameras. We also show that CCA performs similarly to SGM on standard stereo images, while being more memory and computationally efficient. To further improve the performance of CCA, future research will include better priors for multi-scale aggregation, improved cost calculation (e.g. deep features) and confidence scores, and higher-order polynomials for continuous cost representation.





{\small
\bibliographystyle{ieee_fullname}
\bibliography{dpstereo}
}

\clearpage 

\renewcommand{\thesubsection}{\Alph{subsection}}

\section*{Appendix}
\subsection{Adaptive Penalty }
As cost aggregation is performed along straight paths, this sometimes result in artifacts known as streaks, as two neighboring pixels share only a single path and the smoothness cost does not take into account that a possible depth edge is being crossed. In order to avoid these streaks, we reduce the penalty for non-smooth disparity transitions around the edges of the input images, which are assumed to be aligned with depth edges. Similar to~\cite{Hirschmuller-PAMI2008-sgm}, we use an adaptive parameter defined as: 
\BEN
P_a = P \cdot \Alpha_{p-1} \cdot \exp{\left(-(I_p - I_{p-1})^2/\sigma^2\right)},
\EEN
where $I_p$, $I_{p-1}$ are the image values of the current and previous pixel, respectively, in the aggregation path $r$. $P$ and $\sigma$ are configurable parameters that determine the amount of disparity smoothness and the magnitude of considered edges. A small value of $\sigma$ will increase aggregation across faint edges while a large value would allow the aggregation to be performed across harder edges making the resulting disparity smoother. 

\subsection{Multiple Iterations of Cost Aggregation} \label{sec:multi-iter}
As mentioned in the previous section cost aggregation along straight paths might result in streaking artifacts. To mitigate this problem, previous works suggested using quadrants instead of straight paths~\cite{facciolo2015mgm}. We suggest applying multiple iterations, thus extending the smoothness requirement beyond straight paths. After each iteration, due to the parabola propagation, the aggregated parabolas have a significantly higher curvature compared to the local, original parabolas. However, before starting each new iteration, we would like to have lower confidence values for pixels that had a low confidence in the first iteration. Therefore, to enable multiple iterations, we normalize the values of the coefficients at the end of each iteration as follows:
$$\alpha_p^{i+1} = N\Alpha_p^i,\beta_p^{i+1} = N\Beta_p^i,$$
 with $N=\alpha_p^1 / \overline{\alpha_p^1}$ and $i$ being the current iteration. $\alpha_p^{i+1},\beta_p^{i+1}$ are the coefficients after the $i_{th}$ iteration, $\alpha_p^1$ is the original coefficient value before aggregation, and $\overline{\alpha_p^1}$ is the mean value across the image.

\begin{table}[t!]
\centering
\begin{tabular}{|c|c|c|c|c|}
    \hline
	Method & AI(1) & AI(2) &  $ 1-|\rho_s|$ & Geometric \\ 
        &&&& Mean \\ 
    \hline
    1 Iteration & 0.055 & 0.083 & 0.092 & 0.075 \\
    \hline
    2 Iteration & 0.051 & 0.078 & 0.082 &  0.068\\
    \hline
    3 Iteration & 0.041 & 0.068 & 0.061 & 0.055  \\
    \hline
\end{tabular}
\caption{Comparison of different number of iterations on DSLR data~\cite{Punnappurath-ICCP2020-modelingDefocus}.}
\label{tab:supp_iter}
\end{table}
We test the effect of increasing the number of iterations of CCA. We test this on the DSLR-A data-set and the results are summarized in Table~\ref{tab:supp_iter}. We use three scales and the same number of iterations at each scale. The rest of the hyperparameters are as described in~\ref{tab:hyper}. As seen in the table, performance of CCA is improved as the number of iterations is increased.
 
\subsection{Comparison of Number of Scales of CCA}
As discussed in Section 3.4, CCA allows for multi-scale fusion, which improves disparity estimation. In this section, we test the effect of number of scales by varying the number of scale. We experiment with increasing the number of scales from one to three on the DSLR-A data-set. We use three iterations at each scale and the rest of the hyperparameters are as described in~\ref{tab:hyper}. As shown in Table~\ref{tab:supp_scale}, increasing the number of scales improves the quality of disparity estimation.

\begin{table}[b!]
\centering
\begin{tabular}{|c|c|c|c|c|}
    \hline
	Method & AI(1) & AI(2) &  $ 1-|\rho_s|$ & Geometric \\ 
        &&&& Mean \\ 
    \hline
    1 Scale  & 0.052 & 0.078 & 0.084 & 0.069 \\
    \hline
    2 Scales & 0.051 & 0.076 & 0.078 & 0.067\\
    \hline
    3 Scales & 0.041 & 0.068 & 0.061 & 0.055  \\
    \hline
\end{tabular}
\caption{Comparison of different number of scales on DSLR data~\cite{Punnappurath-ICCP2020-modelingDefocus}.}
\label{tab:supp_scale}
\end{table}

\subsection{Comparison of Cost Functions of CCA}
We performed two more tests on the DSLR-A data-set. First, we test different common cost functions used for local matching: sum of squared differences (SSD), sum of absolute differences (SAD), and normalized cross-correlation (NCC). While changing these costs, we still use ENCC for the initial sub-pixel estimation and optimize hyper-parameters for each cost. A comparison is shown in Table~\ref{tab:supp_score}. As seen, SAD outperforms both SSD and NCC. The same conclusion was drawn by~\cite{Heiko2009Eval}, that is, SAD outperforms other metrics when used as a cost function for SGM. 

\begin{table}[t!]
\centering
\begin{tabular}{|c|c|c|c|c|}
    \hline
	Method & AI(1) & AI(2) &  $ 1-|\rho_s|$ & Geometric \\ 
        &&&& Mean \\ 
    \hline
    SAD & 0.041 & 0.068 & 0.061 & 0.053 \\
    \hline
    SSD & 0.044 & 0.072 & 0.069 &  0.060\\
    \hline
    NCC & 0.046 & 0.078 & 0.081 & 0.066  \\
    \hline
\end{tabular}
\caption{Comparison of different cost score functions on DSLR data~\cite{Punnappurath-ICCP2020-modelingDefocus}.}
\label{tab:supp_score}
\end{table}

Next, we study different sub-pixel estimation methods for the initial sub-pixel estimation before aggregation. We tested three methods: traditional parabola fitting, histogram equalization interpolation suggested by~\cite{Miclea2015New} with a calibrated offset, and the interpolation function suggested by~\cite{Psarakis-ICCV2005-ecc}. Recall that this estimation regards the initial sub-pixel offset that shifts the initial parabola for better initialization; then, parabolas are used at the aggregation step. We use SAD with the same hyper-parameters in all these cases. The comparison is summarized in Table~\ref{tab:subpix_score}. As shown, when using ENCC, CCA provides slightly better results, showing that our method benefits from better sub-pixel estimation~\cite{Psarakis-ICCV2005-ecc}.

\begin{table}[t!]
\centering
\begin{tabular}{|c|c|c|c|c|}
    \hline
	Method & AI(1) & AI(2) &  $ 1-|\rho_s|$ & Geometric \\ 
        &&&& Mean \\ 
    \hline
    Parabola & 0.041 & 0.071 & 0.064 & 0.057 \\
    \hline
    Hist eq.\cite{Miclea2015New} & 0.041 & 0.070 & 0.063 & 0.056 \\
    \hline
    ENCC~\cite{Psarakis-ICCV2005-ecc} & 0.041 & 0.068 & 0.061 & 0.053 \\
    \hline
\end{tabular}
\caption{Comparison of different sub-pixel estimation methods for initial parabola positioning on DSLR data~\cite{Punnappurath-ICCP2020-modelingDefocus}.}
\label{tab:subpix_score}
\end{table}
\begin{figure*}[t]
  \begin{center}
   \begin{tabular}{c@{\hspace{5mm}}c@{\hspace{5mm}}c@{\hspace{5mm}}c}

       \includegraphics[width=0.24\linewidth]{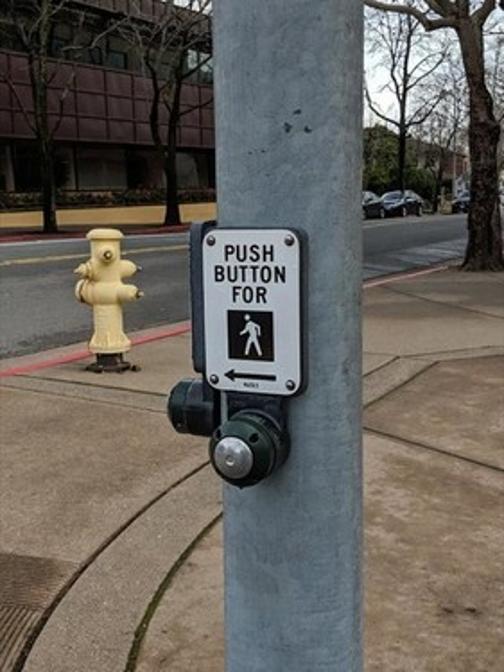} &
       \includegraphics[width=0.24\linewidth]{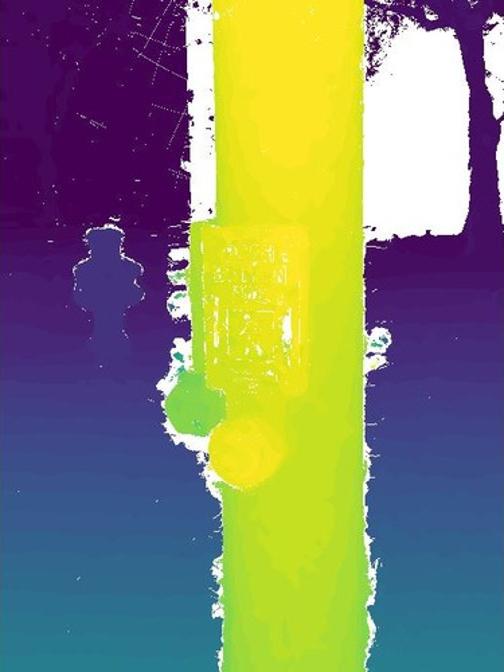} &
       \includegraphics[width=0.24\linewidth]{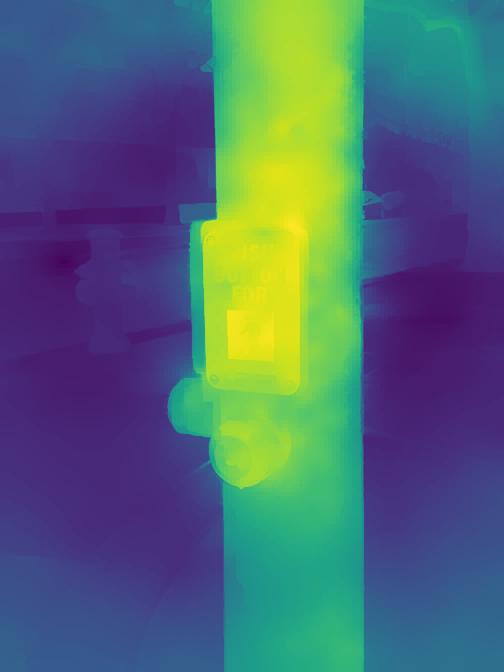} &
       \includegraphics[width=0.24\linewidth]{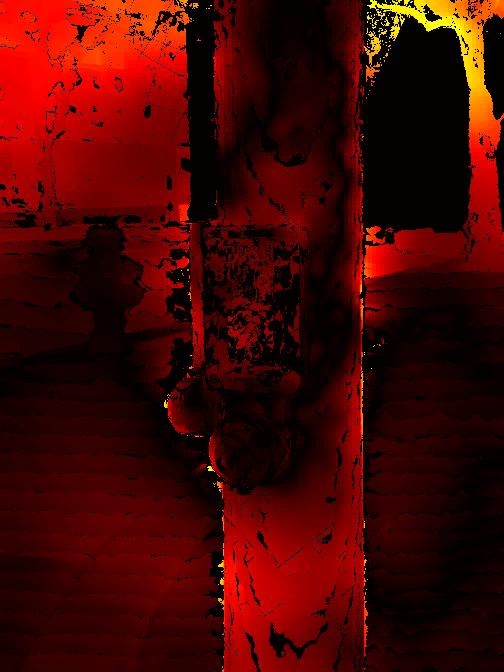} \\
       
      \includegraphics[width=0.24\linewidth]{\Figs/Pixel_result/Images/Im22.jpg} &
       \includegraphics[width=0.24\linewidth]{\Figs/Pixel_result/GT/GT22.jpg} &
       \includegraphics[width=0.24\linewidth]{\Figs/Pixel_result/CSGM_filter/22.jpg} &
       \includegraphics[width=0.24\linewidth]{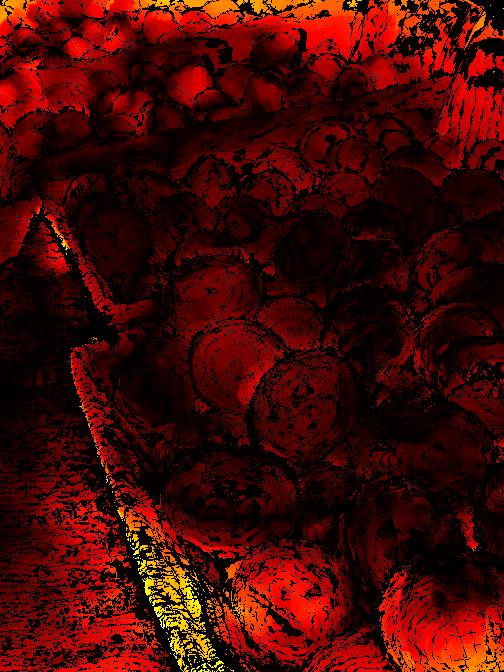} \\

      Image & GT & CCA + filter & Difference Image
    \end{tabular}
    \end{center}
 \caption{We show the difference between GT and CCA output. The rightmost image shows the difference between the GT and CCA estimation, where the hotter colors (i.e. white and yellow) denote larger errors. This emphasizes that disparity estimation is less accurate at the edges of the images because of spherical aberrations.}
 \label{fig:diff_im}
\end{figure*}

To calibrate for the offset of the interpolation function of~\cite{Miclea2015New}, we generated a random image, shifted it by values of $0.1,0.2,0.3...,1$ along the $x$-axis and calculated the sub-pixel disparity estimation. We then calculated the average error in disparity between the estimated and GT disparity. We calculated this parameter once and used it for all the experiments.


\subsection{DP Phone - Spherical Aberration }
Camera lenses suffer from different aberrations impacting the quality of produced images. Aberrations vary between different lenses, where high-end compound lenses use multiple lenses to compensate for them. For DP images, spherical aberrations create radial artifacts where objects near the edges appear closer~\cite{Wadhwa_SIGGRAPH2018_syntheticDoF}. These aberrations are significant for phone captured DP images.  In~\cite{Wadhwa_SIGGRAPH2018_syntheticDoF} the authors suggest a calibration procedure for spherical aberrations. In this work, we do not have such calibration data available and we show the effect of the aberrations in~\figref{fig:diff_im}.



\subsection{CCA Adaption for Large Disparities}
Estimating large disparities with CCA requires several adaptions. These adaptions need to consider regions of large repeated textures and untextured regions which might result in multiple local minimas. In addition, we aim to consider occluded areas which may not have a distinct minimum. To this end we use the confidence score $S_{confidence}$ described in Sec. 3.2 of the main paper and reduce confidence of parabolas in areas where the minimum is not distinct. We use four methods to detect suspicious pixels of low confidence. First, we reduce confidence of pixels where another minimum is close enough to the first minimum for large disparities:
\[    S_{confidence} = 
\begin{cases}
    \epsilon^2, & \text{if } C_{int}(p,d^k)-C_{int}(p,d^0) < T_d  \\ & \cap \; |d^k - d^{0}| > 2 \\ 
    1, & \text{otherwise}.
\end{cases}
\]
Where $T_d$ is a threshold to detect the difference between cost values of different disparities. Second, we adapt Eq. 3 from the main paper to consider more than one minima to multiple minimas, where we empirically found that considering five minimas performs optimally. Third, we use the initial disparity (before aggregation) and reduce confidence in small regions with distinct disparities from their neighborhood. Fourth, we perform left-right (L-R) check on the initial disparity and reduce confidence for inconsistent pixels.

During cost aggregation, SGM treats large disparity differences differently than small differences, thus resulting in a larger penalty. We adopt a similar scheme. To accommodate for large disparity differences we suggest altering the cost aggregation step (Sec. 3.3 of the main paper). As our algorithms allows for multiple iterations of cost aggregation (Sec. \ref{sec:multi-iter}) we use two different aggregation schemes depending on the iteration number. For the \emph{first} iteration, instead of propagating the parabola, we either propagate the coefficients from the previous pixel in the path or keep the current coefficients for large disparities. The propagation stops if the confidence is not high enough, or if an edge is detected. Our updated aggregation scheme is defined as follows: 
\[
    A_p = 
\begin{cases}
    \alpha_p + P_{adapt} \cdot \Alpha_{p-1},& \text{if } |d -m_{p-1}| < 2 \\
    \Alpha_{p-1}, & \text{if } |d -m_{p-1}| > 2 \\
    &  \cap \; \frac{\Alpha_{p-1}}{\alpha_p} > T_{prop} \\& \cap \; I_{edge} < T_{edge}  \\
    \alpha_p, & \text{otherwise}.
\end{cases}
\]
$T_{prop}$ is a threshold to detect if the confidence of the aggregated parabola is higher than the current parabola, $I_{edge} = \exp^{\left(-(I_p - I_{p-1})^2/\sigma^2\right)}$ is an operator used to calculate edges, and $T_{edge}$ is a threshold used to detect edges.  $\Beta_p, \Gamma_p$ follow the same rule of update. However, if multiple iterations of cost aggregations are used we update the rule as multiple iterations might create a strong bias to fronto-parallel surfaces. This is because pixels with high confidence could propagate their disparity to all neighboring pixels if there is no edge. Hence, for the next iterations, we update the aggregation scheme: 
\[
    \Alpha_p^l = 
\begin{cases}
    \alpha_p + P_{adapt}^1 \cdot \Alpha_{p-1}^{l-1},& \text{if } |d - m_{p-1}|<2 \\
    \alpha_p + P_{adapt}^2 \cdot \Alpha_{p-1}^{l-1},& \text{if } |d - m_{p-1}|\geq 2.
\end{cases}
\]
Where $l>1$ is the iteration number and
\BEAN
&P_{adapt}^1 = P_1 \exp^{\left(-(I_p - I_{p-1})^2/\sigma^2\right)}, \\
&P_{adapt}^2 = P_2 \exp^{\left(-(I_p - I_{p-1})^2/\sigma^2\right)},
\EEAN
and $P_2 < P_1$.


After the disparity estimation, we use common post-processing techniques to refine the disparity map. We use speckle reduction to remove small areas of isolated disparity and L-R consistency check to remove inconsistent disparities between the two views. To filter the result, we first use a median filter to fill the disparity of pixels detected in the previous step and then use bilateral filtering~\cite{barron2016fast} for final smoothing. To define a confidence map that guides the filtering, we use the valid pixels after speckle reduction and L-R consistency check (i.e. the confidence value is one for all pixels that remain after these two step, otherwise is zero). We use the same post-processing for SGM for a fair comparison.

\subsection{Configuration of Hyperparameters}

We provide all the hyperparameters of CCA in Table~\ref{tab:hyper}. Note that we use a sliding window approach to calculate the cost function, where we use a Gaussian weighted window; we denote the Gaussian's standard deviation by $\omega$.

\begin{table*}[t!]   
\centering
\begin{tabular}{|c|c|c|c|c|c|}
    \hline
	Parameter & Symbol & DSLR  & DSLR  &  Phone  & Middlebury  \\  
	 &  &  data-set A~\cite{Punnappurath-ICCP2020-modelingDefocus}  & data-set B~\cite{Punnappurath-ICCP2020-modelingDefocus} &  data-set~\cite{Garg-ICCV2019-learningDual} & data-set~\cite{scharstein2014high} \\  
    \hline
    Window STD & $\omega$ & 8 & 8 & 11 & 5 \\ 
    \hline
    Penalty factor & $P (P_1)$ & $3.2$ & $1.3$ & $7$ & $1$ \\
    \hline 
    Number of scales & & $3$ & $4$ & $2$ & $1$ \\
    \hline
    Number of iterations & & $[3,3,2]$ & $[2,2,3,6]$ & $[4,4]$ & $4$\\
    \hline
    Scale factor & $w$ & 1.5 & 2.5 & 0.4 & \\
    \hline
    Edge detect STD & $\sigma$ & 3.25 & 3 & 6 & 3\\
    \hline
    Ratio threshold  & $T_q$ & 2.2 & 2.2 & 2.2 & 2.2 \\
    \hline
    Threshold invalid & $T_a$ & $0.04$ & $0.075$ & $0.01$ & $0.001$\\
    \hline
    Threshold large disparity & $T_{d}$ &  &  & & $0.1$\\
    \hline
    Threshold edge & $T_{edge}$ &  &  & & $0.5$\\
    \hline
    Threshold aggregation & $T_{prop}$ &  &  & & $1000$\\
    \hline
    Penalty large factor & $P_2$ &  &  &  & $0.05$ \\
    \hline
    Bilateral STD luma~\cite{barron2016fast} & $\sigma_{luma}$ & 4 & 4 & 16 & $16$ \\
    \hline
    Bilateral STD color~\cite{barron2016fast} & $\sigma_{color}$ & 4 & 4 & 8 & 4 \\
    \hline
    Bilateral STD spatial~\cite{barron2016fast} & $\sigma_{xy}$ & 32 & 32 & 8 & 4 \\
    \hline
    Bilateral Smoothness~\cite{barron2016fast} & $\lambda$ & 512 & 512 & 15 & 1 \\
    \hline
    Guided Neighbourhood~\cite{he2012guided} &  & [75,75] & [75,75] & [15,15] &  \\
    \hline
    Guided Smoothness~\cite{he2012guided} &  & 0.2 & 0.2 & 0.1 &  \\
    \hline
\end{tabular}
\caption{Hyperparameters used in all experiments. We also include the the symbol as it appears in the main paper. At the end of the table we report parameters used for post-filtering. }
\label{tab:hyper}
\end{table*}


\end{document}